\documentclass[twocolumn]{fairmeta}
\usepackage{graphicx} 
\usepackage{placeins}
\usepackage{float}
\usepackage{xcolor}
\PassOptionsToPackage{hyphens}{url}
\usepackage{hyperref}
\usepackage{enumitem}
\usepackage{xcolor}

\definecolor{lrred}{RGB}{199,46,46}
\definecolor{noiseblue}{RGB}{39,92,213}

\makeatletter
\def\Hy@autoref@space{}%
\makeatother
\AtBeginDocument{%
}

\usepackage[most]{tcolorbox}
\tcbset{
  colback=gray!5,
  colframe=gray!50,
  boxrule=0.5pt,
  arc=3pt,
  left=6pt,
  right=6pt,
  top=6pt,
  bottom=6pt
}

\title{An Empirical Study on Noisy Data and LLM Pretraining Loss Divergence}

\author[1,\dagger]{Qizhen Zhang}
\author[\dagger]{Ankush Garg}
\author[1,2]{Jakob Foerster}

\author[\dagger,*]{Niladri Chatterji}
\author[\dagger, *]{Kshitiz Malik}
\author[2, *]{Mike Lewis}

\affiliation[1]{University of Oxford}
\affiliation[2]{FAIR at Meta}

\contribution[\dagger]{Work done at Meta}
\contribution[*]{Joint last author}

\abstract{
Large-scale pretraining datasets drive the success of large language models (LLMs). However, these web-scale corpora inevitably contain large amounts of noisy data due to unregulated web content or randomness inherent in data. Although LLM pretrainers often speculate that such noise contributes to instabilities in large-scale LLM pretraining and, in the worst cases, loss divergence, this phenomenon remains poorly understood.
In this work, we present a systematic empirical study of whether noisy data causes LLM pretraining divergences and how it does so. By injecting controlled synthetic uniformly random noise into otherwise clean datasets, we analyze training dynamics across model sizes ranging from 480M to 5.2B parameters.
We show that noisy data indeed induces training loss divergence, and that the probability of divergence depends strongly on the noise type, amount of noise, and model scale. We further find that noise-induced divergences exhibit activation patterns distinct from those caused by high learning rates \citep{wortsman2023small}, and we provide diagnostics that differentiate these two failure modes.  Together, these results provide a large-scale, controlled characterization of how noisy data affects loss divergence in LLM pretraining.
}

\date{\today}
\correspondence{qizhen.zhang@eng.ox.ac.uk}

\begin{document}

\maketitle

\section{Introduction}
\label{section:intro}
\begin{figure*}[h]
    \centering
    \includegraphics[trim=0 0 795 550, clip, width=0.24 \textwidth]{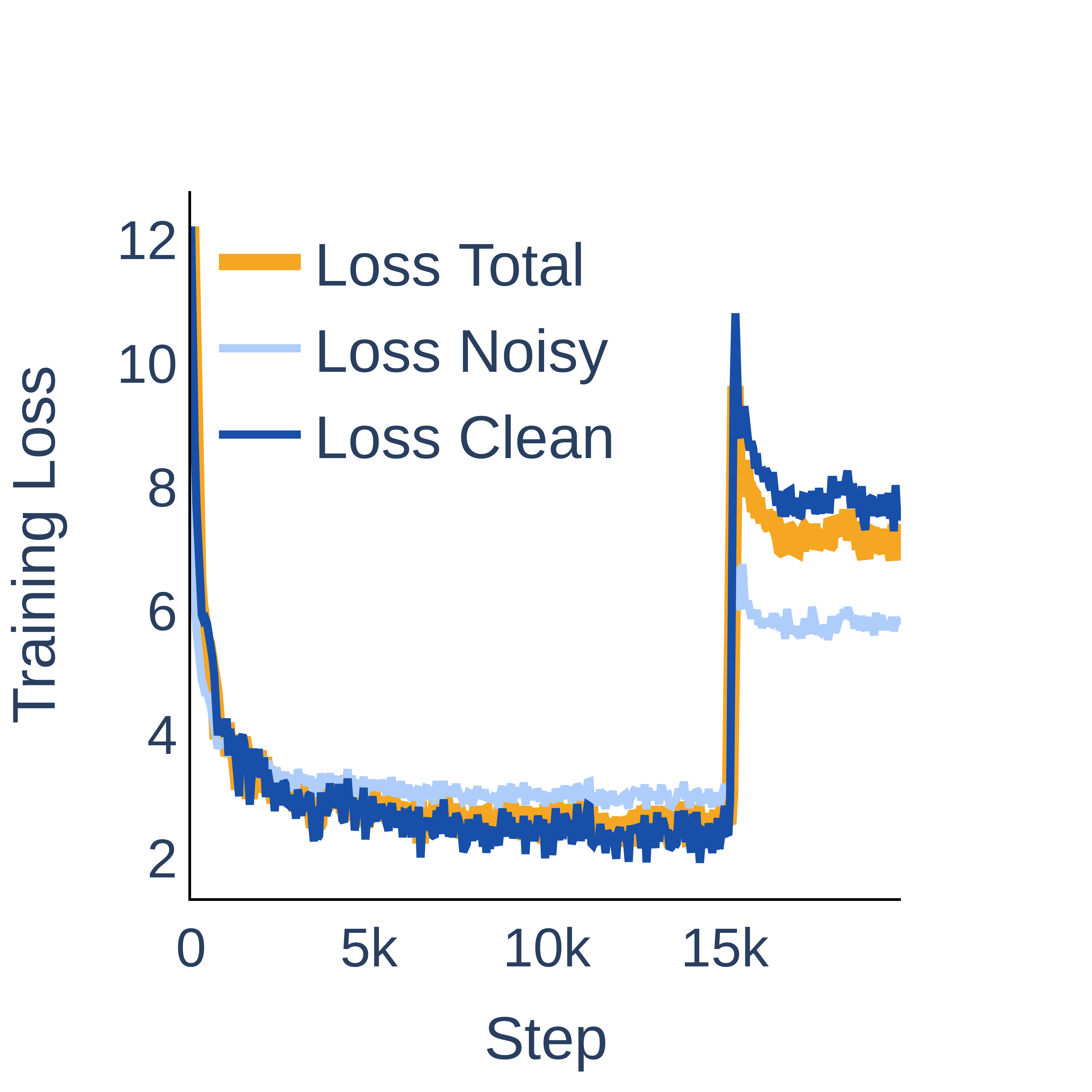}
    \includegraphics[trim=0 0 795 550, clip, width=0.24\textwidth]{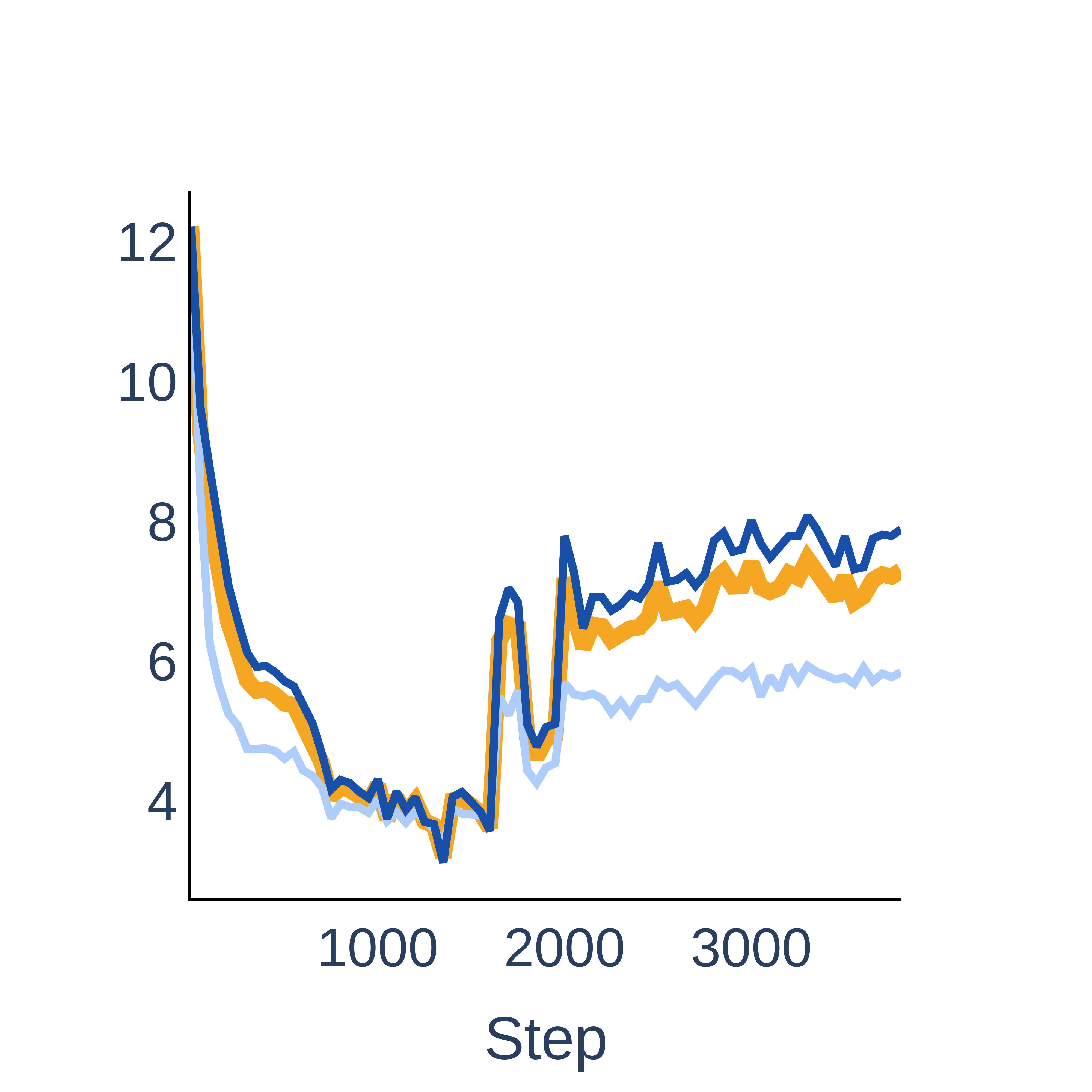}
    \includegraphics[trim=0 0 795 550, clip, width=0.24\textwidth]{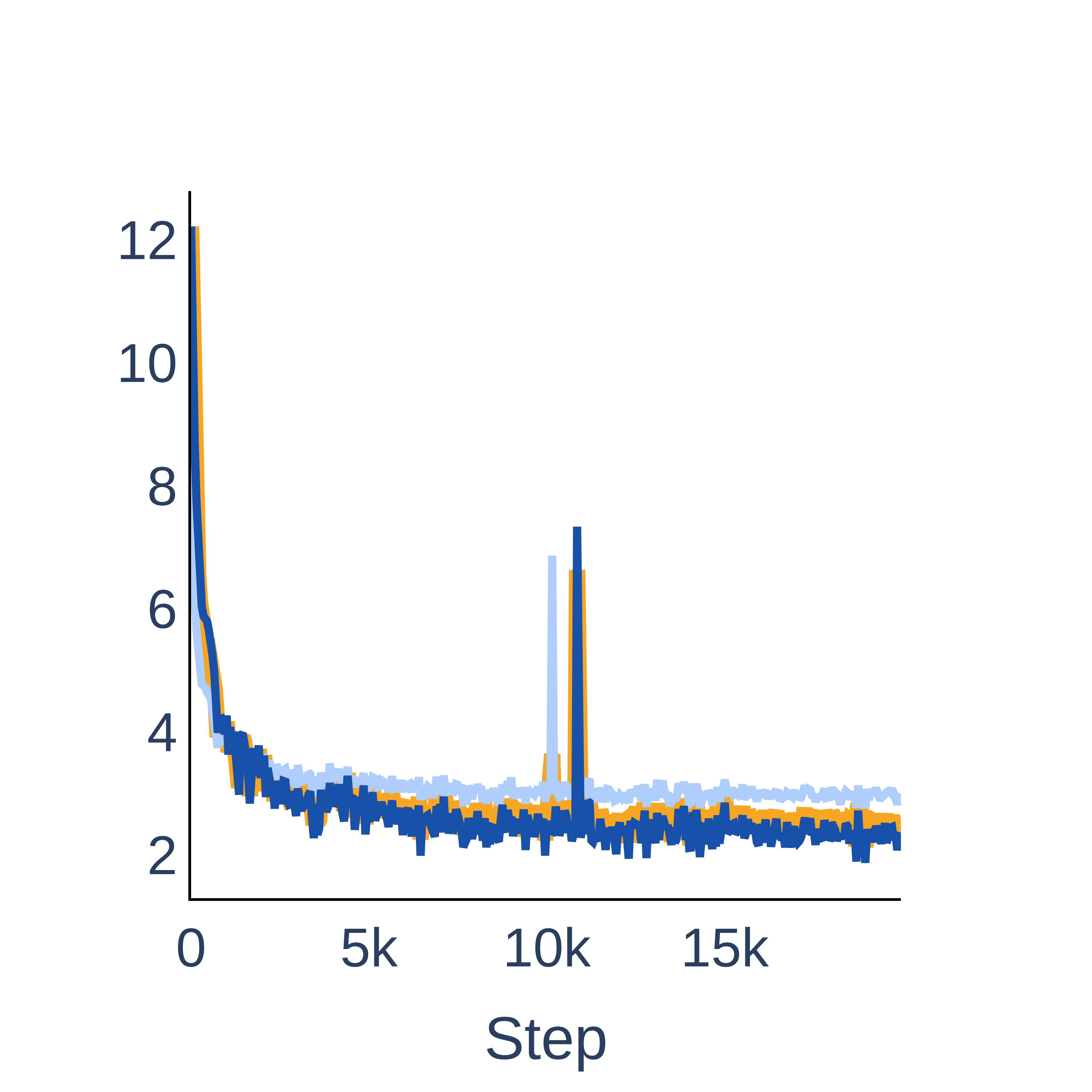}
    \includegraphics[trim=0 0 795 550, clip, width=0.24\textwidth]{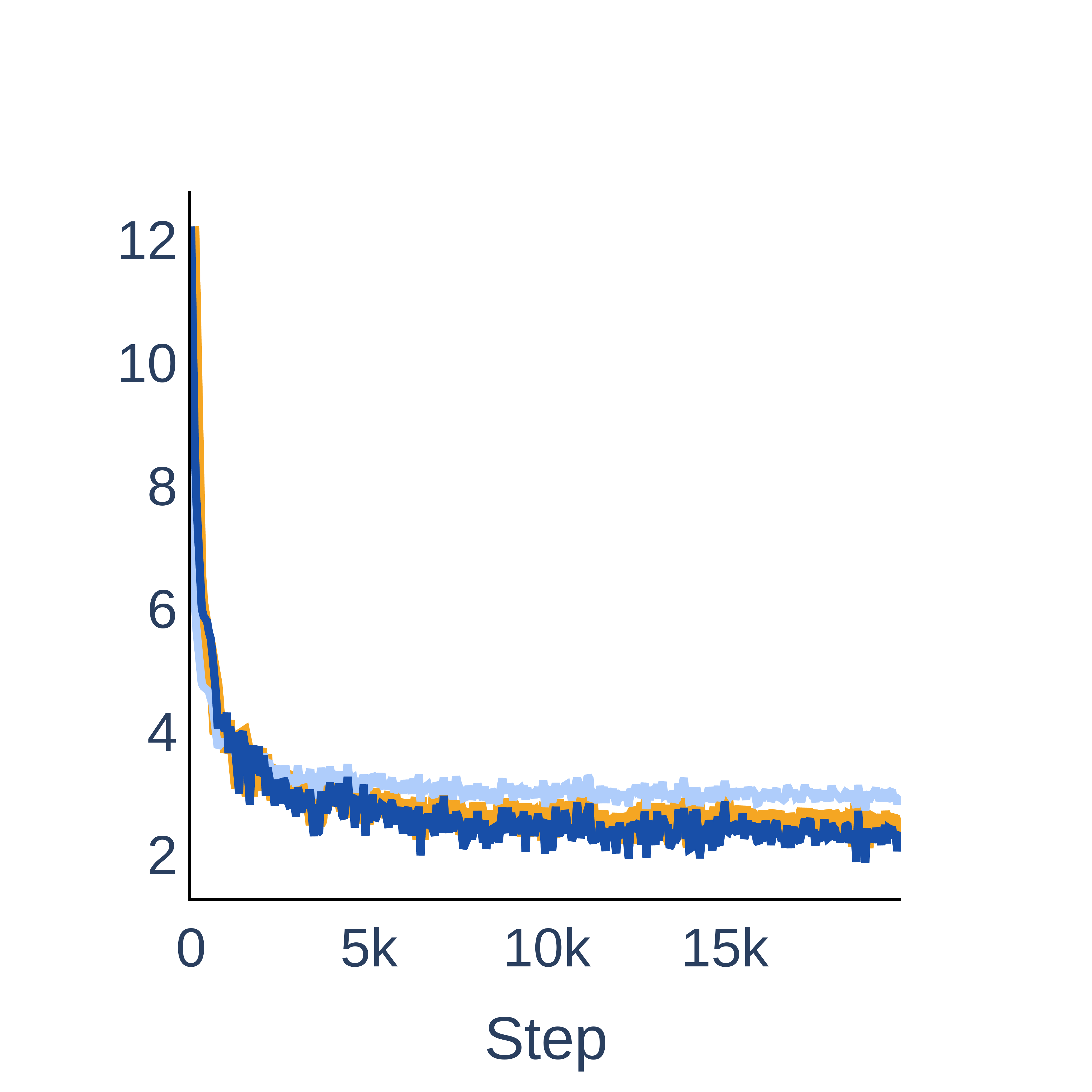}
    \caption{
    Examples of four pretraining runs using the same 1.3B model architecture and 15\% noisy data, differing only in random seed. The left two runs illustrate cases where training diverges, while the right two show stable behavior. We focus on divergences rather than fast spikes, so the third run is categorized as stable because its loss spikes quickly recover and the loss continues to decrease.
    }
  \label{fig:divergence_exmaples}
\end{figure*}

Scaling up transformer models in terms of model size and datasets has led to remarkable capabilities, but the resulting large-scale pretraining runs are extremely expensive \citep{dubey2024llama, adcock2026llama}. Furthermore, not every pretraining run succeeds \citep{chowdhery2023palm,zhang2022opt}: researchers frequently observe instabilities that slow or disrupt learning, and in the worst case, the pretraining loss diverges entirely. Understanding these failures is therefore essential. 

One hypothesized cause of loss divergence is noisy data in the training corpus. Web-scale datasets inevitably contain substantial noise due to unregulated web content or randomness inherent in data. Yet whether such noise truly drives divergence remains poorly understood.

In this work, we present a systematic empirical study of how uniform random noisy data impacts LLM pretraining stability. We focus on the worst-case instabilities that lead to loss \textit{divergence} rather than fast loss spikes. By injecting controlled synthetic uniform random noise into otherwise clean datasets, we examine training dynamics across model sizes from 480M to 5.2B parameters. Our main contributions are fourfold:

\begin{enumerate}[leftmargin=*]
\item \textbf{Noise can cause divergence \& the type of noise matters}
(\autoref{sec:types_of_noise}): We show that injected noise can indeed cause pretraining loss divergence. We also find that different noise types affect stability to varying degrees, and that some forms of noise cause divergence even in $<$ 1B parameter models, which are typically very stable to train.

\item \textbf{Scaling trends of noisy data}
(\autoref{sec:scaling_laws}): We show that higher data noise ratios increase the probability of loss divergence, and that larger models, particularly deeper models rather than wider models, are substantially more likely to diverge.

\item \textbf{Noisy data divergence differs from high learning rate divergence}
(\autoref{sec:lr_vs_noise}): We show that loss divergences induced by noisy data exhibit activation patterns that differ from those caused by overly high learning rates, allowing clear diagnostic separation between the two failure modes.

\item \textbf{Dense vs. MoE sensitivity to noisy data}
(\autoref{sec:moes}): Finally, we show that dense models and active-parameter-matched Mixture-of-Experts (MoE) models have comparable sensitivity to noisy data.
\end{enumerate}

Together, these results provide a large-scale, controlled characterization of how noisy data causes LLM pretraining loss divergence, which we hope will encourage future research to better understand and address the mechanisms through how data quality causes training instability.

\section{Related work}

\paragraph{LLM pretraining instabilities} 
Many studies analyze pretraining instabilities through the lens of abnormal activation and parameter behavior. Approaches such as QK-layer normalization \citep{dehghani2023scaling}, $z$-loss regularization \citep{chowdhery2023palm}, and weight decay \citep{loshchilov2017decoupled} stabilize training by modifying model architectures or loss functions to mitigate these abnormal behaviors. In contrast, relatively few works examine how \textit{training recipe decisions} give rise to these activation and parameter pathologies in the first place. Prior research in this direction has examined numerical precision and quantization, showing that reduced precision can induce pathological activation and gradient behavior \citep{micikevicius2017mixed,fishman2024scaling,peng2023fp8}, as well as learning rate choices, where \citet{wortsman2023small} demonstrate that overly high learning rates can cause abnormal activation and parameter growth and lead to loss divergence even in small language models. Our work studies an orthogonal training recipe choice: data quality. We show that noisy pretraining data can cause loss divergence, and that noisy data-induced failures exhibit activation and parameter patterns distinct from those caused by high learning rates. Based on this distinction, we introduce diagnostic methods that separate learning rate induced failures from noisy data induced failures, enabling practitioners to identify when data cleaning is required.

\paragraph{LLM pretraining data quality}
The most closely related work is \citet{ru2025we}, which argues that noisy data has a limited impact on training dynamics, reporting only modest loss increases and no training spikes or divergence. However, their analysis considers a narrow noise setting: injecting clean text with tokens sampled from the full tokenizer vocabulary. We show that noise type is important: certain forms of random noise, especially when restricted to a \textit{subset} of the tokenizer vocabulary, significantly increase the probability of loss divergence. By systematically varying noise types, noise ratios, and model sizes, we demonstrate that noisy data can destabilize training, and we characterize when this instability arises and how it can be diagnosed.

\paragraph{Noisy label works prior to LLMs}
A substantial body of theoretical and empirical research has studied neural network training under noisy labels in supervised learning settings \citep{zhang2018generalized,chen2019understanding,ghosh2017robust,sukhbaatar2014training,chen2023understanding, zhou2019toward, rolnick2017deep, patrini2017making, natarajan2013learning}. These works typically focus on small-scale models and datasets, studying how noisy labels affect model performance, generalization, and overfitting. In contrast, our work studies \textit{large-scale}, \textit{transformer-based} language models and investigates how noisy data leads to \textit{training loss divergence}. As demonstrated in \autoref{sec:scaling_laws}, these instabilities become increasingly pronounced at larger model scales: configurations that are stable for small models, where most prior noisy-label studies are conducted, do not remain stable when scaled to the regimes we consider. Moreover, we provide diagnostic insights for identifying noisy data induced failures specific to transformer architectures.

\section{Experimental methodology}
\subsection{Noisy data generation}
Our study focuses on \textit{uniform random noise}, which can arise from unregulated web content\footnote{Example: \url{https://docs.oracle.com/cd//E19528-01//820-0890//WebSvr.html\#wp35099}} or inherently random sequences such as hash codes\footnote{Example: \url{https://cdn.kernel.org/pub/linux/kernel/v6.x/ChangeLog-6.16?}}. Collecting a sufficiently large corpus of genuine random noise from web crawls is computationally infeasible. Instead, we simulate uniform random noise synthetically in a controlled setting. 

We construct noisy training data from a clean corpus $D_c$, consisting of a subset of the Llama 4 pretraining data mixture \citep{adcock2026llama}. Let $V$ denote the tokenizer vocabulary and let $\alpha \in (0,1)$ be the target noise ratio, the percentage of tokens that are artificial noise. Each noise token is sampled independently from a designated noise vocabulary, $n_i \sim \text{Uniform}(V_N)$, where $V_N \subseteq V$. We introduce a restricted noise vocabulary because real-world noise typically occupies a limited subset of the tokenizer vocabulary, for example, hash-like strings are composed of digits and lowercase letters (e.g., \texttt{4f2a9c1e0d}). We defer details on the choice of $V_N$ to \autoref{sec:types_of_noise}. For each document with $n$ tokens, we inject uniform random noise using one of the following two approaches. \textbf{Inserting:} We sample $\tfrac{n\alpha}{1-\alpha}$ insertion positions. At each sampled position $i$, we insert a noise token $n_i$ between the $i$-th and $i+1$-th clean tokens. Positions are sampled with replacement, allowing consecutive noise tokens.
\textbf{Overwriting:} Each token in the document is replaced by a noise token with probability $\alpha$.

\begin{figure}[t]
  \includegraphics[trim=0 0 145 350, clip, width=0.49\textwidth]{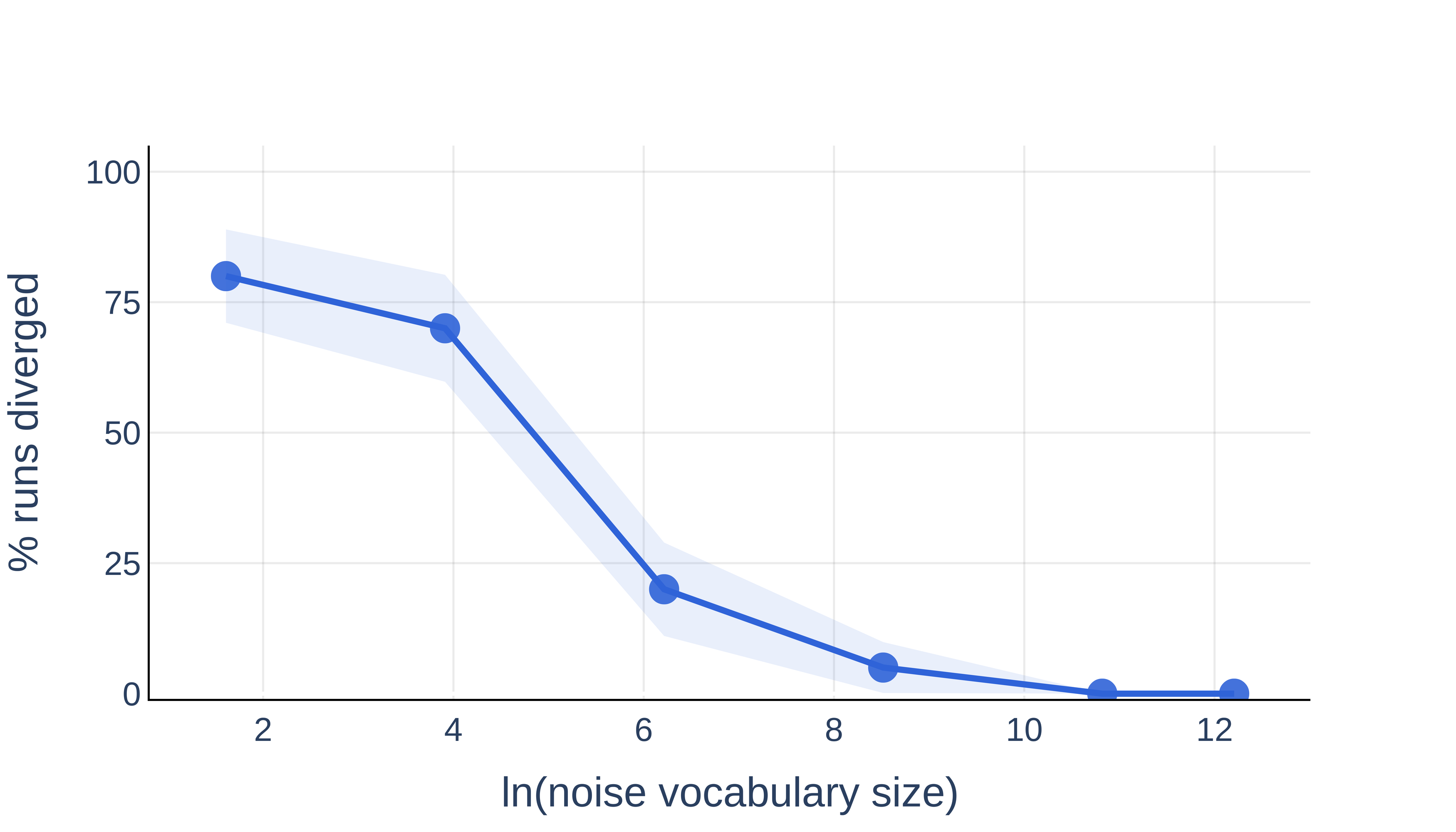}
  \caption{
\textbf{Effect of noise vocabulary size on training stability} for 540M dense models with noise ratio ${\alpha = 55\%}$.
We observe that reducing the size of the noise vocabulary significantly increases the probability of divergence. In all plots throughout the paper, we use shaded regions to denote the standard error.}
  \label{fig:noise_vocab}
\end{figure}

\subsection{Dense model architecture details}
We train decoder-only transformer models \citep{vaswani2017attention} using the standard auto-regressive next-token prediction objective. Our dense architecture follows the Llama 3 family of models \citep{dubey2024llama}: we use pre-normalization transformers, we do not use QK-layernorm \citep{dehghani2023scaling}, we do not use biases, we do not tie the input and output embedding weights, and we use rotary positional embeddings \citep{su2024roformer} and group query attention \citep{ainslie2023gqa}. ~\autoref{tab:model_configs} summarizes the shared hyperparameters across all models.

To vary model size, we adjust the number of layers and the model dimension which are detailed in \autoref{sec:scaling_laws} and \autoref{tab:arch_more_details}. We always jointly scale up the model dimension and number of query heads. 
Reported parameter counts include all parameters, including input and output embeddings.

\subsection{MoE architecture details}

For Mixture of Experts \citep[MoE;][]{shazeer2017outrageously,fedus2022switch} experiments, we use dropless MoEs \citep{gale2023megablocks, liu2024deepseek} with 16 feed-forward network (FFN) experts and token-choice top-2 routing (see \autoref{sec:apdx_moe_arch} for more details).  
For fair comparison with dense models, we active-parameter match top-$2$ MoE models by scaling the FFN dimension by $0.5$. All other architectural components remains the same as the dense setting.

\begin{figure}[t]
  \includegraphics[trim=0 0 145 350, clip, width=0.49\textwidth]{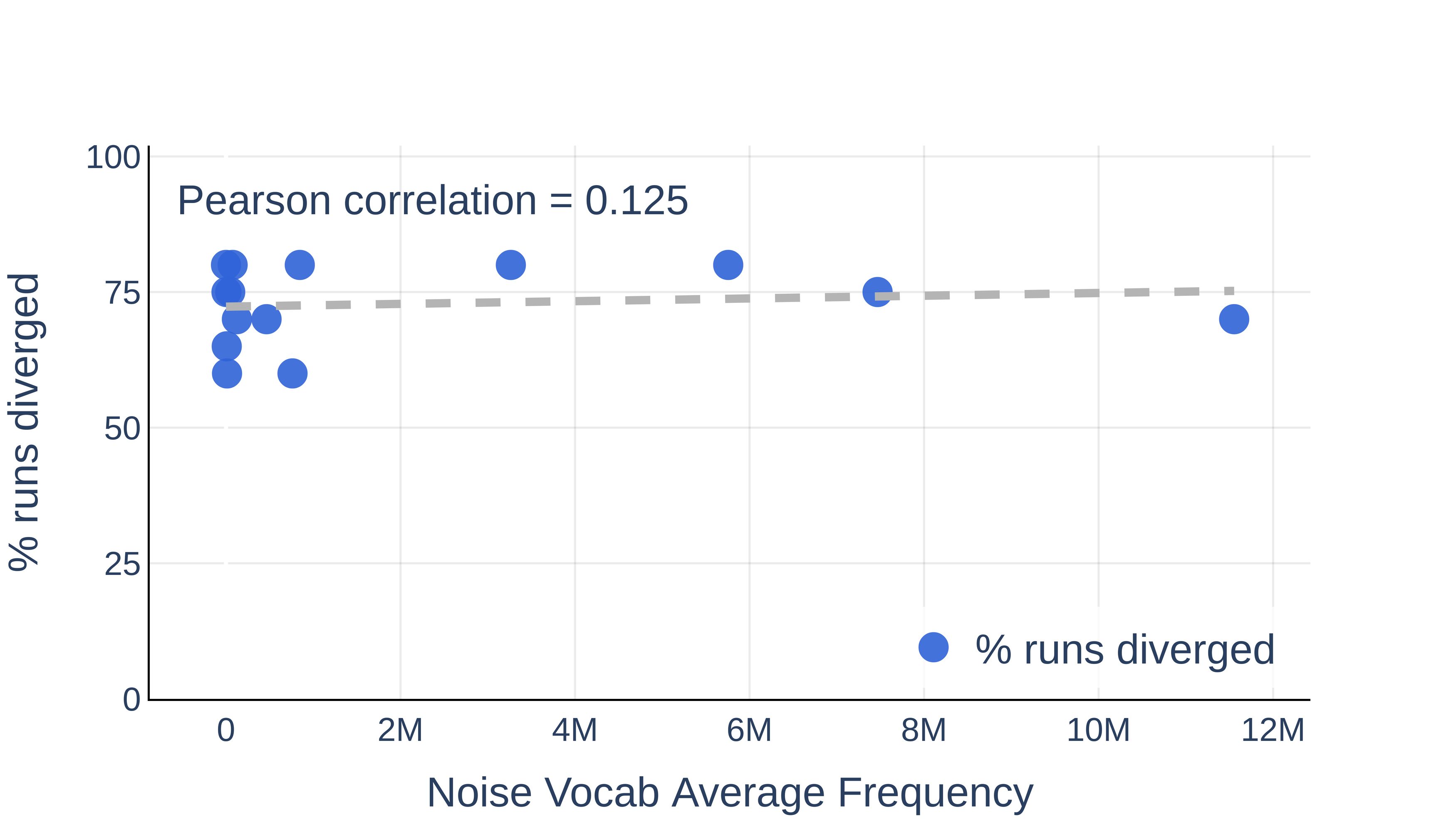}
  \caption{\textbf{Effect of noise vocabulary content on training stability.} Fixing the noise vocabulary size to $|V_n|{=}5$, we plot the divergence rate against the average frequency of the selected noise tokens in the clean corpus. We observe that the content of the noise vocabulary has little effect on loss divergence, with a near zero Pearson correlation.}
  \label{fig:noise_content}
\end{figure}

\subsection{Training details}
We use AdamW \citep{loshchilov2017decoupled} with $\beta_1 = 0.9$, $\beta_2 = 0.95$, $\epsilon=1e-8$, global norm gradient clipping of $1.0$, and  weight decay $1\mathrm{e}-4$. We apply a learning rate (LR) schedule with a $2000$-step linear warm-up followed by cosine decay, with a minimum LR ratio of $0.1$ and $2\mathrm{e}{4}$ total steps.  We use a peak LR of $lr=1.85e-2$ and a batch size of $2.6\mathrm{e}{5}$ tokens, and we train for less than one epoch in all experiments. To enable hyperparameter transfer across model sizes, we adopt Maximal Update Parametrization \citep{yang2021tuning} with a base width of $256$. All experiments are trained on 16 to 64 H100 GPUs using PyTorch FSDP2 \citep{pytorch}. 

\begin{figure}[t]
  \centering
  \includegraphics[trim=0 0 145 350, clip, width=0.49\textwidth]{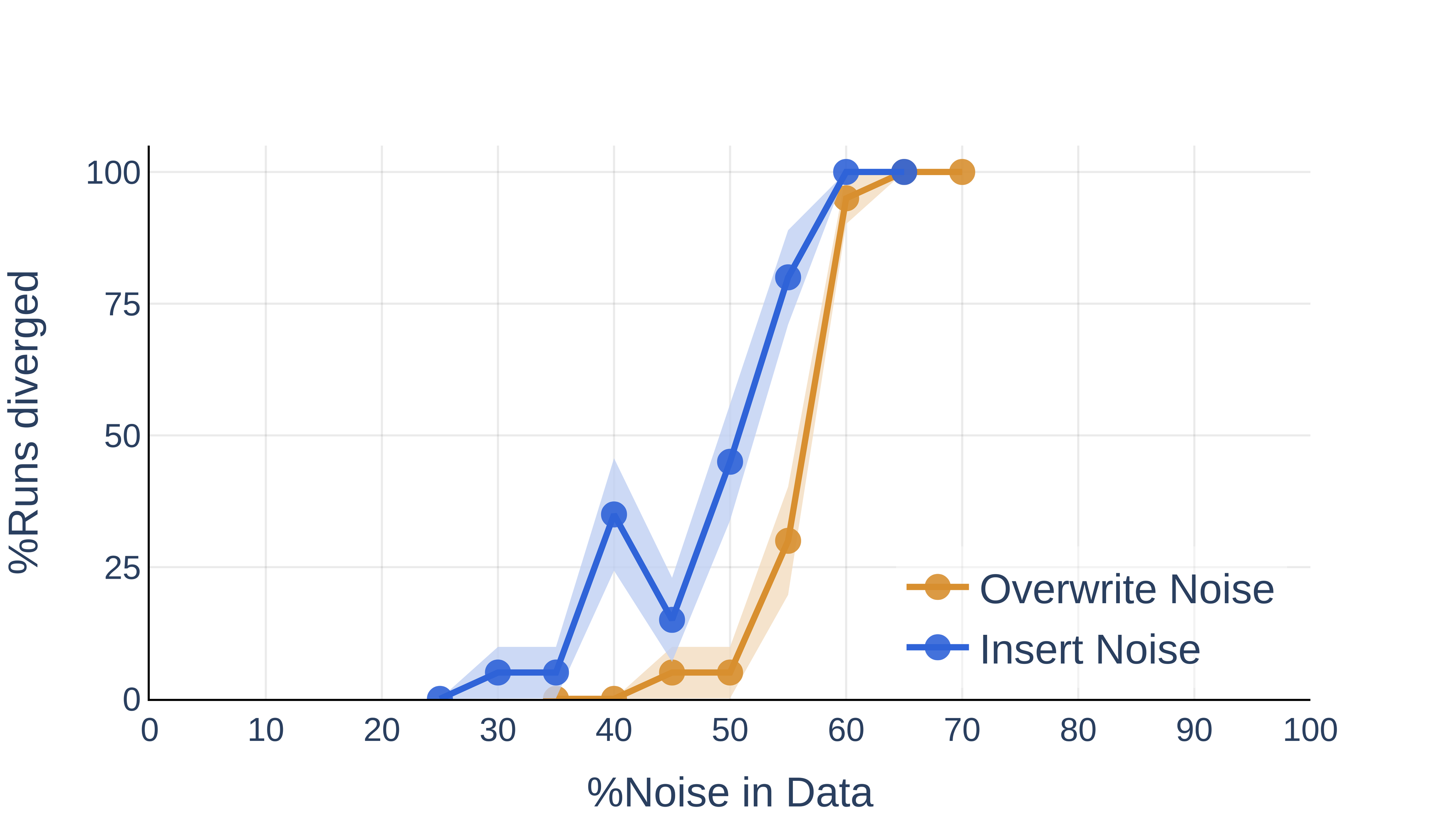}
  \caption{\textbf{Comparison of inserting versus overwriting noise} for a fixed noise vocabulary of five tokens. Inserting noise results in a higher probability of loss divergence than overwriting noise.}
  \label{fig:insert_noise}
\end{figure}
\subsection{Measuring training stability}
Training stability is inherently non-deterministic. Even with identical architectures and training setups, runs that differ only by random seed can behave differently\footnote{In addition to different initialization, non-determinism from GPU execution may also be a cause.}. As shown in \autoref{fig:divergence_exmaples}, among four such runs, two diverge while two remain stable. Because this work focuses on loss \textit{divergence}, we treat the third run as stable, since it quickly recovers from a loss spike and continues to decrease. We mark a run as diverged if its loss exceeds the minimum observed so far by more than 0.5 nats/token for at least 600 consecutive steps. To quantify training stability, we estimate the probability of divergence by repeating each experiment with 20 random seeds and reporting the percentage of runs that diverge. To ensure that any observed divergences are due to the injected noise, we verify that for all model sizes studied, none of the 20 seeds trained on the clean corpus $D_c$ (i.e., without any noise injection) diverges when using the same training setup and hyperparameters.

For analyses comparing relative stability across settings (\autoref{sec:types_of_noise}, \autoref{sec:scaling_laws}), we instead use an LR schedule designed for 15 trillion tokens, following \citet{dubey2024llama}, to approximate large-scale pretraining. Due to computation constraints, these runs are truncated at $2\mathrm{e}{4}$ steps, and we consider only divergences occurring within this horizon.

\section{Results and analysis}
We first show that uniform random noise can indeed cause pretraining loss divergence and identify the most destabilizing noise types (\autoref{sec:types_of_noise}). Using the most destabilizing noise type, we then study how divergence scales with noise ratio $\alpha$ and model size (\autoref{sec:scaling_laws}). Next, we present diagnostics that distinguish divergences caused by high LRs, as described by \citeauthor{wortsman2023small}, from those induced by noisy data (\autoref{sec:lr_vs_noise}). Finally, we show that dense and MoE models exhibit similar sensitivity to noisy data training conditions (\autoref{sec:moes}).

\subsection{Noisy data causes divergence \& the type of noisy data matters}
\label{sec:types_of_noise}
We first establish that noisy training data can induce loss divergence, and then study which types of noise are most destabilizing. All experiments in this subsection use a noise ratio of $\alpha = 55\%$ and a 540M dense model following the architecture in \autoref{tab:model_configs} and \autoref{tab:arch_more_details}, with model dimension 1024 and 10 layers. 

\begin{tcolorbox}
\textbf{Question:} Can noisy data cause training divergence? If so, does the \textbf{size} of the noise vocabulary $|V_n|$ affect the probability of divergence?

\textbf{Answer: Yes to both.}
\end{tcolorbox}

\begin{figure}[t]
  \centering
  \includegraphics[trim=0 0 145 350, clip, width=0.49\textwidth]{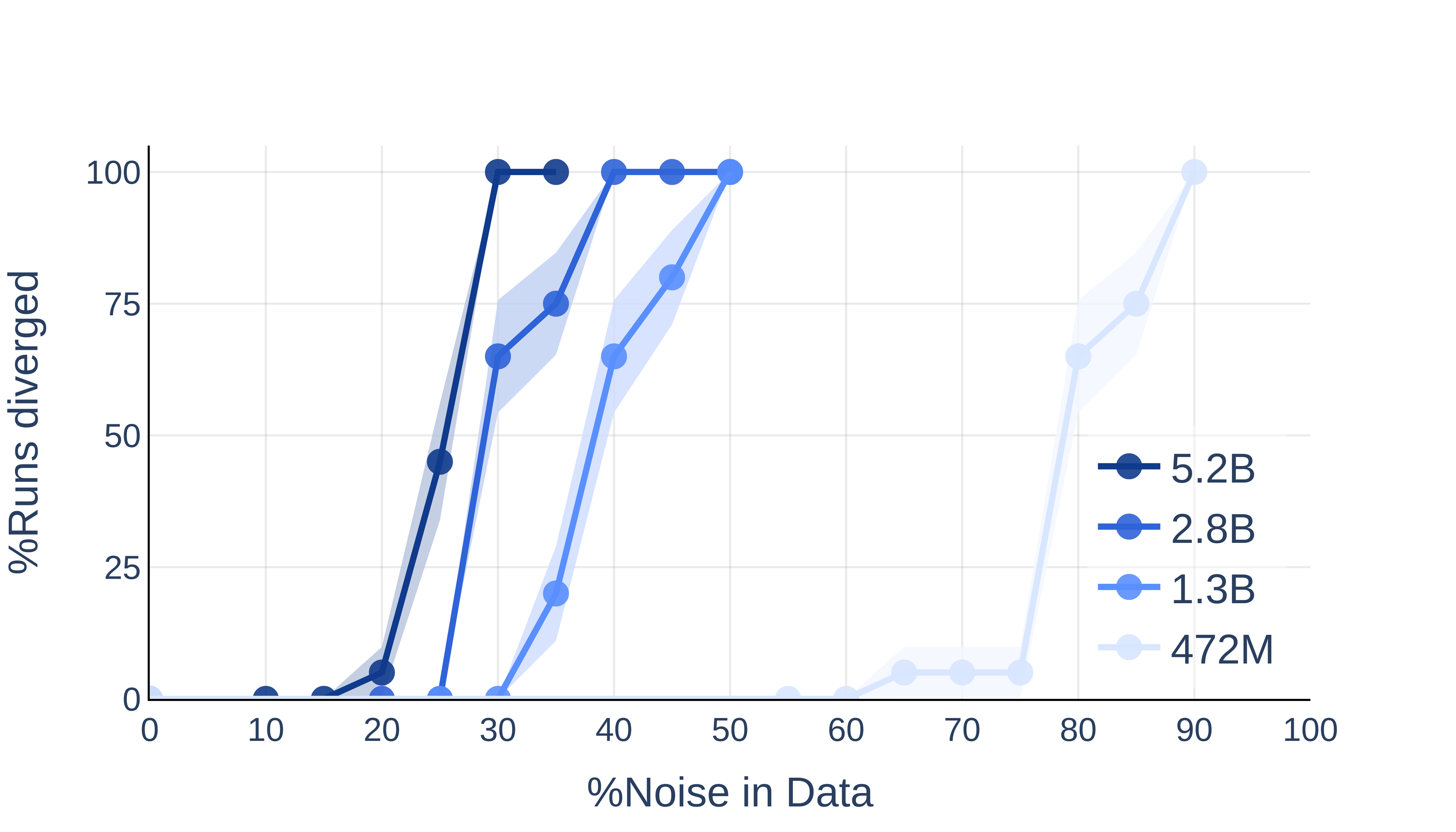}%
  \caption{Scaling model size by jointly scaling the depth and width of the model. We observe that larger models are more sensitive to noise, and a higher noise ratio leads to more divergences.}
  \label{fig:scaling1}
\end{figure}

\begin{figure*}[t]
  \centering
    \includegraphics[trim=0 0 145 350, clip, width=0.49\textwidth]{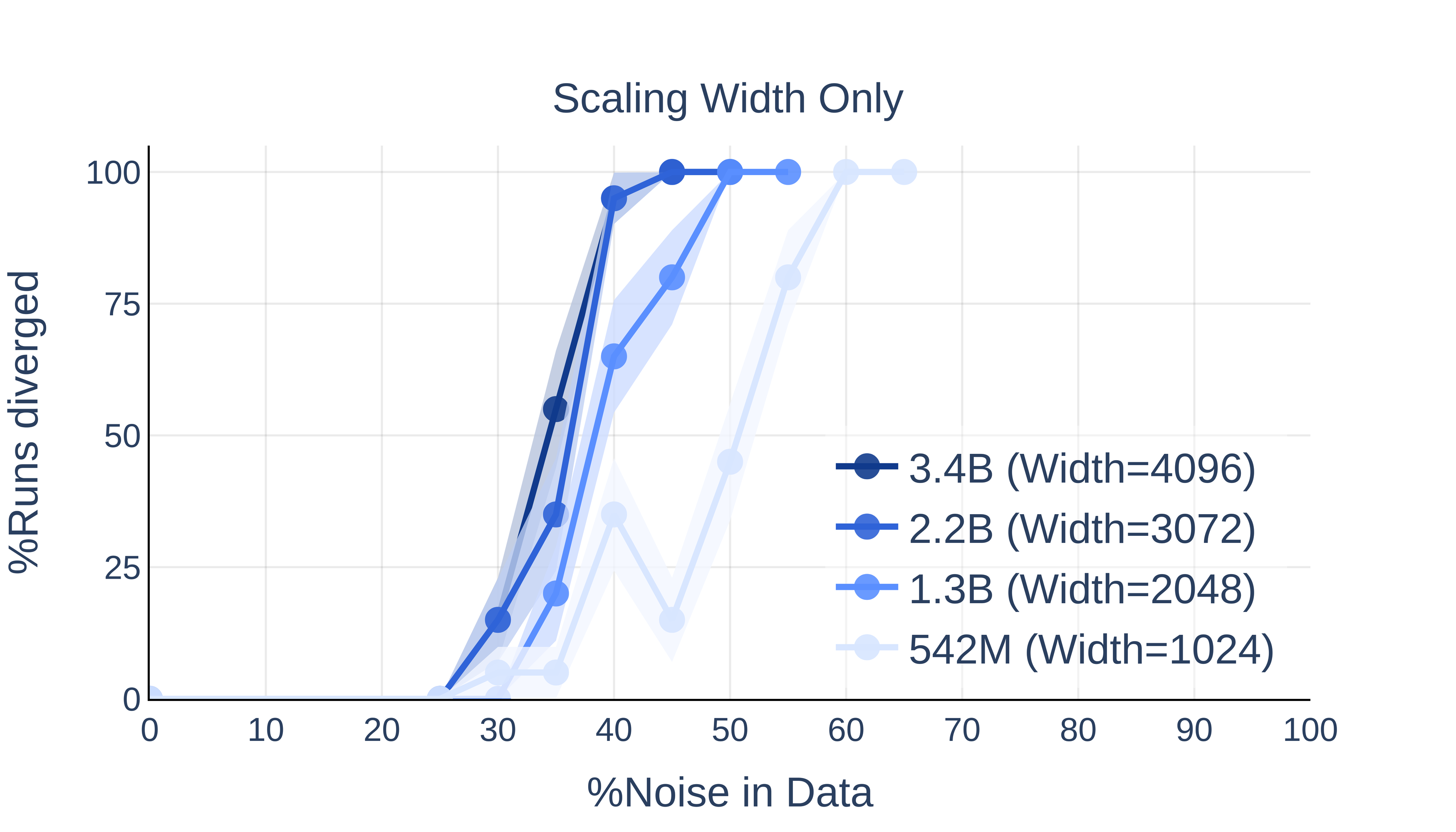}
    \includegraphics[trim=0 0 145 350, clip, width=0.49\textwidth]{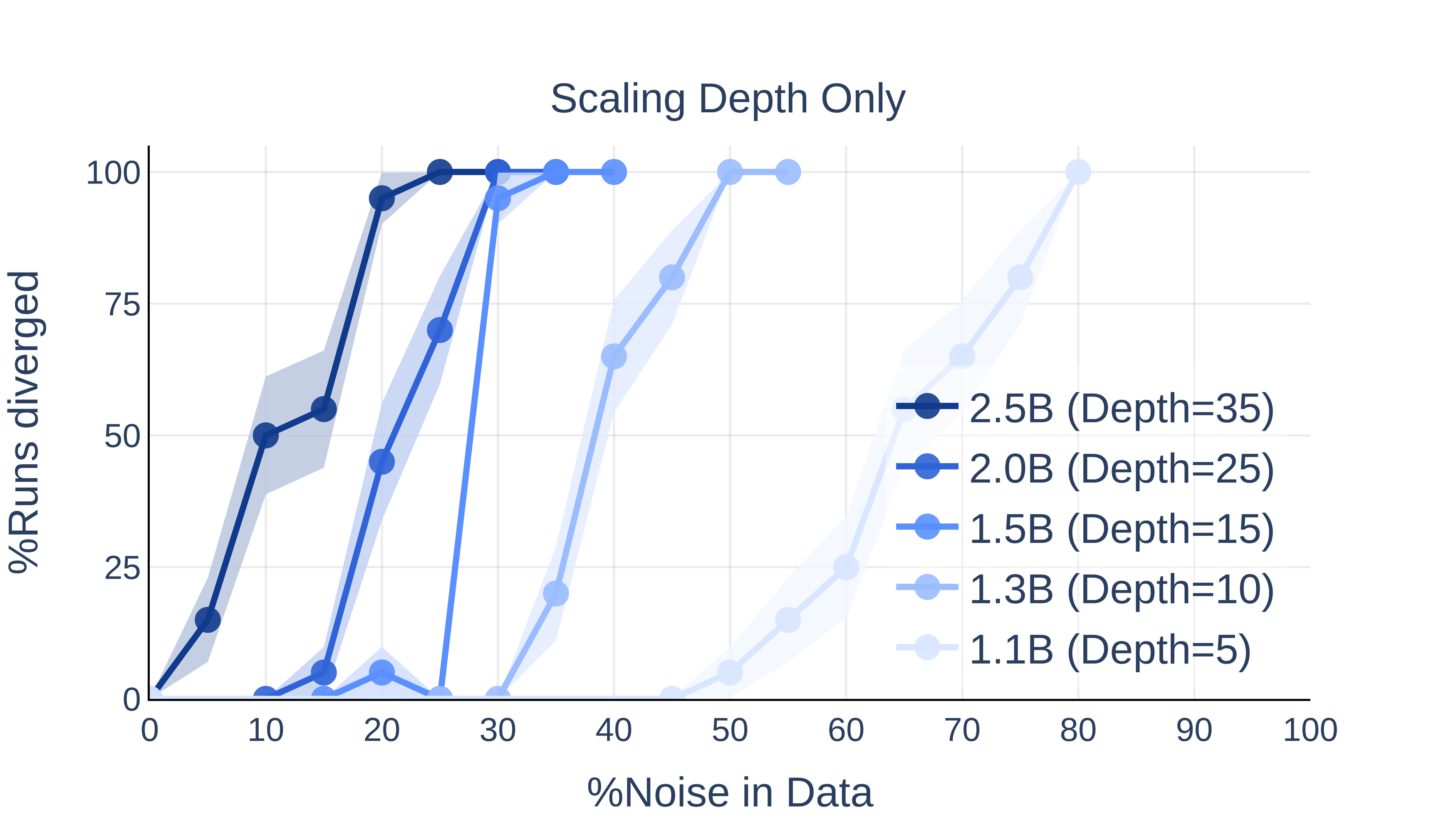}
    \label{fig:scaling_depth}
  \caption{\textbf{Left:} Scaling model size by width only. \textbf{Right:} Scaling model size by depth only. We observe that increasing depth induces more divergences than increasing width, with deeper models diverging far more frequently.}
  \label{fig:scaling_width_vs_depth}
\end{figure*}

In \autoref{fig:noise_vocab}, we vary the noise vocabulary size $k=|V_n|$ by restricting noise tokens to the first $k \in {5, 50, 500, 5000, 50000}$ tokens in the tokenizer vocabulary, and also consider using the full vocabulary ($|V_n| = |V| = 200{,}000$). These experiments use insertion noise. We defer discussion of alternative choices for the $k$ tokens, as well as insertion versus overwriting noise, to later in this section. We see that injecting noise indeed induces loss divergence, with smaller noise vocabularies increasing the probability of divergence. 

\begin{tcolorbox}
\textbf{Question:} Does the \textbf{content} of the noisy token vocabulary affect the probability of divergence?

\textbf{Answer: No.}
\end{tcolorbox}

One may suspect that inserting common versus rare tokens as noise would affect training dynamics differently. To test this, we fix the noise vocabulary size to $|V_n| = 5$, and we vary its \textit{content} by selecting different sets of five tokens from the tokenizer vocabulary $V$. We construct token sets ${t_1,\dots,t_5} \subset V$ whose average frequency in the clean corpus $D_c$ span several orders of magnitude, ranging from sets of extremely common tokens (each appearing on average $\sim$ 12 million times) to sets of tokens that never appear in $D_c$.

In \autoref{fig:noise_content}, the x-axis shows the average frequency of the five noise tokens in the clean dataset,
$
\frac{1}{5} \sum_{i=1}^{5} \mathrm{frequency}(t_i, D_c),
$
while the y-axis reports the percentage of runs that diverge. We find that the actual noisy token content has minimal effect on training stability, with a near-zero Pearson correlation of $0.125$.

\begin{tcolorbox}
\textbf{Question:} Does \textbf{inserting vs overwriting} noise affect divergence in LLM pretraining?

\textbf{Answer: Yes.}
\end{tcolorbox}
  
In \autoref{fig:insert_noise}, we compare inserting versus overwriting noise, fixing the noise vocabulary to the first five tokens in the tokenizer vocabulary for both settings. We observe that inserting noise leads to higher probability of loss divergence than overwriting noise.

\subsection{Scaling trends of noisy data} 
\label{sec:scaling_laws}
Next, we study how sensitivity to noisy data scales along two axes: model size and noise ratio. For all remaining experiments, we use the most destabilizing noise setting identified in \autoref{sec:types_of_noise}, namely inserting noisy tokens from a noise vocabulary of size $|V_n|=5$.

In \autoref{fig:scaling1}, we scale up model size by jointly increasing the number of layers and the model dimension while maintaining a fixed ratio
$
\frac{\text{num\_layers}}{\text{model\_dim}} = 204.8.
$
We consider four dense models ranging from 472M parameters (1024 dimensions, 5 layers) to 5.2B parameters (4096 dimensions, 20 layers). We observe that i) at a fixed noise ratio, larger models diverge more frequently than smaller ones, and ii) increasing the noise ratio consistently raises the divergence rate across all model sizes.

In \autoref{fig:scaling_width_vs_depth}, we isolate the effects of width and depth. The left panel fixes depth at 10 layers and scales width from 1024 to 4096, increasing model size from 542M to 3.4B parameters. The right panel fixes the width at 2048 and scales the depth from 5 to 35 layers, increasing the model size from 1.1B to 2.5B parameters. Scaling width has a limited effect on training stability: despite spanning a larger parameter range, width-scaled models exhibit more similar divergence rates. In contrast, increasing depth substantially degrades stability. In the most extreme case, the 35-layer, 2.5B parameter model diverges in 15\% of runs even at a noise ratio of 5\%.

\subsection{Divergences due to high LR vs. noisy data}
\label{sec:lr_vs_noise}
\begin{figure*}[th]
  \includegraphics[trim=0 0 45 0,width=0.95\textwidth]{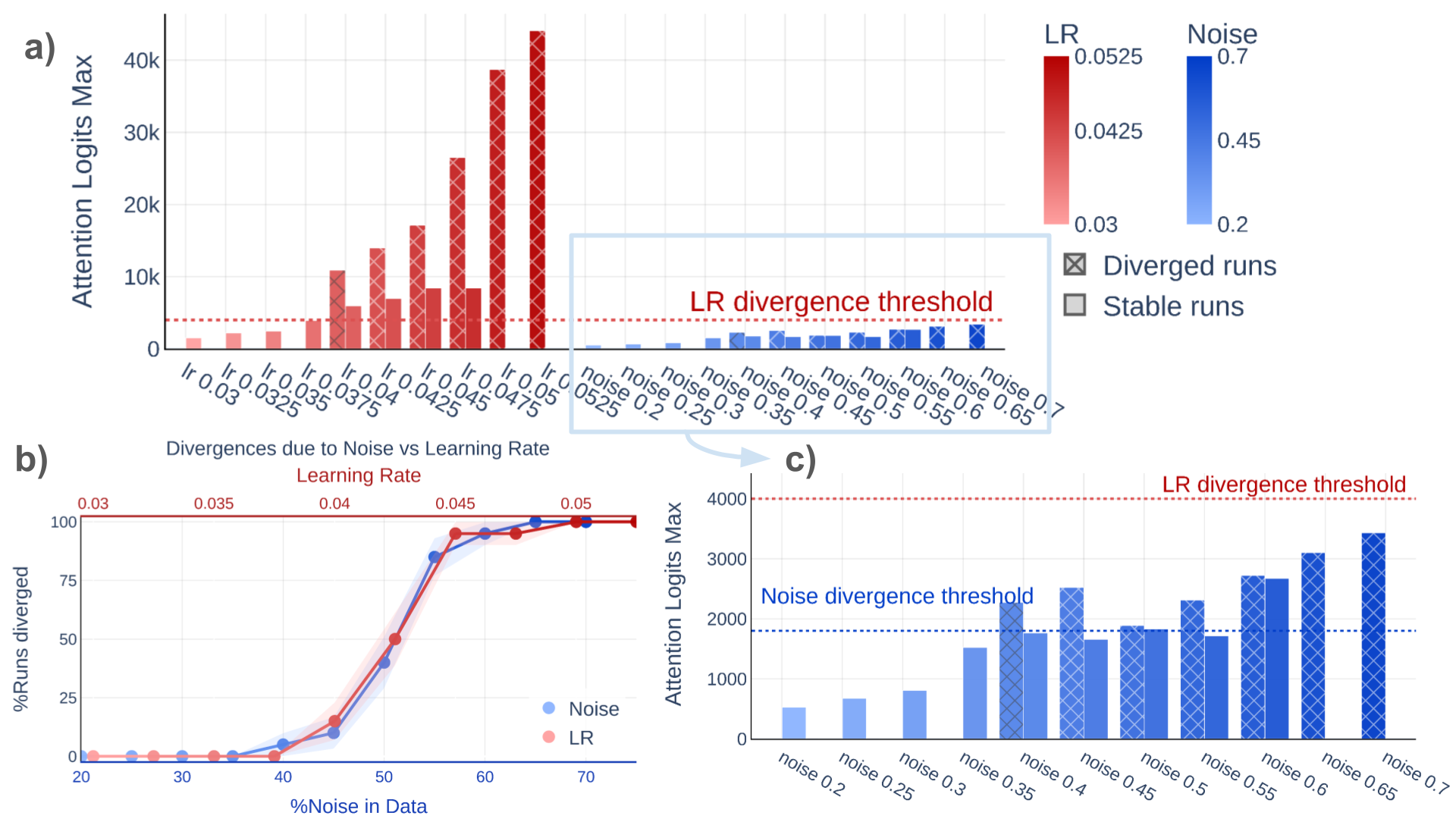}
    \caption{\textbf{Noisy data and overly high LR divergences exhibit different activation behavior.}
    We compare 540M dense models trained on clean data with \textcolor{lrred}{varying high-LRs (red)} versus models trained on a small LR with \textcolor{noiseblue}{varying noise ratios (blue)}.
    \textbf{(a)} Bars show the maximum attention logit at step 1000 averaged across seeds. Solid bars correspond to stable runs within the 20 seeds, while cross-hatched bars correspond to divergent runs within the 20 seeds. Missing bars indicate that no runs of that type occur for the configuration (i.e., lr=0.03 has no cross-hatched bars, meaning none of the 20 seeds for this configuration have diverged).
    \citeauthor{wortsman2023small} identify divergence threshold for high-LR runs, shown by the dotted red line at 4000.
    \textbf{(b)} For comparability, LR and noise ratio ranges are selected to match divergence probabilities across the two settings.
    \textbf{(c)} A zoom-in on noisy data runs from (a) reveals a different and lower divergence threshold for noisy data settings at approximately 1800 (blue dotted lines). As shown in \autoref{sec:apdx_more_sizes}, the same numerical thresholds, $\sim 4000$ for high-LR runs and $\sim 1800$ for noisy data runs, also holds across other model sizes.}
    \label{fig:act_analysis}
\end{figure*}

\begin{tcolorbox}
\textbf{Question:} Can we distinguish divergences caused by high LRs from those caused by noisy data?

\textbf{Answer: Yes, through examining activations.}
\end{tcolorbox}

\subsubsection{Examining maximum attention logits}
In a transformer's self-attention layer, attention logits are computed as
$z_{ij} = \frac{\langle q_i , k_j \rangle}{\sqrt{d_h}}$.
\citet{dehghani2023scaling} shows that an excessively large maximum attention logit causes the softmax to collapse toward one-hot weights, destabilizing training. Building on this, \citet{wortsman2023small} find that overly high LRs drive such logit growth and lead to loss divergence. They also show that divergence can be predicted early in training when the maximum attention logit exceeds a certain model-size-invariant threshold.

In \autoref{fig:act_analysis}, we analyze the maximum attention logits in runs that diverge due to noisy training data and compare them to runs that diverge due to high LRs using 540M dense models. See \autoref{sec:apdx_more_sizes} for other model sizes, including MoE models. 
For noisy data runs, we train 20 seeds per noise ratio, with $\alpha \in [30\%, 70\%]$, using a fixed LR of $1.85\times10^{-2}$. For high-LR runs, we train 20 seeds per LR, with $\text{LR} \in [3.25\times10^{-2}, 5.25\times10^{-2}]$, on the clean dataset $D_c$. As shown in \autoref{fig:act_analysis}(b), these ranges are chosen so that the probability of divergence under \textcolor{lrred}{high LRs (red)} closely matches that under \textcolor{noiseblue}{noisy data (blue)}, enabling direct comparisons of activation and parameter statistics across the two settings. 
For each configuration, we report the maximum attention logit at step 1000, computed across seeds and reported separately for diverged runs (cross-hatched bars) and stable runs (solid bars). Missing bars indicate that no runs of that type occur for the configuration (i.e., the lr=0.03 configuration does not have a cross-hatched bar, meaning none of the 20 seeds for this configuration have diverged). We chose to examine the activations at step 1000 because the earliest divergence occurs at this point. We make two observations:

\textbf{High LR divergences produce significantly larger maximum attention logits.} In \autoref{fig:act_analysis}(a), we first reproduce the findings of \citet{wortsman2023small}: across model sizes (see \autoref{sec:apdx_more_sizes}, including MoE architectures), high LR configurations lead to a non-zero probability of divergence once the maximum attention logit exceeds a threshold of $4000${\interfootnotelinepenalty=10000\footnote{In \citet{wortsman2023small}, the reported threshold is $10^{4}$ at step 2000. The lower threshold here is expected, as we measure activations earlier due to a shorter warm-up and fewer total training steps.}}.
Importantly, this threshold should be interpreted at the \emph{configuration level}: if the maximum attention logits exceed this threshold value, then the configuration is unstable, i.e., at least one of the 20 seeds diverges. Within an unstable configuration, stable runs may or may not exceed the threshold, but all diverged runs consistently do. Beyond reproducing this result, we find that high-LR runs have significantly larger maximum attention logits than noisy data runs; even at the highest noise ratios where noisy runs diverge 100\% of times, the noisy data runs' maximum attention logits never exceed the high LR divergence threshold of $4000$.

\begin{figure}[t]
  \centering
   \includegraphics[trim=0 0 800 50, clip, width=0.49\textwidth]{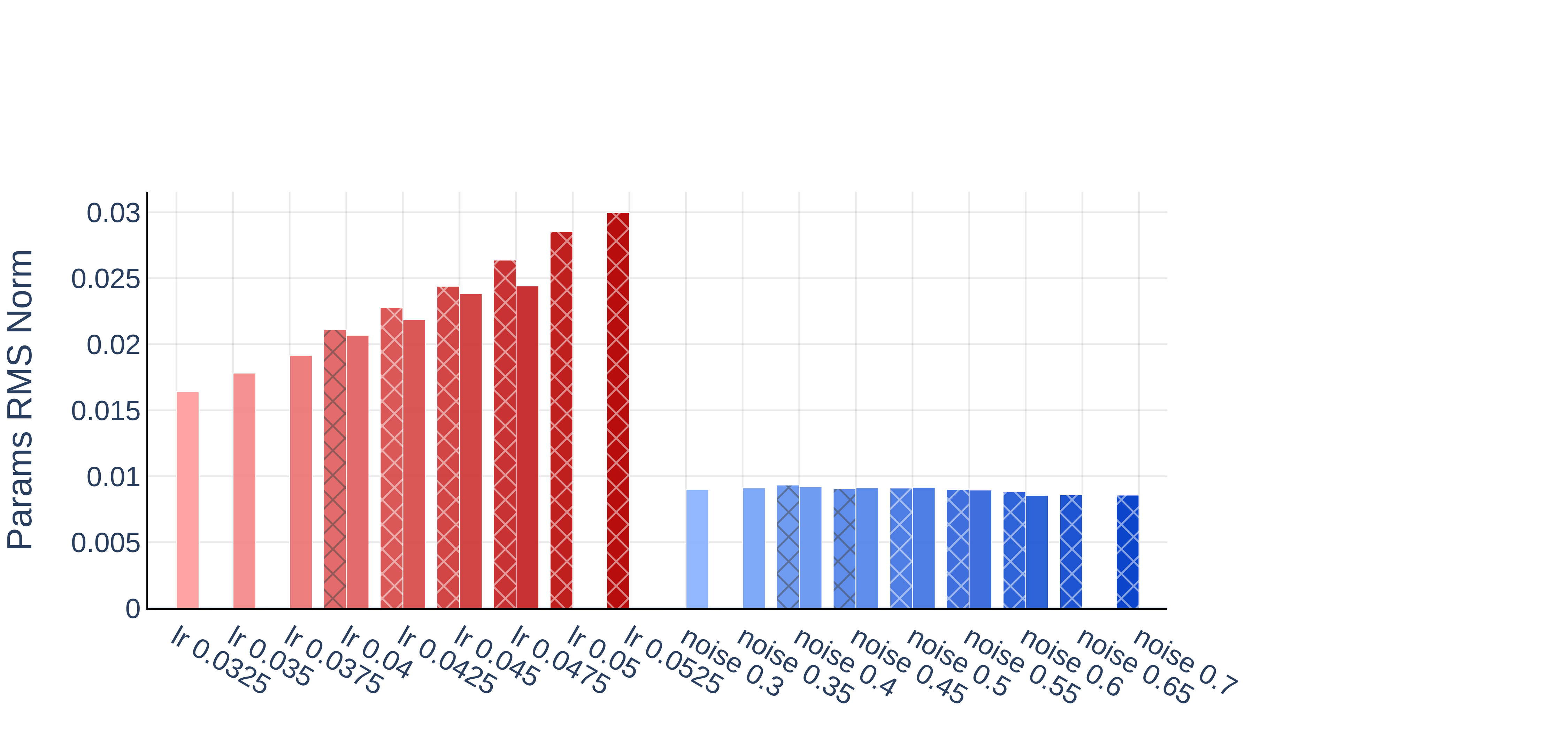}%
  \caption{Parameter RMS norms for 540M dense models trained on clean data with \textcolor{lrred}{varying high LRs (red)} versus with a fixed small LR on \textcolor{noiseblue}{varying noisy data (blue)}. Divergences induced by noisy data exhibit significantly smaller parameter norms than those induced by high LRs. Increasing the LR also leads to systematic growth in parameter norms, whereas increasing the noise ratio does not.}
  \label{fig:activation_absmean_and_norm}
\end{figure}

\textbf{Noisy data divergences occur at their own threshold, around 1800.} In \autoref{fig:act_analysis}(c), we see that within the noisy data setting, configurations whose maximum attention logits exceed $\sim 1800$ exhibit a non-zero probability of divergence. As shown in \autoref{sec:apdx_more_sizes}, the same numerical thresholds, $\sim 4000$ for high-LR runs and $\sim 1800$ for noisy data runs, also apply across different model sizes, including MoE models.

Together, these results show that divergence from noisy data is mechanistically distinct from divergence induced by high LRs. Maximum attention logits also provide a simple diagnostic for identifying the underlying cause of divergence. If a run diverges without exceeding the high-LR divergence threshold ($\sim 4000$) but exceeds the noisy data threshold ($\sim 1800$), then the divergence is likely due to noisy data rather than overly high LR.

\subsubsection{Examining parameter norms and other activations}
Beyond differences in maximum attention logits, we observe additional distinctions between divergences induced by high LRs and those caused by noisy data. As shown in Figure~\ref{fig:activation_absmean_and_norm}, even at comparable divergence rates, runs that diverge due to noisy data exhibit smaller parameter norms than those diverging due to high LRs. Moreover, increasing the LR leads to larger parameter norms, whereas increasing the noise ratio does not. See \autoref{sec:apdx_more_sizes} for analysis on other activation statistics.

\begin{figure}[t]
    \centering
    \includegraphics[trim=0 0 145 100,width=0.49\textwidth]{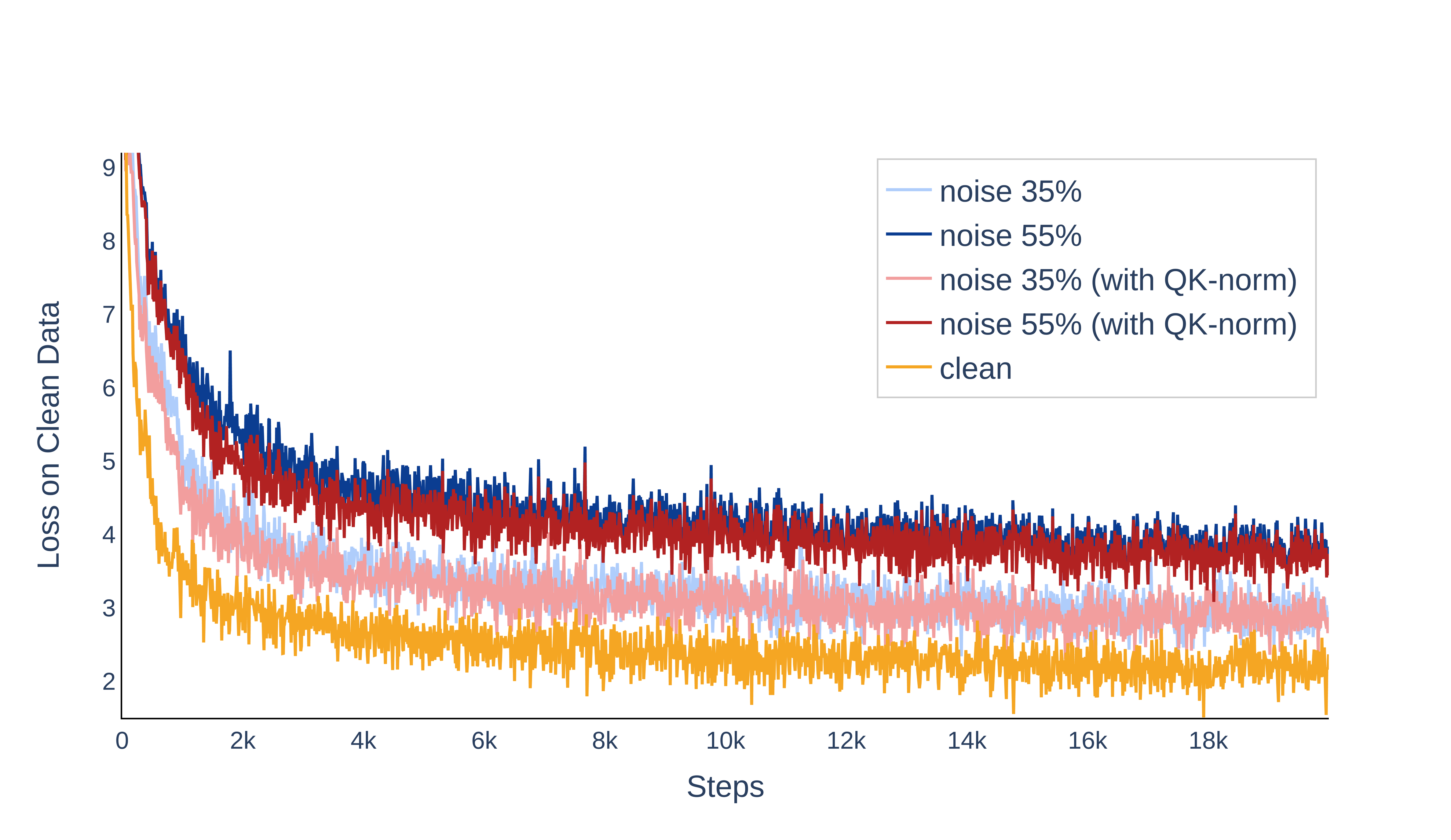}
  \caption{Average loss for stable runs of a 540M-parameter dense model, evaluated on clean tokens only. Results are averaged over seeds that do not diverge.}
  \label{fig:qk_loss}
\end{figure}

\subsection{Interventions for noisy data}
\begin{tcolorbox}
\textbf{Question:} If we diagnose a run as diverging due to noisy data, how can we stabilize training?

\textbf{Answer: Cleaning the data is most effective; when this is infeasible, apply QK-layernorm.}
\end{tcolorbox}

\begin{figure*}[t]
  \includegraphics[trim=0 0 445 0, width=0.3\textwidth]{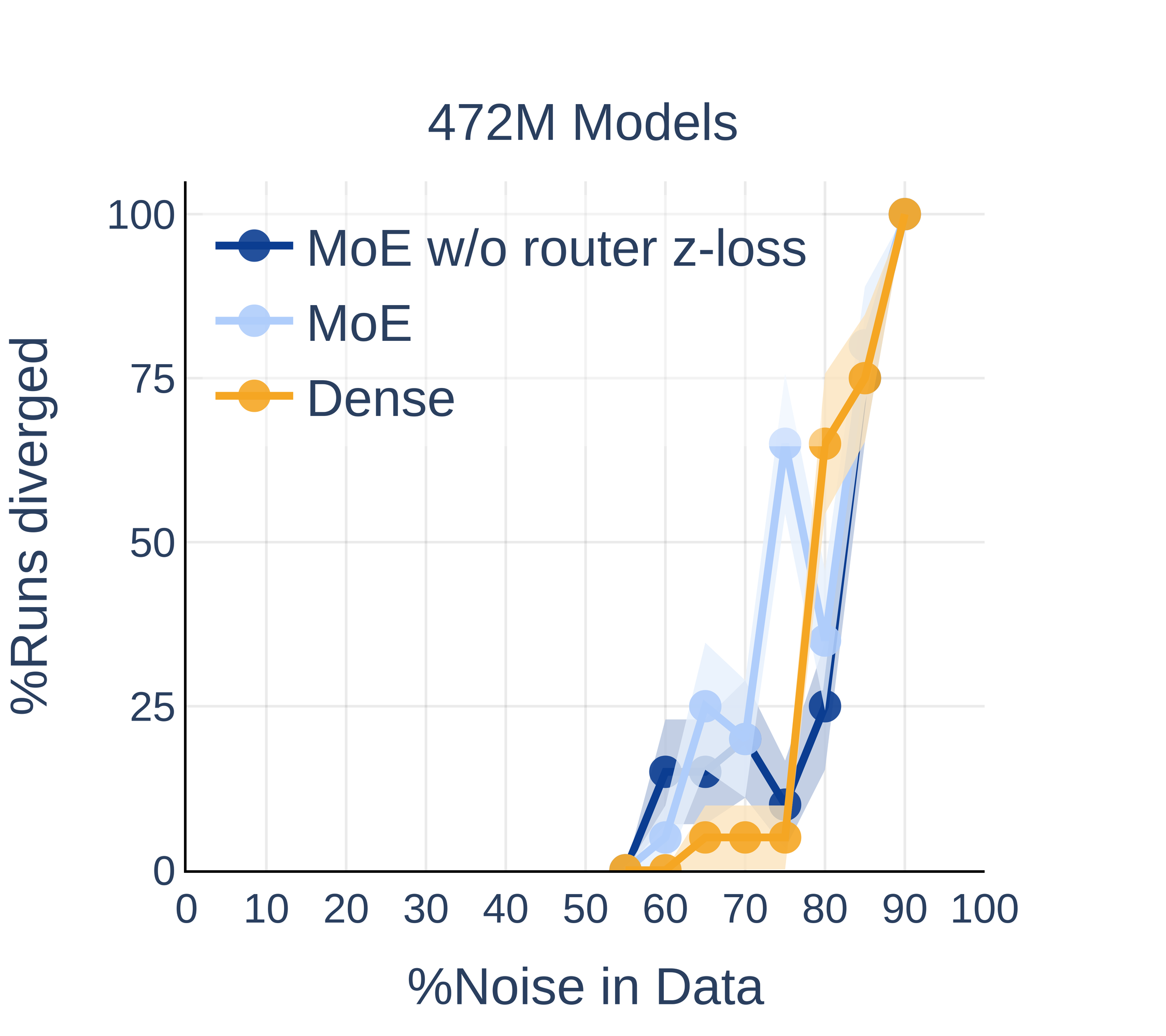}
  \includegraphics[trim=0 0 445 0,width=0.3
  \textwidth]{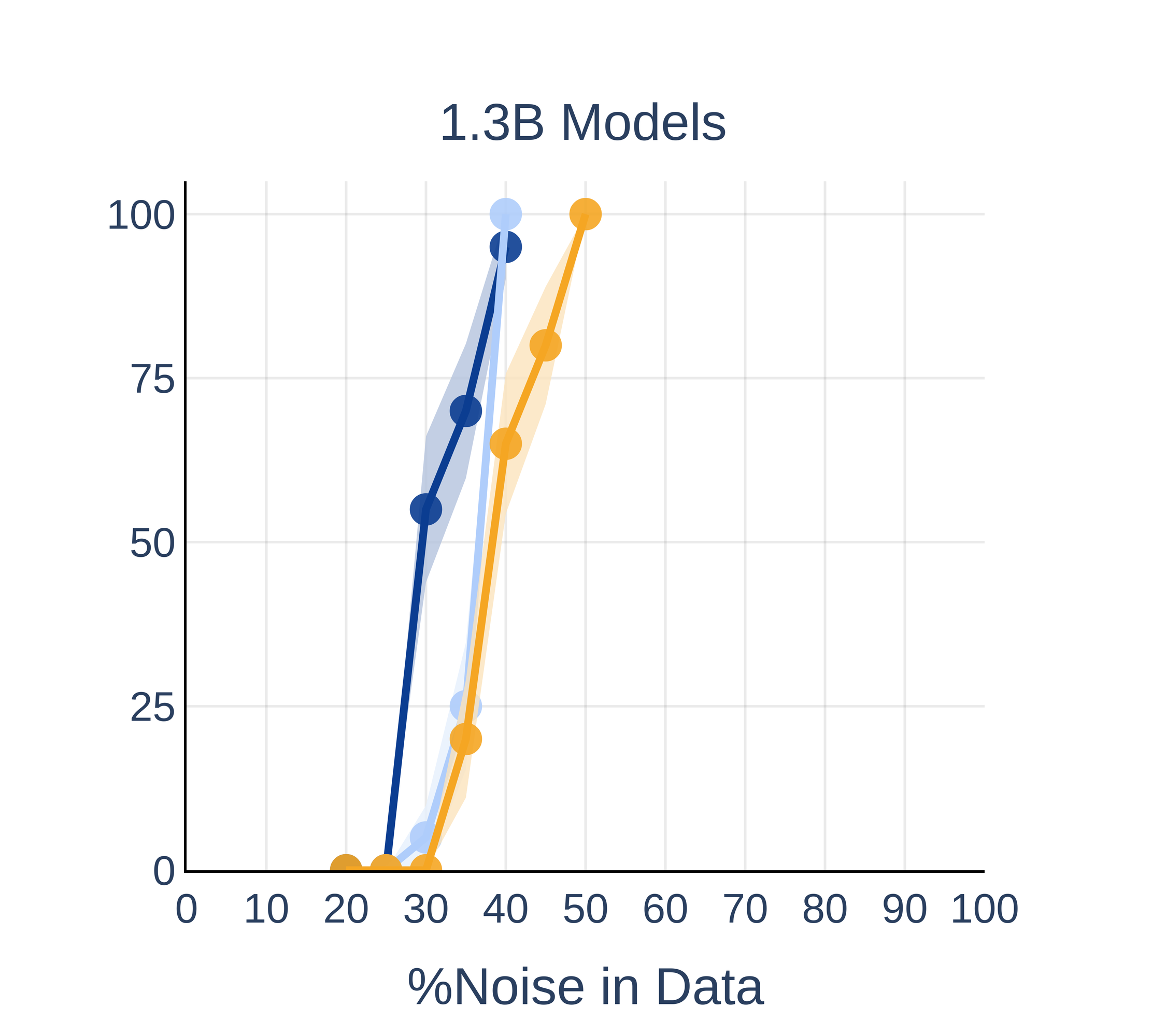}
\includegraphics[trim=0 0 445 0,width=0.3
  \textwidth]{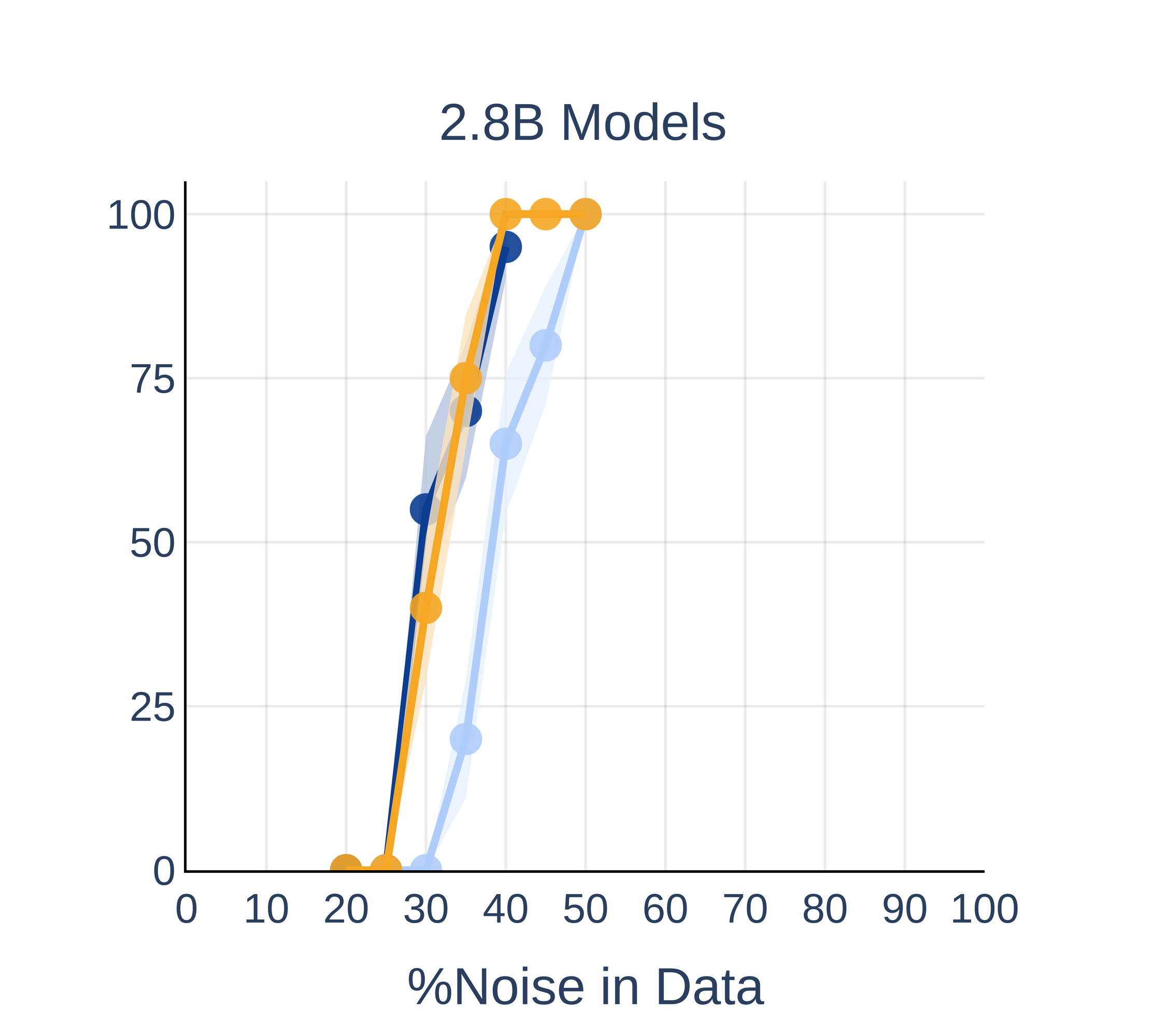}
  \caption{
  Divergence probability due to noisy data for dense and active-parameter–matched MoE models at three scales
  (472M, 1.3B, and 2.8B parameters). We report results for MoEs trained with
  and without router $z$-loss regularization. We observe that MoE models diverge at similar rates to their
  dense counterparts across all scales. 
  }
  \label{fig:moe_scaling}
\end{figure*}

We train 540M dense models with noise ratios ranging from $30\%$ to $70\%$. At $30\%$ noise, no runs diverge, while at $70\%$ noise all runs diverge across 20 seeds; intermediate noise levels yield intermediate divergence rates (see \autoref{fig:act_analysis}b). In \autoref{fig:qk_loss}, we report the average loss of only runs that remain stable across the 20 seeds.  We show the loss on clean tokens only, since this reflects the test-time distribution. For clarity, we only show results for $\alpha =0\%$ (clean data), $35\%$, and $55\%$ noise, although the trend is consistent across all noise ratios. We make two observations:

\textbf{Cleaning data is the best solution.}
Even when some runs remain stable under noisy training data settings, increasing the noise ratio consistently degrades performance on clean tokens. Training on clean data yields the best loss.

\textbf{When data cleaning is not possible, use QK-layernorm.}
Applying QK-layernorm substantially reduces maximum attention logits for all noisy runs (see \autoref{fig:qnorms_zoomed}) and eliminates all divergences across all noise ratios. Even at $70\%$ noise, a configuration that exhibited $100\%$ divergence, QK-layernorm fully stabilizes training across all seeds. We also observe that when trained on noisy data, QK-layernorm yields slightly lower training loss compared to stable runs without QK-layernorm.

\begin{figure*}[t]
    \centering
    \includegraphics[trim=0 0 45 0,width=0.98\textwidth]{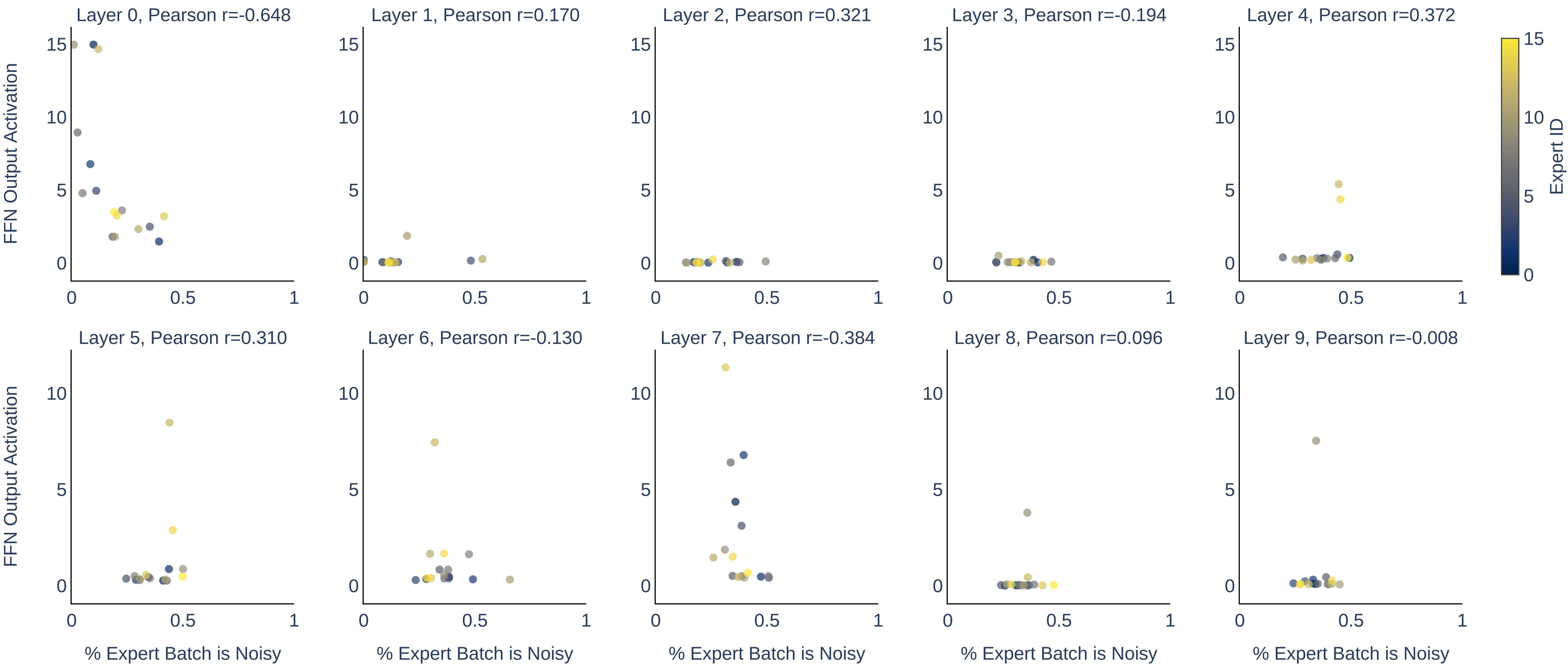}
  \caption{Each subplot corresponds to one of the ten layers in the 1.3B MoE model. The x-axis shows the fraction of noisy tokens
  in the expert batch, and the y-axis shows the absolute mean of the FFN output activation.
  The average Pearson correlation across layers is $-0.009$, indicating little 
  correlation between the two.}
  \label{fig:router0_main1}
\end{figure*}

\subsection{Dense vs. MoE models}
\label{sec:moes}

Prior work shows that MoE architectures can be less stable to train than dense models, often requiring additional stabilization techniques \citep{zoph2022st,fedus2022switch}. With noisy data, a natural concern is whether expert specialization amplifies this instability. In particular, one can question if some experts may disproportionately receive noisy tokens, effectively behaving like a more unstable sub-network trained on highly noisy data.

\begin{tcolorbox}
\textbf{Question:} Do MoEs diverge more easily than dense models under noisy data?

\textbf{Answer: No.}
\end{tcolorbox}

We compare dense models with active-parameter–matched MoE models at three scales
(472M, 1.3B, and 2.8B parameters). We evaluate MoEs both with and without
router $z$-loss regularization \citep{zoph2022st}, which is known to improve the router's training stability.

As shown in \autoref{fig:moe_scaling}, MoE models have divergence rates similar to those of dense models across all three scales. This indicates that MoEs are not more sensitive than dense models to noisy data.

To understand this behavior, we analyze how noisy tokens are routed to experts in 1.3B active-parameter MoEs trained with 35\% noise. We visualize token routing for a representative run that diverged\footnote{See leftmost panel in \autoref{fig:moe_divergence_exmaples} for the corresponding loss curve. We also refer readers to \autoref{sec:router_analysis} for analysis on additional diverged and stable runs.}. 
Each subplot in  \autoref{fig:router0_main1} corresponds to one MoE layer, with each point representing an expert. The x-axis reports the fraction of noisy tokens in the expert’s batch, and the y-axis reports the absolute mean of the expert's FFN output activation. We examine the magnitude of the expert's output activation because growing activations usually correlate with training instability. We observe that, although experts receive varying proportions of noisy tokens, this fraction is uncorrelated with activation magnitude, with an average Pearson correlation of $-0.009$ across layers. This reinforces the conclusion that the MoE routing mechanism does not introduce additional sensitivity to noisy data.

\section{Conclusion}
We present a large-scale, controlled study of how uniform random noise causes LLM pretraining loss divergence. We systematically inject varying types and amounts of synthetic noise into clean datasets and evaluate training dynamics across model sizes ranging from 480M to 5.2B parameters.

Our findings show that noise can indeed cause loss divergence, even in small-scale regimes that are typically stable to train. We find that the divergence rate increases with noise ratio and model size, and model depth plays a larger role than width in increased sensitivity to noise. We further demonstrate that divergences driven by noisy data exhibit activation patterns distinct from those caused by overly high LRs, enabling practitioners to differentiate these two major failure modes. Finally, we observe that dense and active-parameter-matched MoE models behave similarly under noisy data, suggesting that noisy data sensitivity is not substantially altered by sparse architectural design. Together, these results provide an understanding of how noisy data impacts LLM pretraining, and offer practical guidance for improving training robustness, data curation, and model design.

\section*{Acknowledgments}
Authors thank Abhishek Kadian and Punit Singh Koura for helpful pointers regarding data loader implementations. The authors are grateful for Meng Zhang, Hanwen Zha, Vinay Rao, and Haroun Habeeb for insightful research discussions. QZ would also like to thank the interns in the legacy GenAI team for brainstorming together, exchanging ideas, and providing feedback, including Wannan (Winnie) Yang, Rachit Bansal, Julian Coda-Forno, Sai Surya Duvvuri, Sriyash Poddar, Arushi Rai, Rishabh Tiwari, and Devvrit Khatri. QZ thanks Chris Lu for helpful discussions on the project. 

\bibliographystyle{assets/plainnat}
\bibliography{paper}

\newpage
\appendix
\onecolumn
\section{Model architecture details}
\autoref{tab:model_configs} reports the default model configuration that is shared across all experiments and model sizes.

\begin{table}[h!]
\centering
\renewcommand{\arraystretch}{1.3}
\begin{tabular}{l | c}
\toprule
\textbf{FFN Dim} &  Model Dim $\times$ 4\\
\textbf{Sequence Length} &  8192\\
\textbf{Head Dim} &  128\\
\textbf{Key/Value Heads} & 8 \\
\textbf{Activation Function} & {SwiGLU} \\
\textbf{Tokenizer Vocabulary Size} & {200,000} \\
\textbf{Positional Embeddings} & {{RoPE} ($\theta = 500,000$)} \\
\bottomrule
\end{tabular}
\caption{Summary of architecture hyperparameters shared across all models trained in this paper.}
\label{tab:model_configs}
\end{table}

\autoref{tab:arch_more_details} provides the additional architectural variations used in all experiments. For MoE models, the `` $\#$ Parameters'' column reports the number of active parameters.

\begin{table}[h!]
\centering
\begin{tabular}{lcccc}
\toprule
\textbf{} & \textbf{\# Parameters} & \textbf{Model Dim} & \textbf{\# Layers} & \textbf{\# Query Heads} \\
\midrule
\multirow{5}{*}{\textbf{Scaling Depth}}
 & 1.1B & 2048 & 5  & 16 \\
 & 1.3B & 2048 & 10 & 16 \\
 & 1.5B & 2048 & 15 & 16 \\
 & 2.0B & 2048 & 25 & 16 \\
 & 2.5B & 2048 & 35 & 16 \\
\midrule

\multirow{4}{*}{\textbf{Scaling Width}}
 & 540M & 1024 & 10 & 8  \\
 & 1.3B & 2048 & 10 & 16 \\
 & 2.2B & 3072 & 10 & 24 \\
 & 3.4B & 4096 & 10 & 32 \\
\midrule

\multirow{4}{*}{\textbf{Scaling Both}}
 & 480M & 1024 & 5  & 8  \\
 & 1.3B & 2048 & 10 & 16 \\
 & 2.8B & 3072 & 15 & 24 \\
 & 5.2B & 4096 & 20 & 32 \\
\bottomrule
\end{tabular}
\caption{Model configurations for experiments scaling depth, width, and scaling both depth and width. }
\label{tab:arch_more_details}
\end{table}

\subsection{MoE architecture details}
\label{sec:apdx_moe_arch}
For Mixture of Experts \citep[MoE;][]{shazeer2017outrageously,fedus2022switch} experiments, we use dropless MoEs\citep{gale2023megablocks, liu2024deepseek}  with 16 feed-forward network (FFN) experts and token-choice top-2 routing. 
Given a token representation $x_t$, the router produces logits $r_{t,e}$ over experts $e \in \{1,\dots,16\}$. 
We select the top-2 experts
\[
S_t = \operatorname{TopK}(\operatorname{Router}(x_t), k=2),
\]
We compute the mixture weights $p_{t,e}$ by applying a softmax  over the selected experts $S_t$:
\[
p_{t,e} = \operatorname{softmax}(r_{t,e}) \quad \text{for } e \in S_t
\]

The final MoE layer output is the top-2 experts weighted by the mixture weights
\[
y_t = \sum_{e \in S_t} p_{t,e} \operatorname{FFN}_e(x_t).
\]

To maintain a balanced load across experts, we adopt loss-free balancing \citep{wang2024auxiliary, liu2024deepseek} with an expert-bias coefficient of $1\mathrm{e}{-3}$. For some experiments, we additionally apply the router z-loss \citep{zoph2022st}, using an auxiliary loss coefficient of $1\mathrm{e}{-3}$. 

For fair comparison with dense models, our top-$2$ MoE models are active-parameter matched by scaling the FFN  dimension by $0.5$. All remaining architectural components follow the same configuration as the dense models.

\newpage
\section{Additional model parameter and activation analysis}
\label{sec:apdx_more_sizes}
We provide additional analyses of model parameters and activations extending \autoref{sec:lr_vs_noise}. Specifically, we study three dense models with parameter counts ranging from 540M to 2.8B, as well as an MoE model with 1.3B active parameters (3.7B total parameters). We find that the conclusions of \autoref{sec:lr_vs_noise} hold consistently across model sizes and for both dense and MoE architectures.

\subsection{540M dense model}

\begin{figure}[H]
    \includegraphics[trim=0 0 1100 0,width=0.49
  \textwidth]{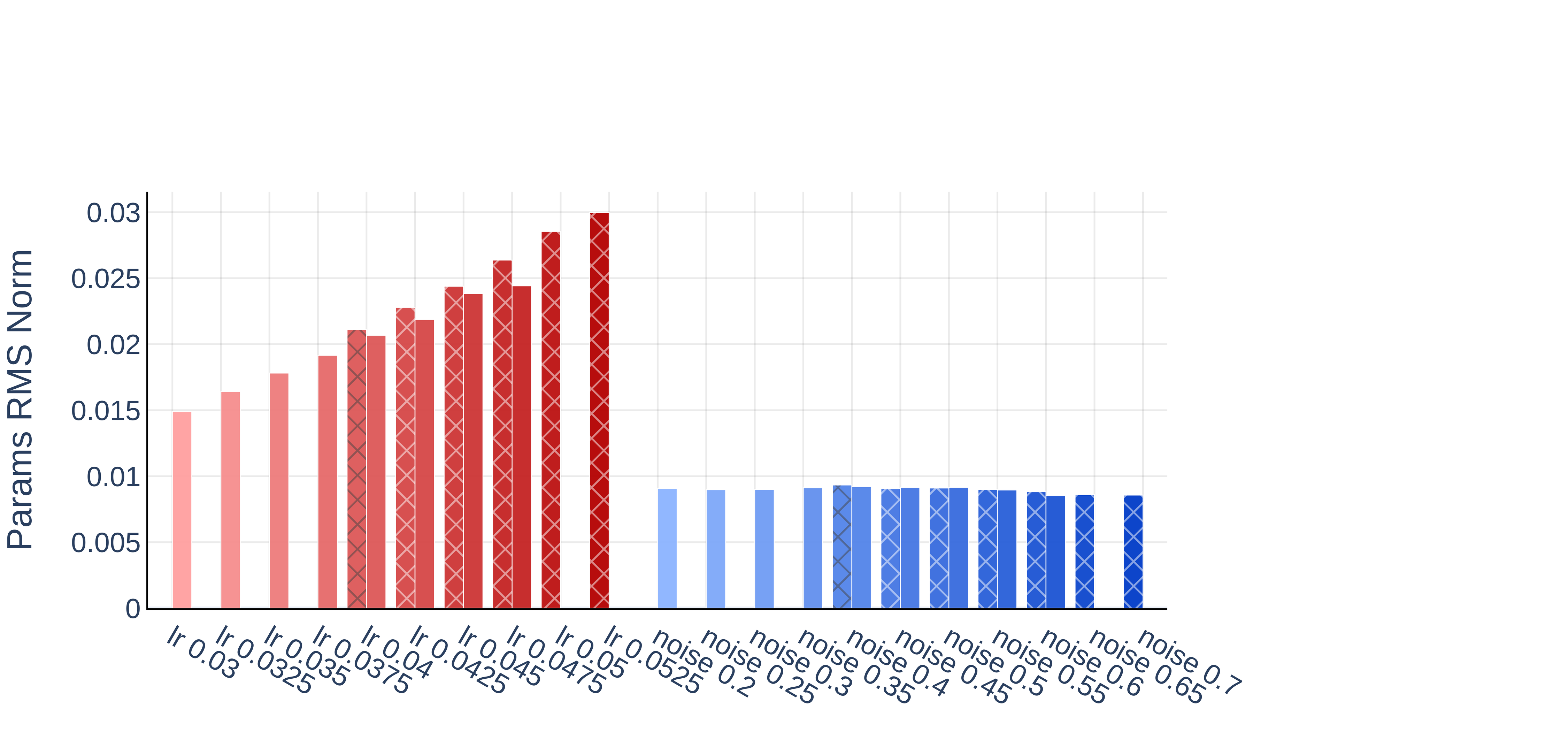}
    \includegraphics[trim=0 0 1100 0,width=0.49
  \textwidth]{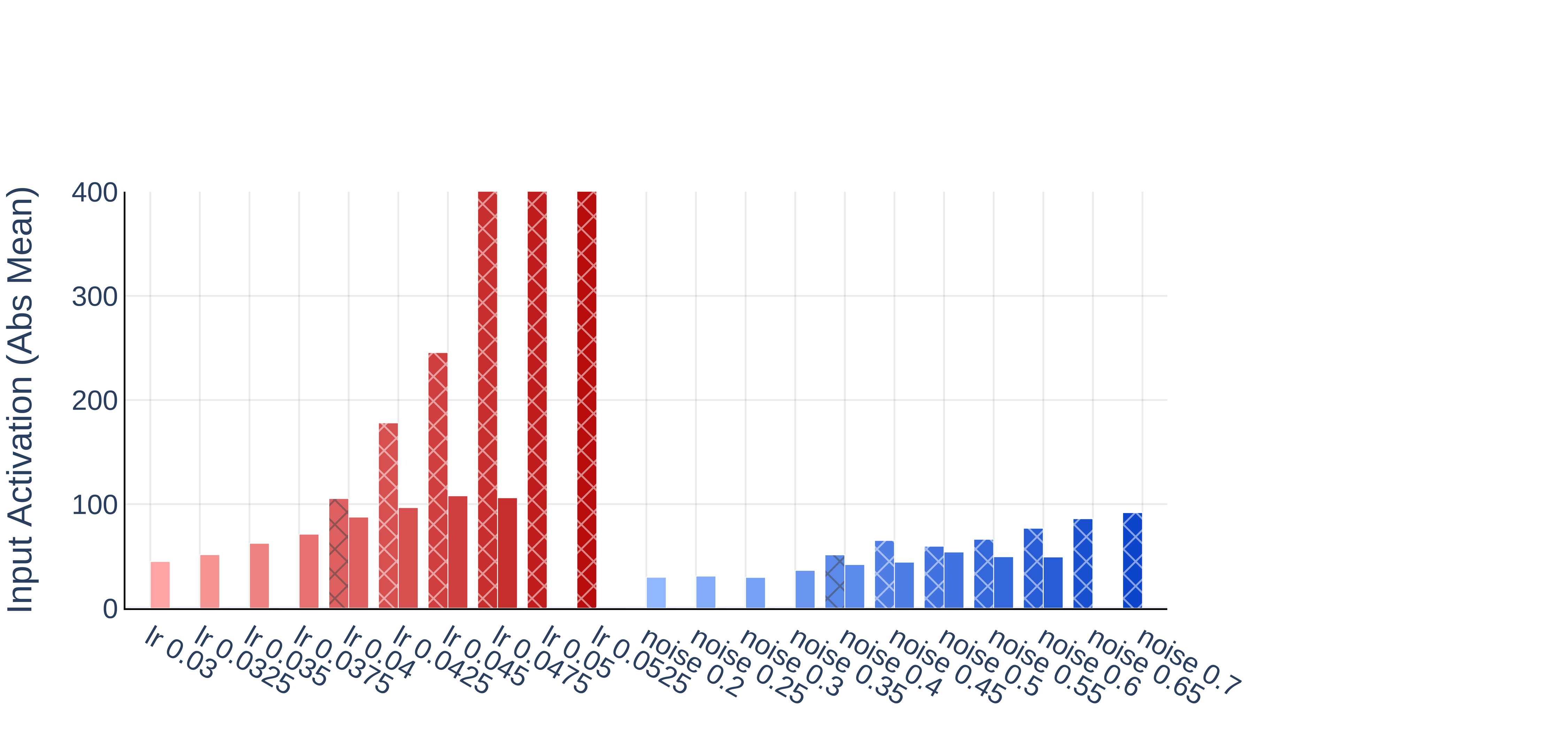}
  \caption{540M dense model run statistics. \textbf{Left:} Parameter RMS norm. \textbf{Right:} Absolute mean of activation input to transformer layers.}
  \label{fig:act_analysis0}
\end{figure}

\begin{figure}[H]
\centering
\includegraphics[trim=0 0 1100 0, width=0.49\textwidth]{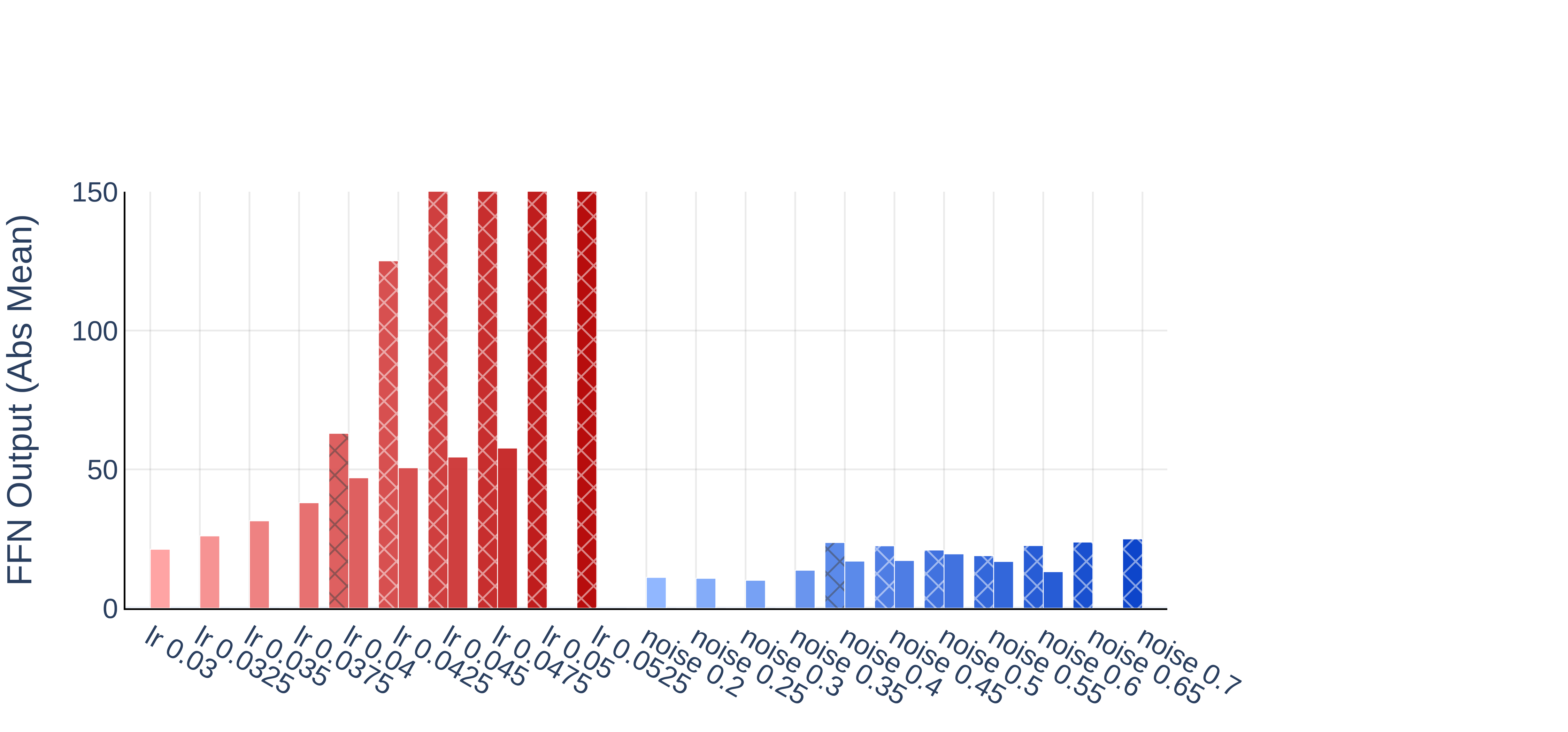}
\hfill
\includegraphics[trim=0 0 1100 0, width=0.49\textwidth]{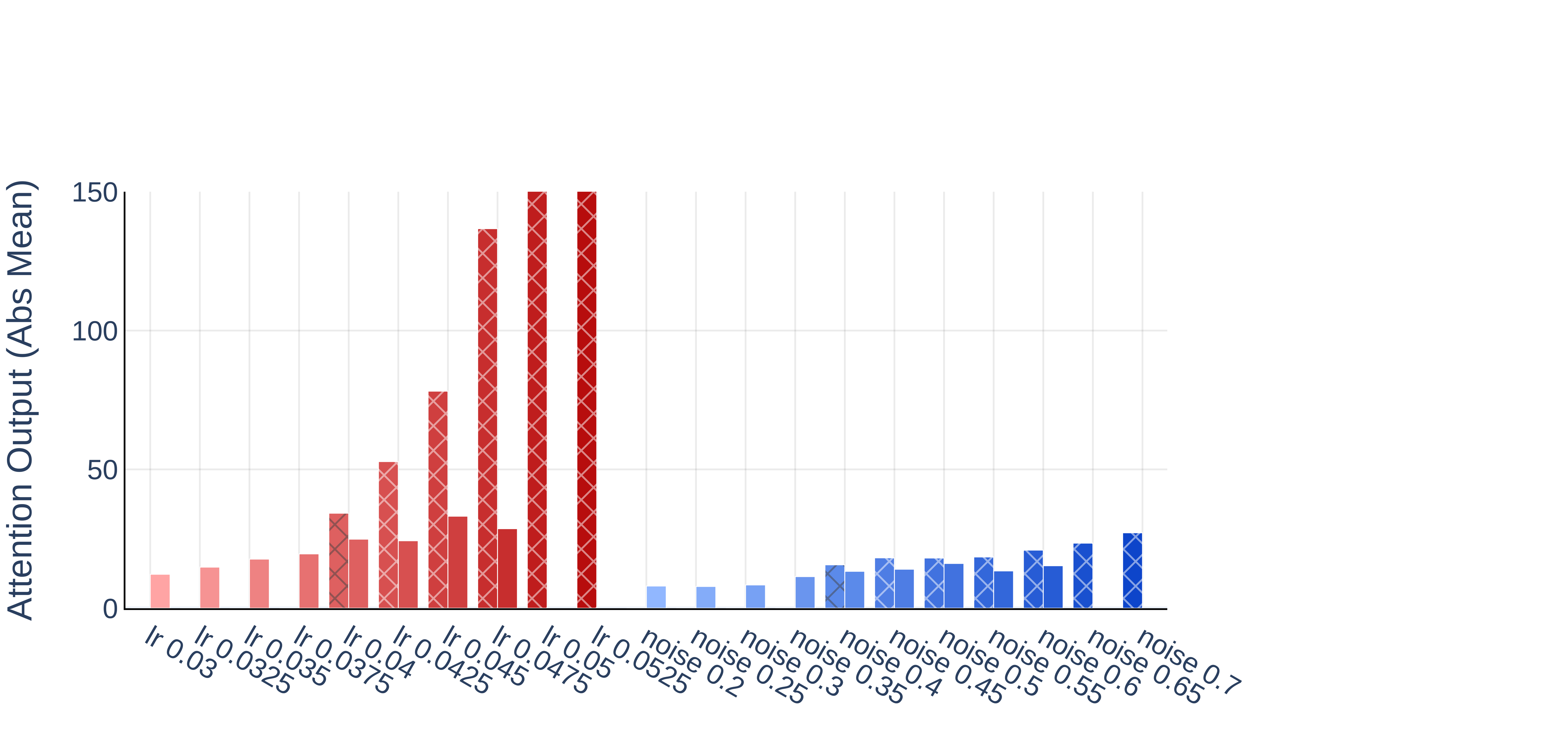}

\includegraphics[trim=0 0 1100 0, width=0.49\textwidth]{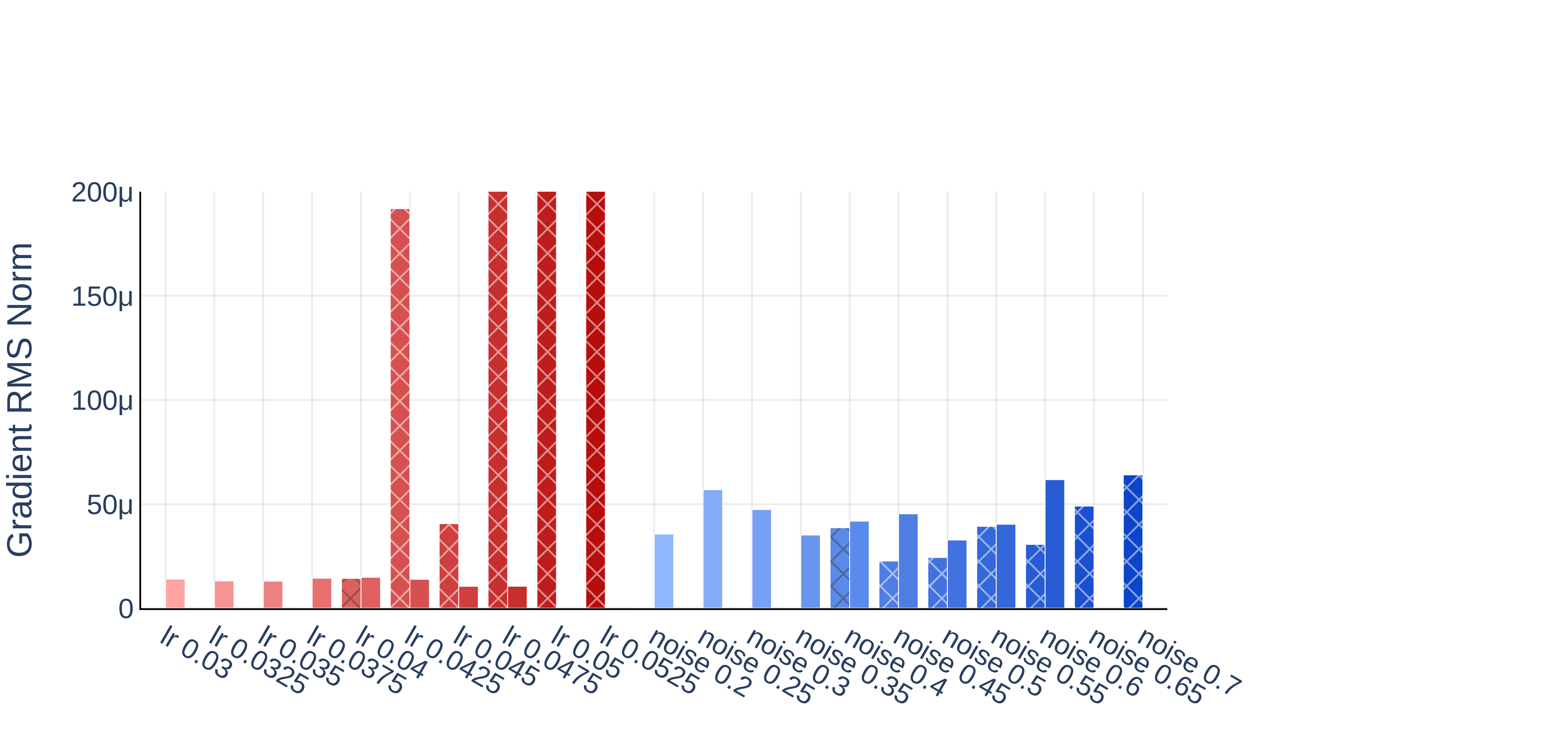}

\caption{540M dense model run statistics. \textbf{Top left}: Absolute mean of pre-residual FFN output. \textbf{Top right}: Absolute mean of pre-residual attention output. \textbf{Bottom}: Gradient RMS norm.}
\label{fig:540m_more_acts}
\end{figure}

\subsection{1.5B dense model}

\begin{figure}[H]
    \includegraphics[trim=0 0 45 0,width=0.95
  \textwidth]{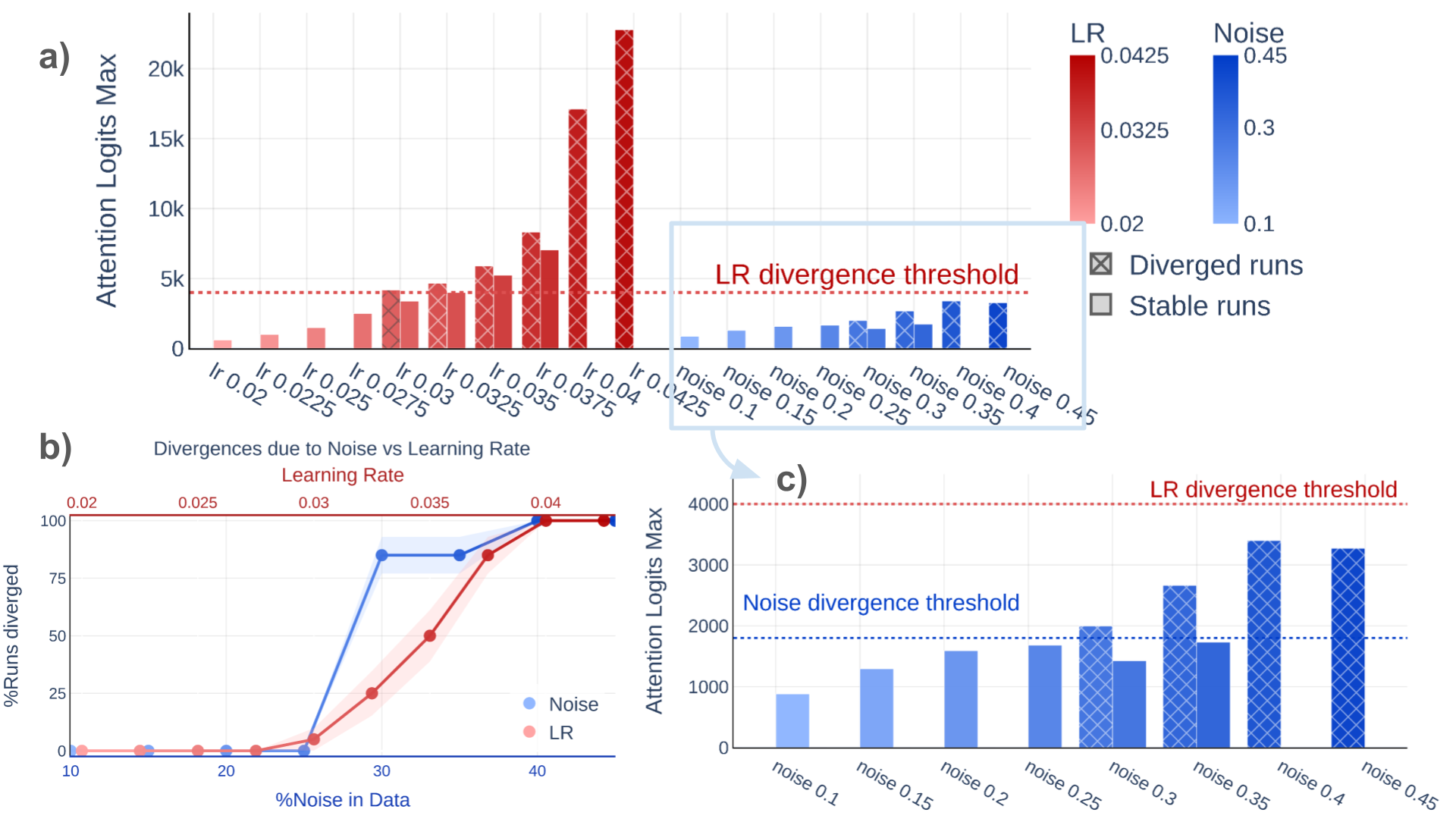}
  \caption{Examining the maximum attention logits for 1.5B Dense models .\textbf{(a)}
    \citeauthor{wortsman2023small} identify a model-size-invariant divergence threshold for high LR runs, shown by the dotted red line at 4000.
    \textbf{(b)} For comparability, LR and noise ratio ranges are selected to match divergence probabilities across the two settings.
    \textbf{(c)} A zoom-in on noisy data runs from (a) reveals a different and lower divergence threshold for noisy data settings at approximately 1800 (see the blue dotted lines). This noisy data divergence threshold also holds across different model sizes and MoE architectures.}
  \label{fig:act_analysis2}
\end{figure}

\begin{figure}[H]
    \includegraphics[trim=0 0 1100 0,width=0.49
  \textwidth]{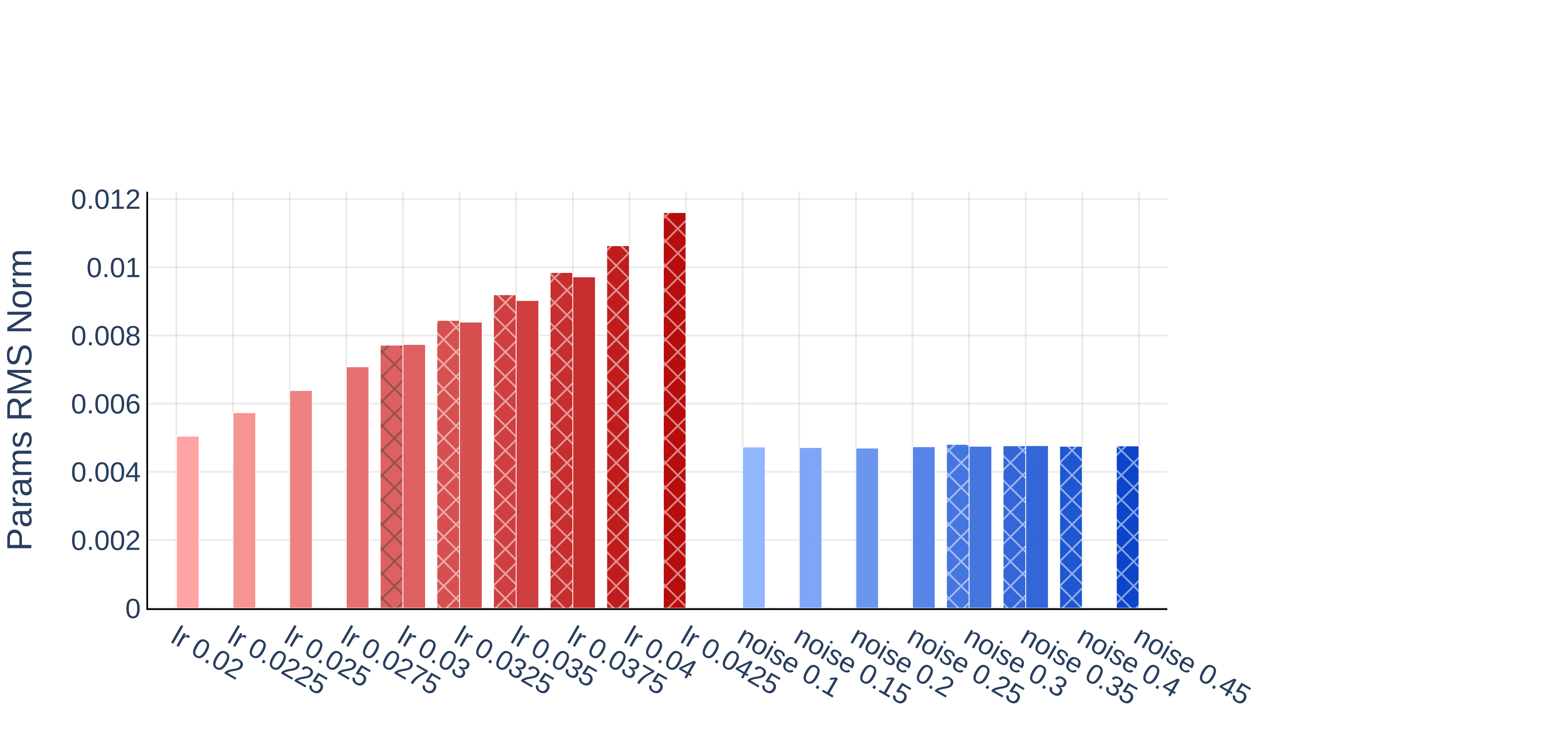}
    \includegraphics[trim=0 0 1100 0,width=0.49
  \textwidth]{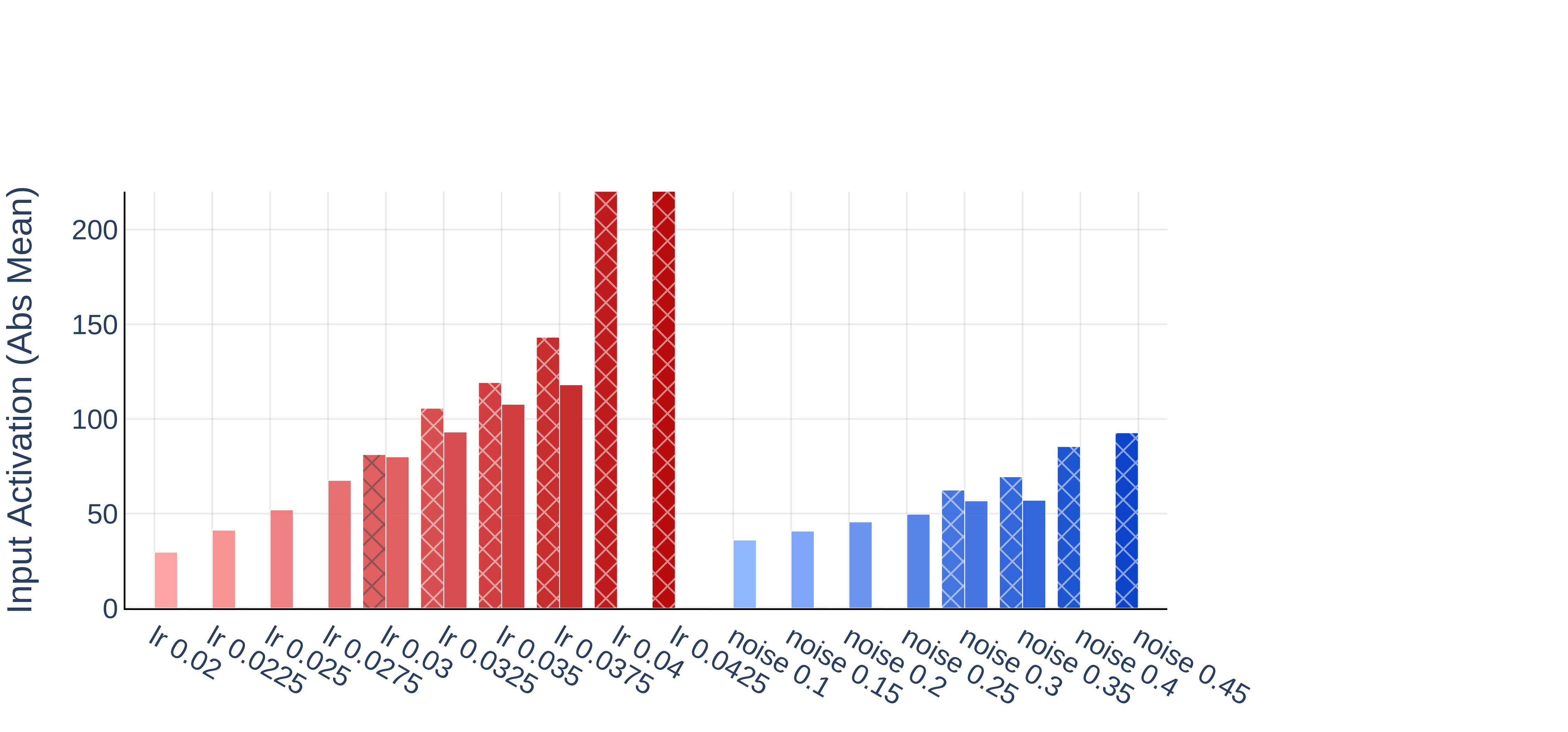}
  \caption{1.5B dense model run statistics. \textbf{Left:} Parameter RMS norm. \textbf{Right:} Absolute mean of activation input to transformer layers.}
  \label{fig:act_analysis3}
\end{figure}

\begin{figure}[H]
    \centering
    \includegraphics[trim=0 0 1100 0,width=0.49
  \textwidth]{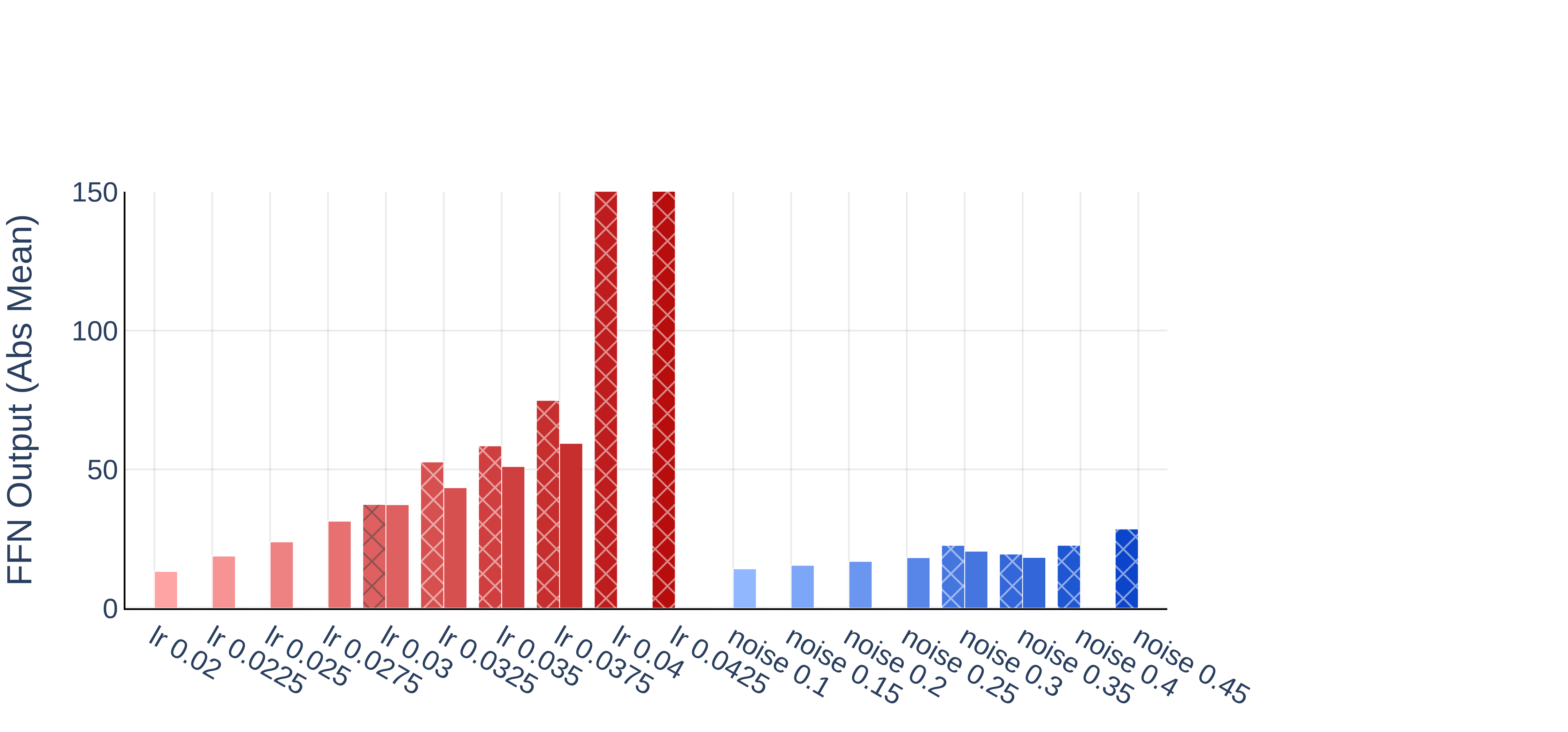}
    \includegraphics[trim=0 0 1100 0,width=0.49
  \textwidth]{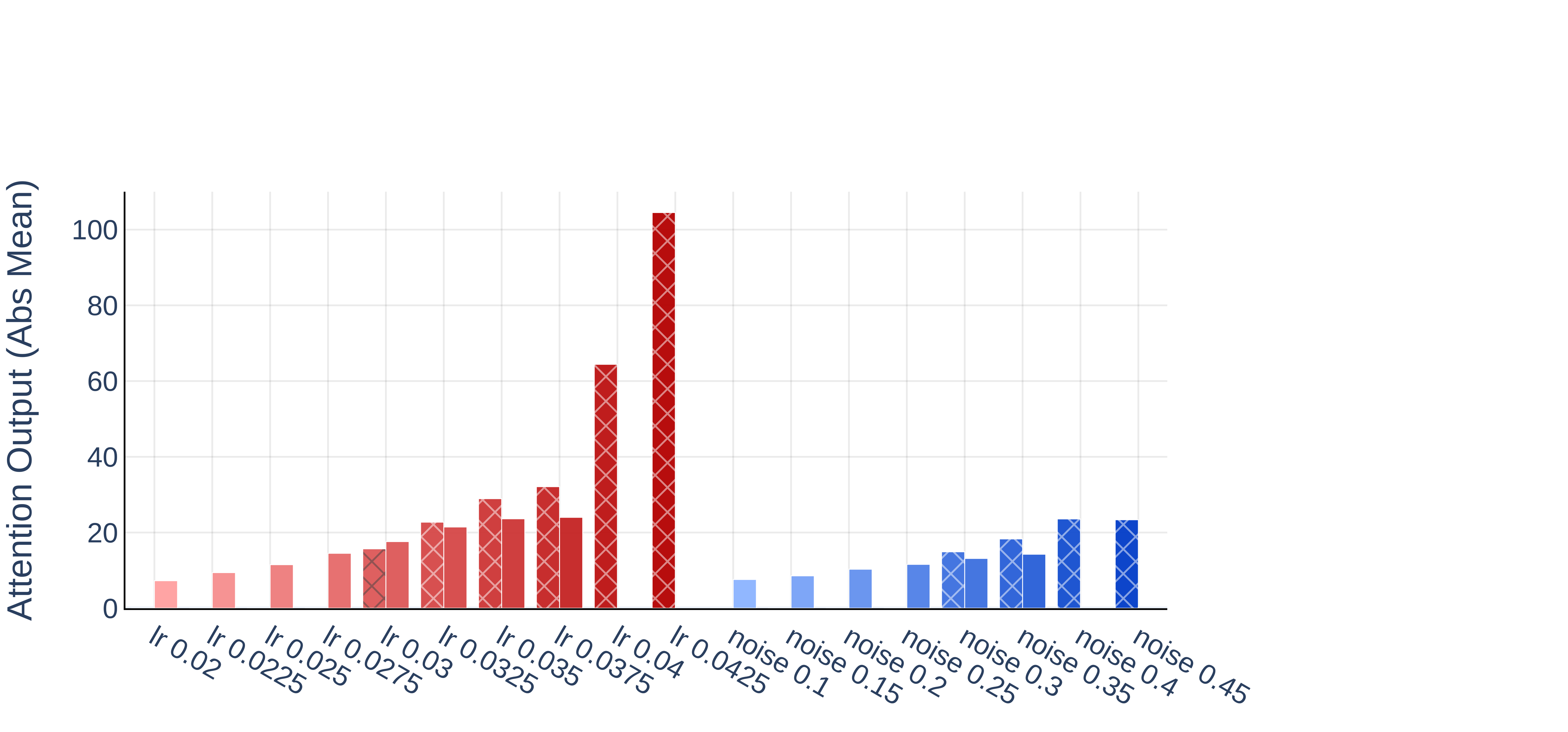}
      \includegraphics[trim=0 0 1100 0,width=0.49
  \textwidth]{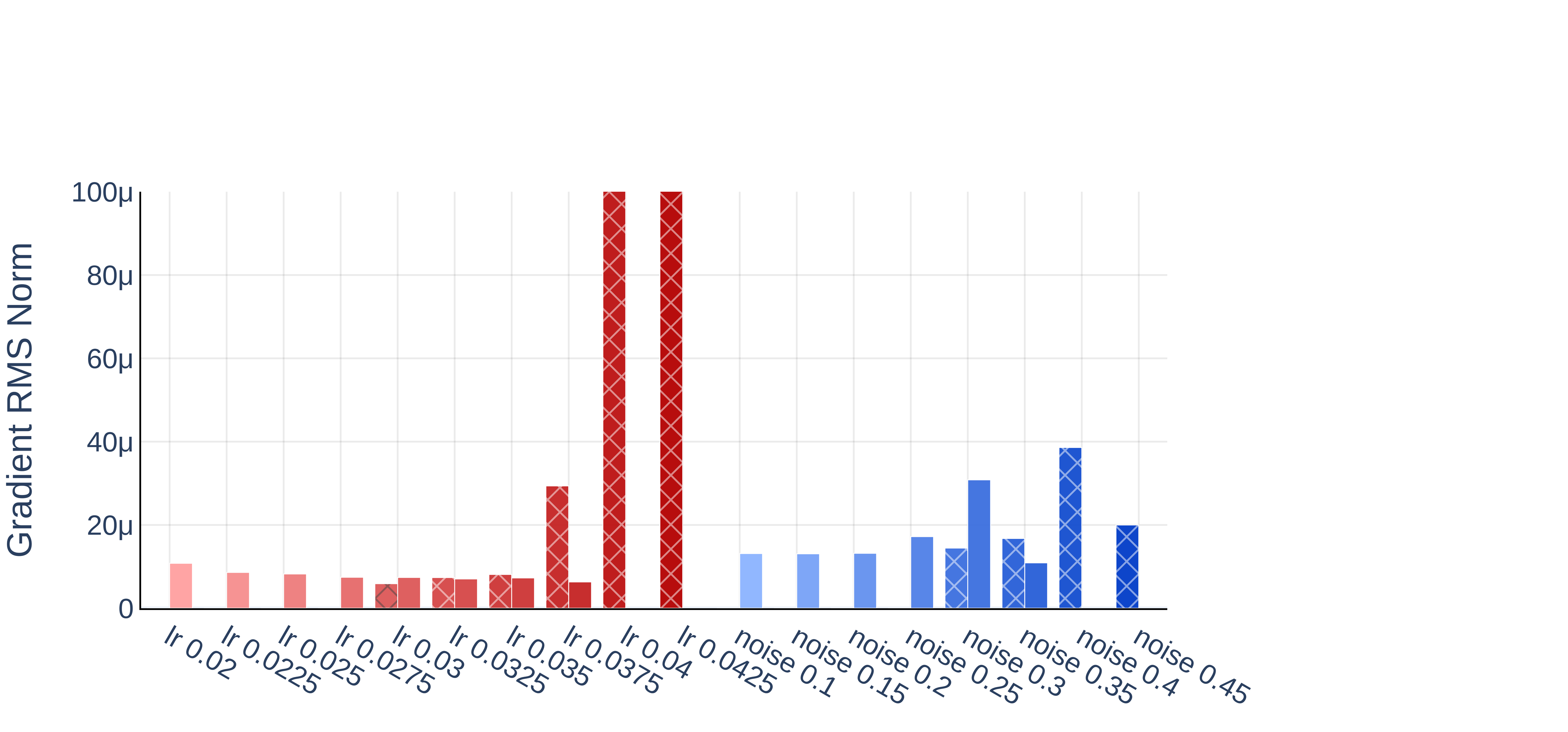}
  \caption{1.5B dense model run statistics. \textbf{Top left}: Absolute mean of pre-residual FFN output. \textbf{Top right}: Absolute mean of pre-residual attention output. \textbf{Bottom}: grad RMS norm.}
  \label{fig:act_analysis4}
\end{figure}

\subsection{2.8B dense Model}
\begin{figure}[H]
    \includegraphics[trim=0 0 45 0,width=0.95
  \textwidth]{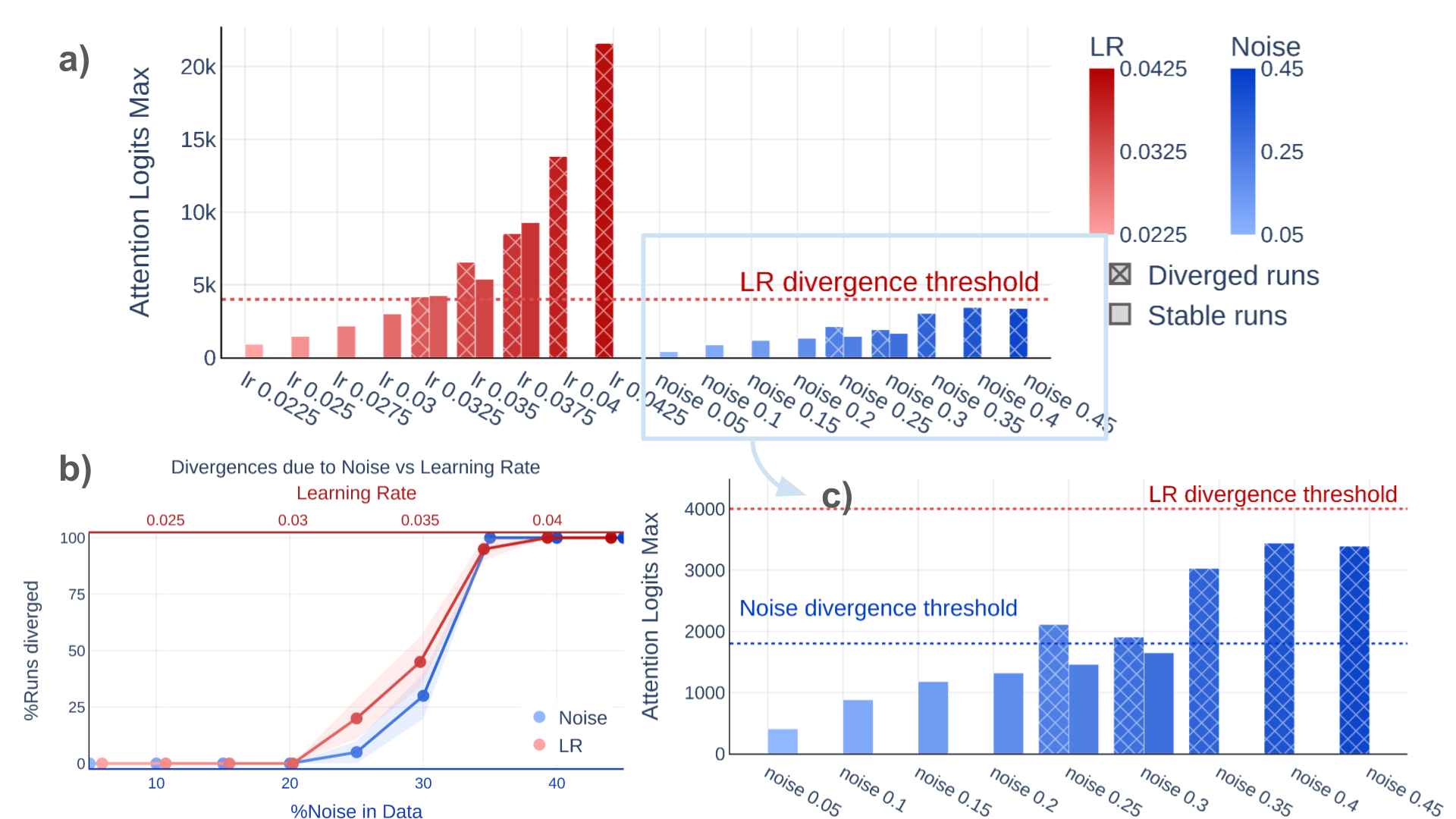}
  \caption{
  Examining the maximum attention logits for 2.8B dense models .\textbf{(a)}
    \citeauthor{wortsman2023small} identify a model-size-invariant divergence threshold for high LR runs, shown by the dotted red line at 4000.
    \textbf{(b)} For comparability, LR and noise ratio ranges are selected to match divergence probabilities across the two settings.
    \textbf{(c)} A zoom-in on noisy data runs from (a) reveals a different and lower divergence threshold for noisy data settings at approximately 1800 (see the blue dotted lines). This noisy data divergence threshold also holds across different model sizes and MoE architectures.}
  \label{fig:act_analysi4}
\end{figure}

\begin{figure}[H]
    \includegraphics[trim=0 0 1100 0,width=0.49\textwidth]{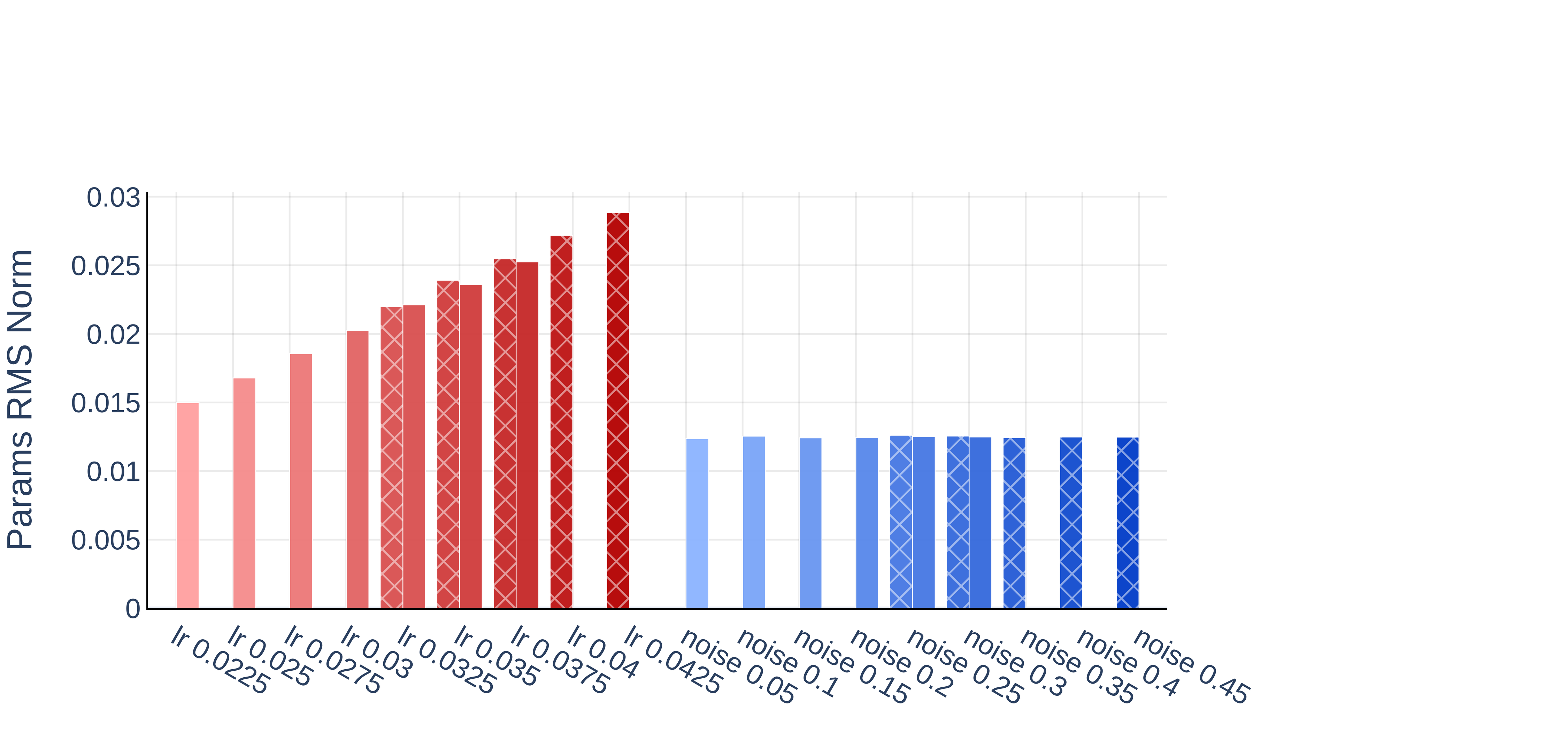} 
    \includegraphics[trim=0 0 1100 0,width=0.49\textwidth]{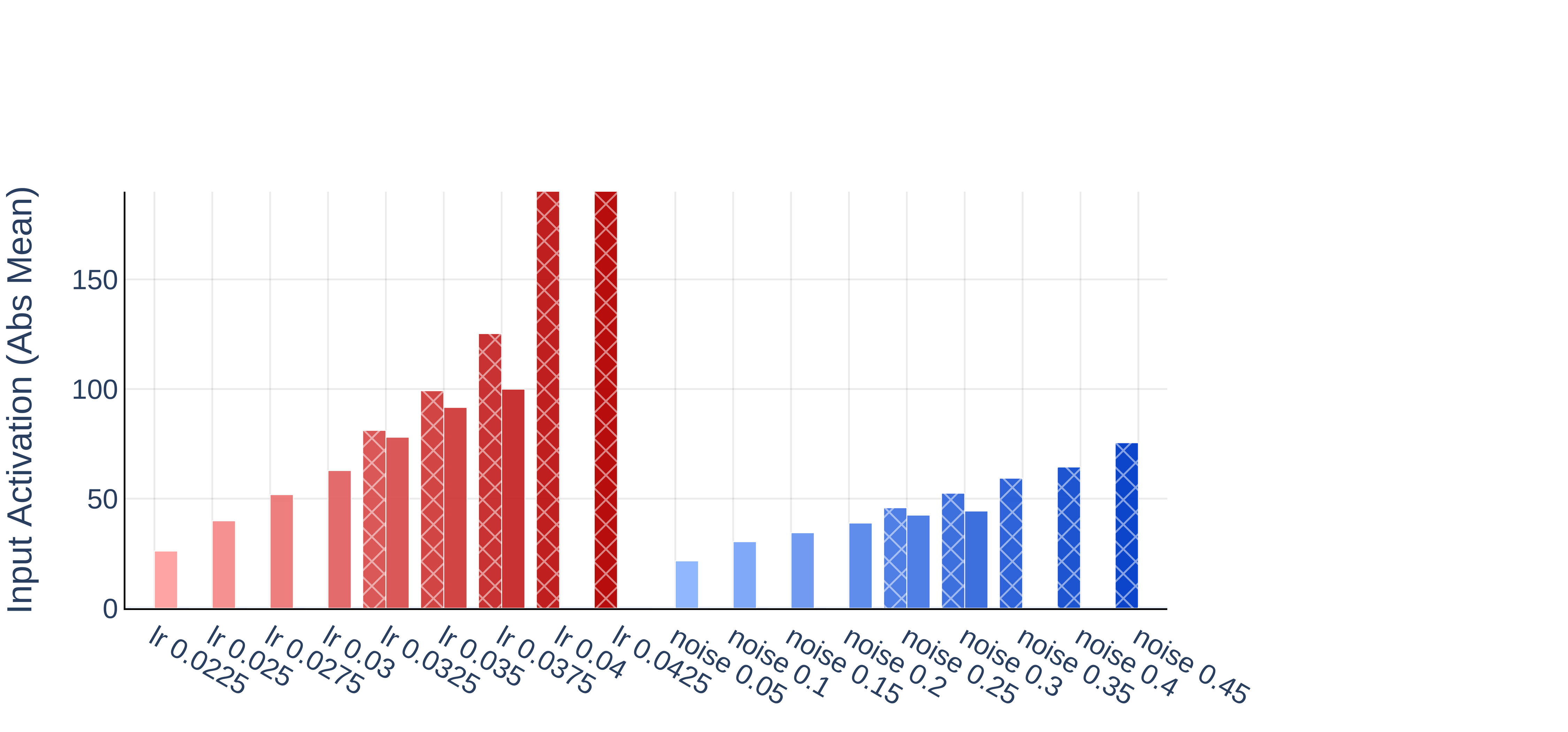} 
  \caption{2.8B dense model run statistics. \textbf{Left:} Parameter RMS norm. \textbf{Right:} Absolute mean of activation input to transformer layers.}
  \label{fig:act_analysis5}
\end{figure}

\begin{figure}[H]
    \centering
    \includegraphics[trim=0 0 1100 0,width=0.49
  \textwidth]{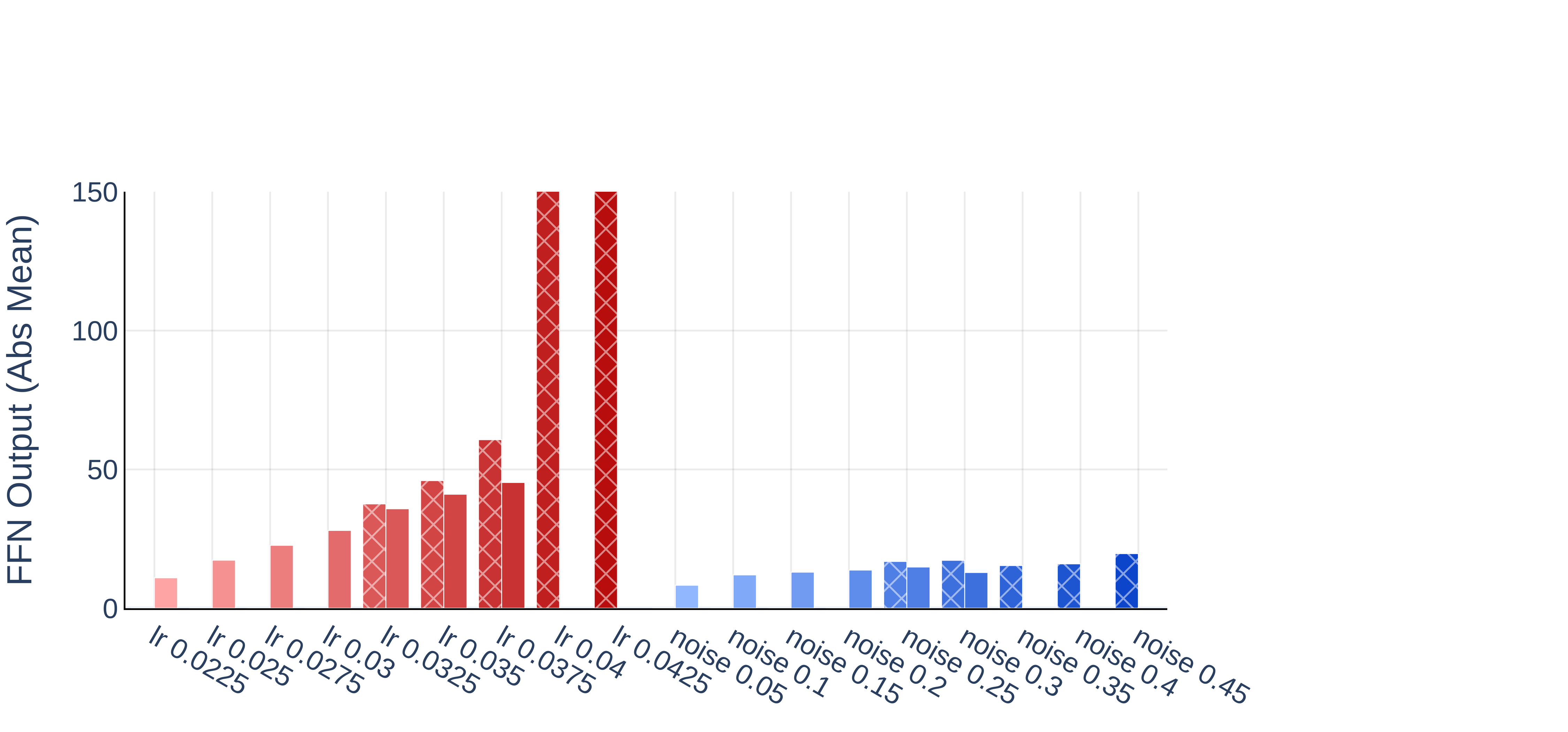}
    \includegraphics[trim=0 0 1100 0,width=0.49
  \textwidth]{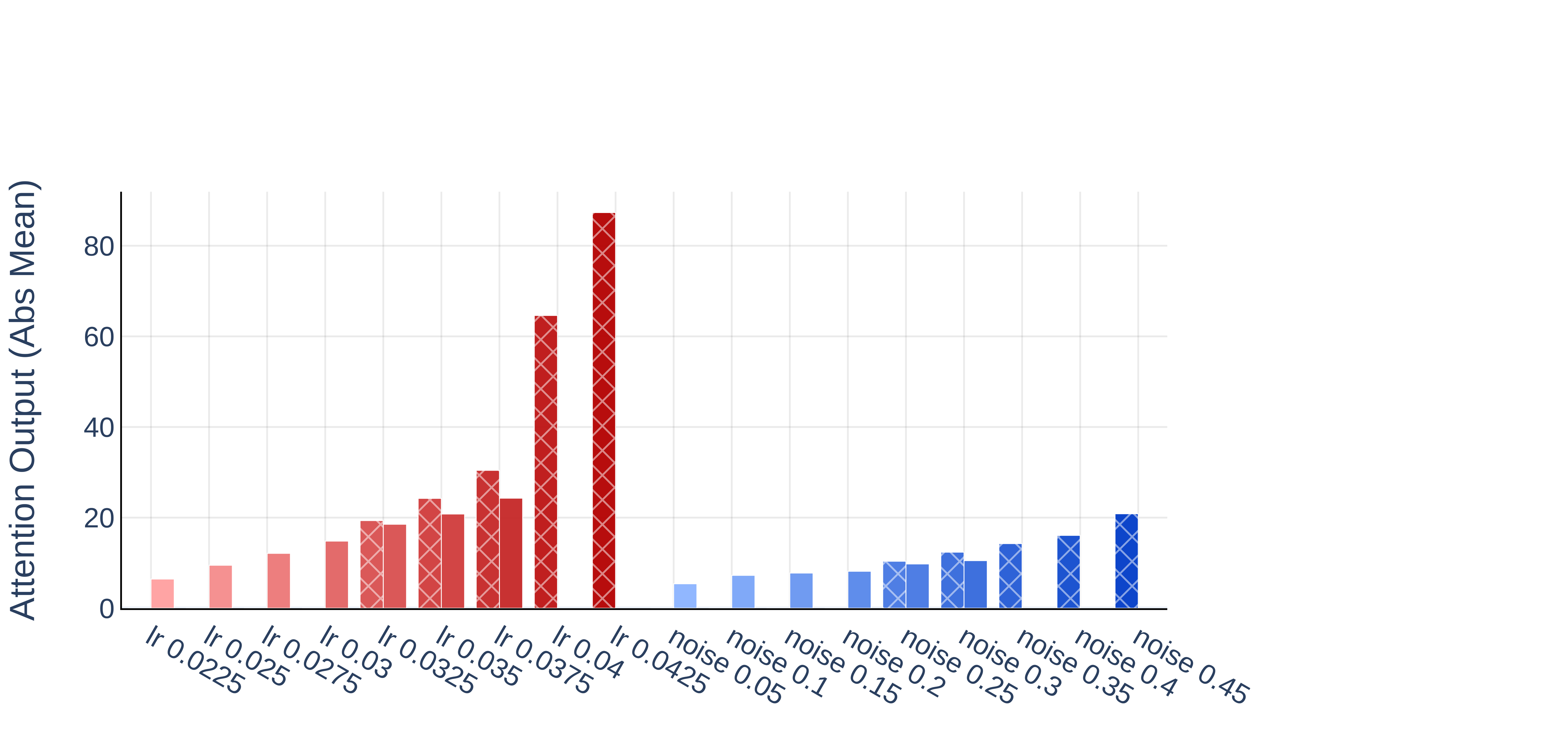}
      \includegraphics[trim=0 0 1100 0,width=0.49
  \textwidth]{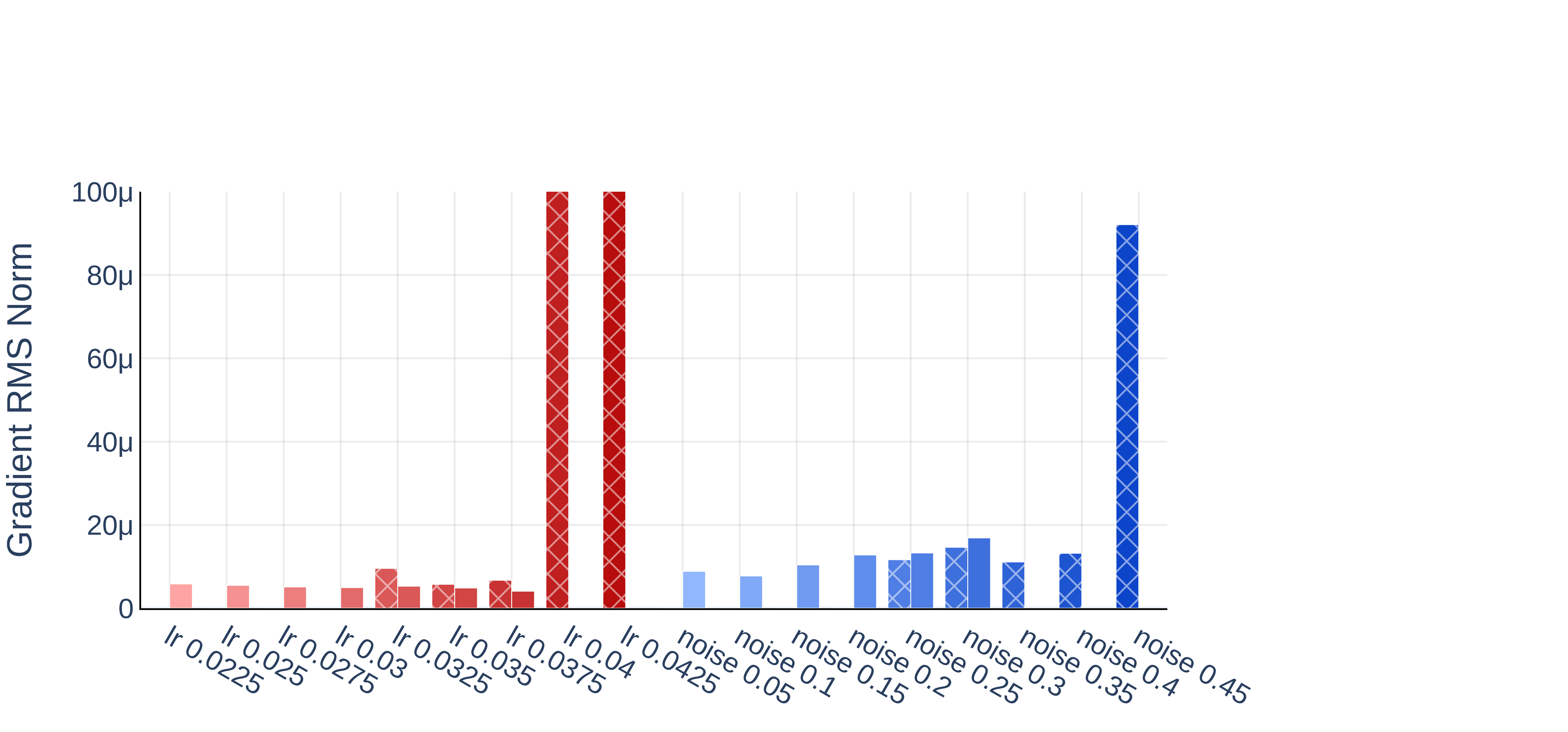}
  \caption{2.8B dense model run statistics. \textbf{Top left}: Absolute mean of pre-residual FFN output. \textbf{Top right}: Absolute mean of pre-residual attention output. \textbf{Bottom}: Gradient RMS norm.}
  \label{fig:act_analysis6}
\end{figure}

\subsection{1.3B active parameter MoE}
\begin{figure}[H]
    \includegraphics[trim=0 0 45 0,width=0.95
  \textwidth]{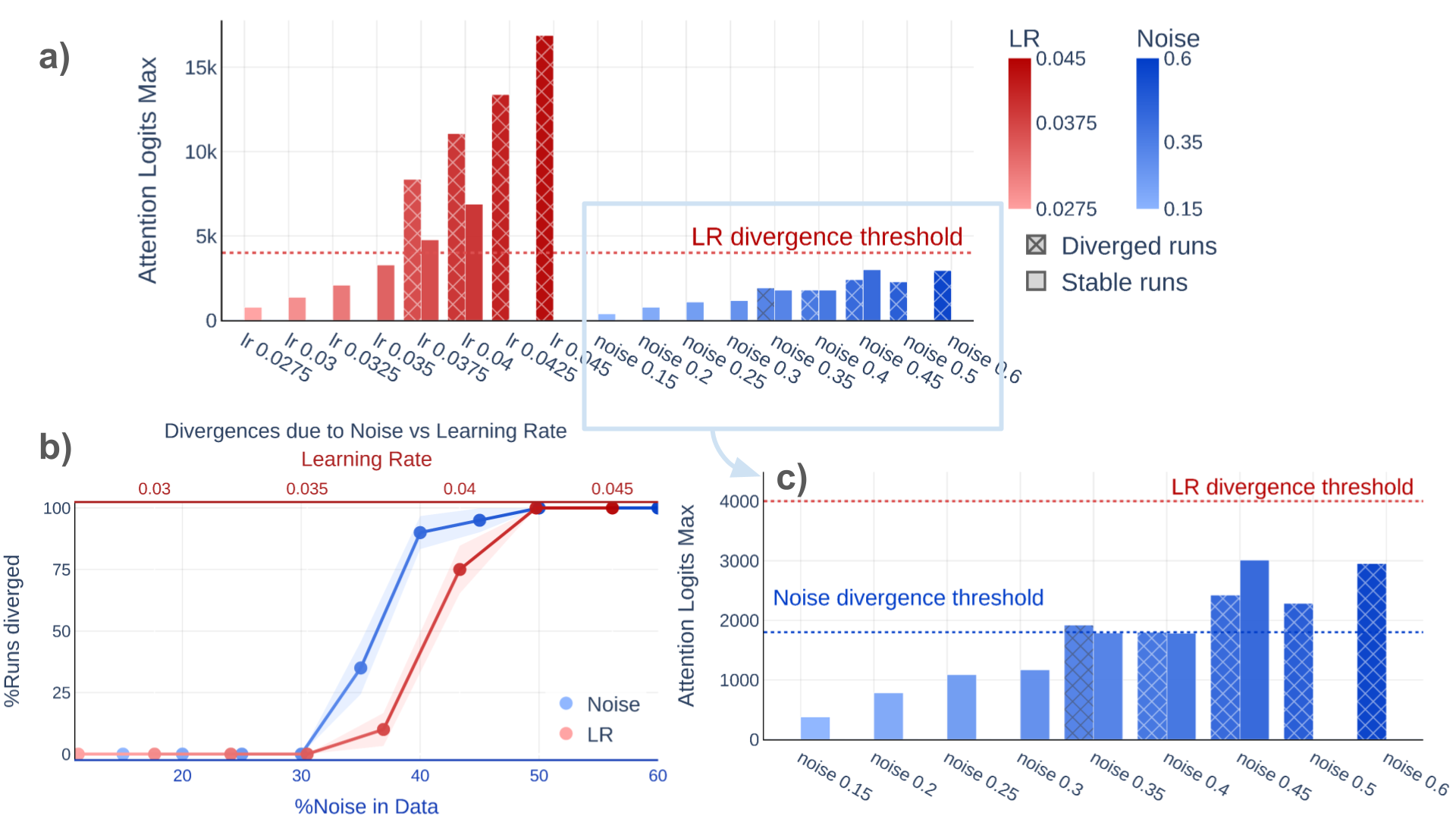}
  \caption{
    Examining the maximum attention logits for 1.3B active parameter MoE models .\textbf{(a)}
    \citeauthor{wortsman2023small} identify a model-size-invariant divergence threshold for high LR runs, shown by the dotted red line at 4000.
    \textbf{(b)} For comparability, LR and noise ratio ranges are selected to match divergence probabilities across the two settings.
    \textbf{(c)} A zoom-in on noisy data runs from (a) reveals a different and lower divergence threshold for noisy data settings at approximately 1800 (see the blue dotted lines). This noisy data divergence threshold also holds across different model sizes and MoE architectures.}
  \label{fig:act_moe_analysis2}
\end{figure}

\begin{figure}[H]
    \includegraphics[trim=0 0 1100 0,width=0.49
  \textwidth]{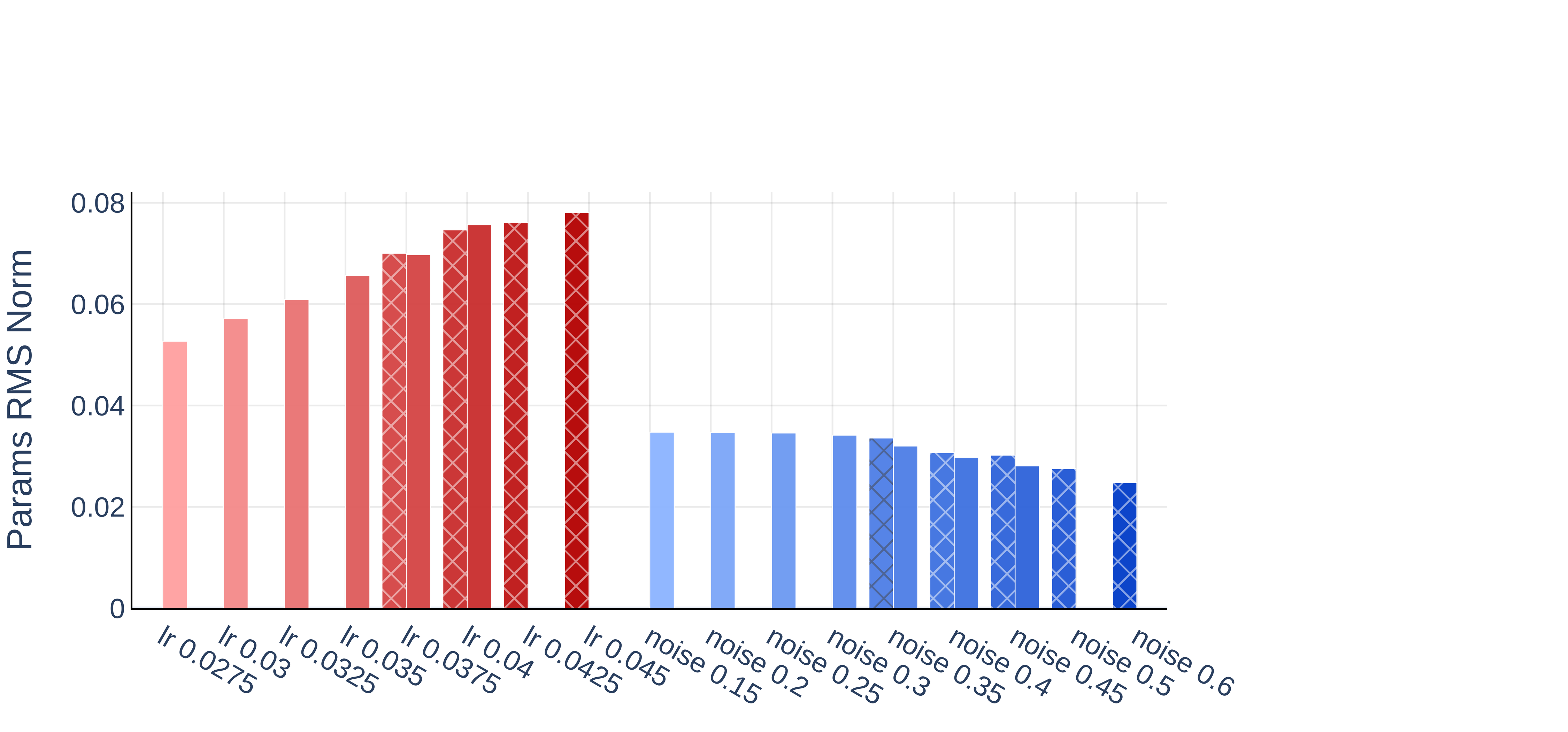} 
    \includegraphics[trim=0 0 1100 0,width=0.49
  \textwidth]{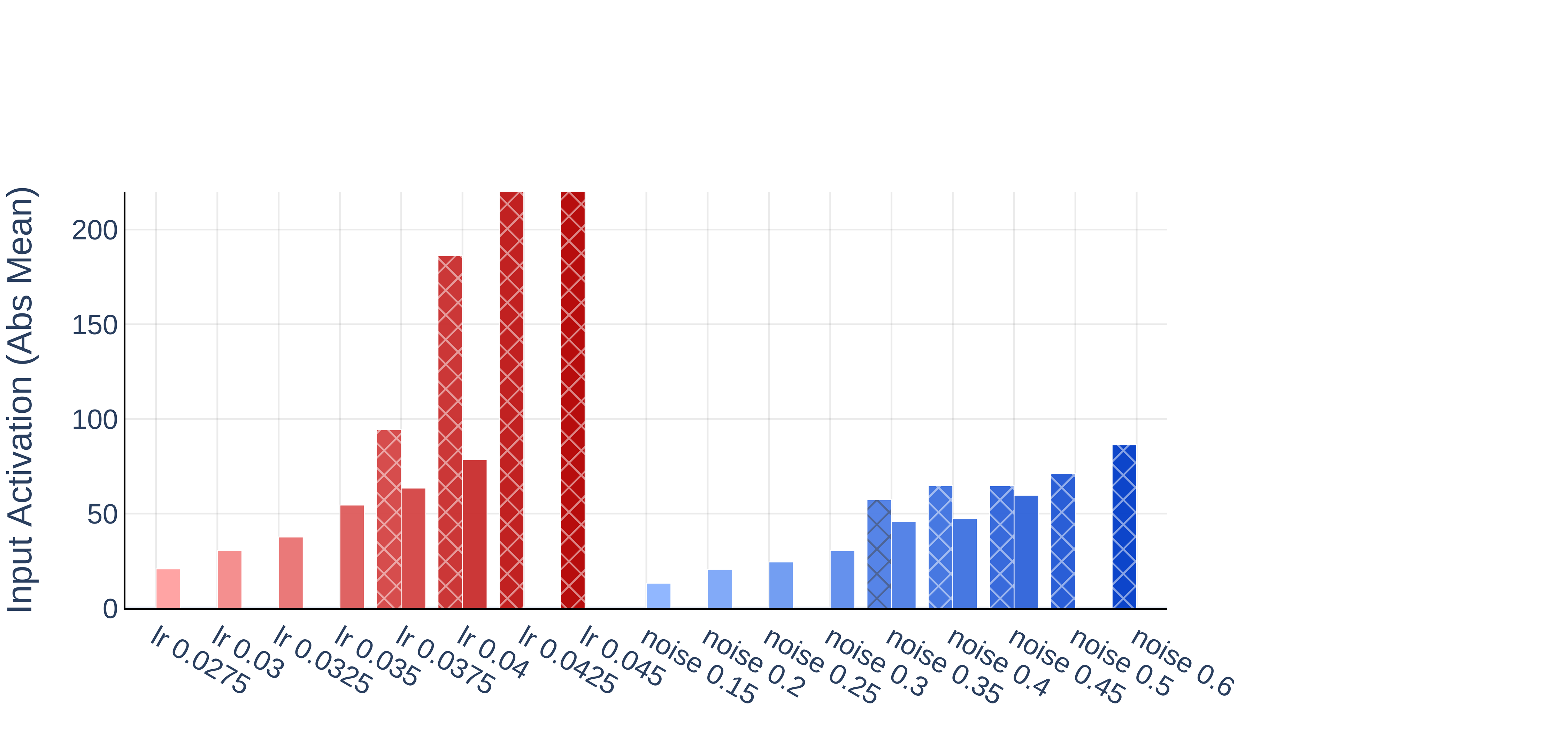}
  \caption{1.3B MoE model run statistics. \textbf{Left:} Parameter RMS norm. \textbf{Right:} Absolute mean of activation input to transformer layers.}
  \label{fig:act_moe analysis3}
\end{figure}

\begin{figure}[H]
    \includegraphics[trim=0 0 1100 0,width=0.49
  \textwidth]{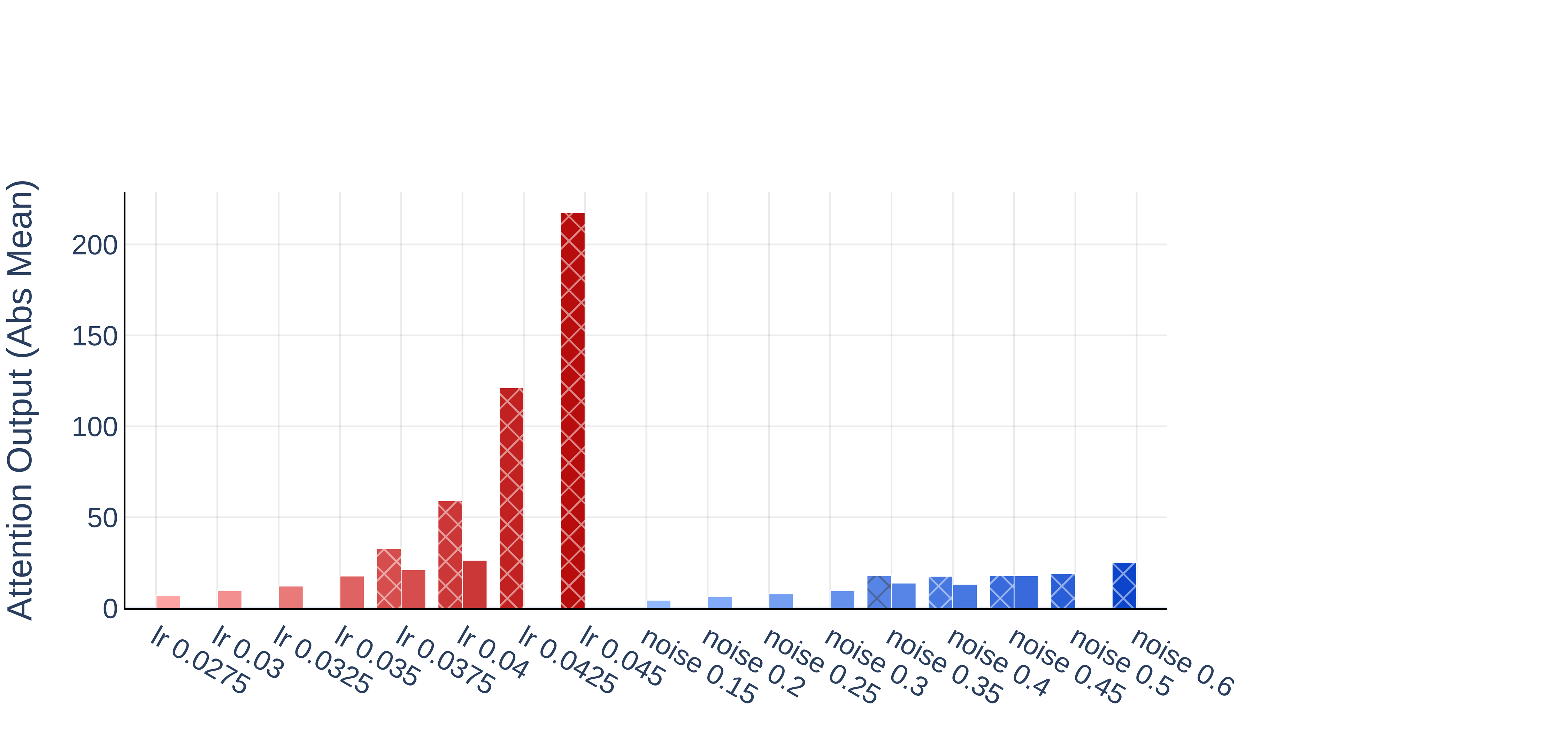}
      \includegraphics[trim=0 0 1100 0,width=0.49
  \textwidth]{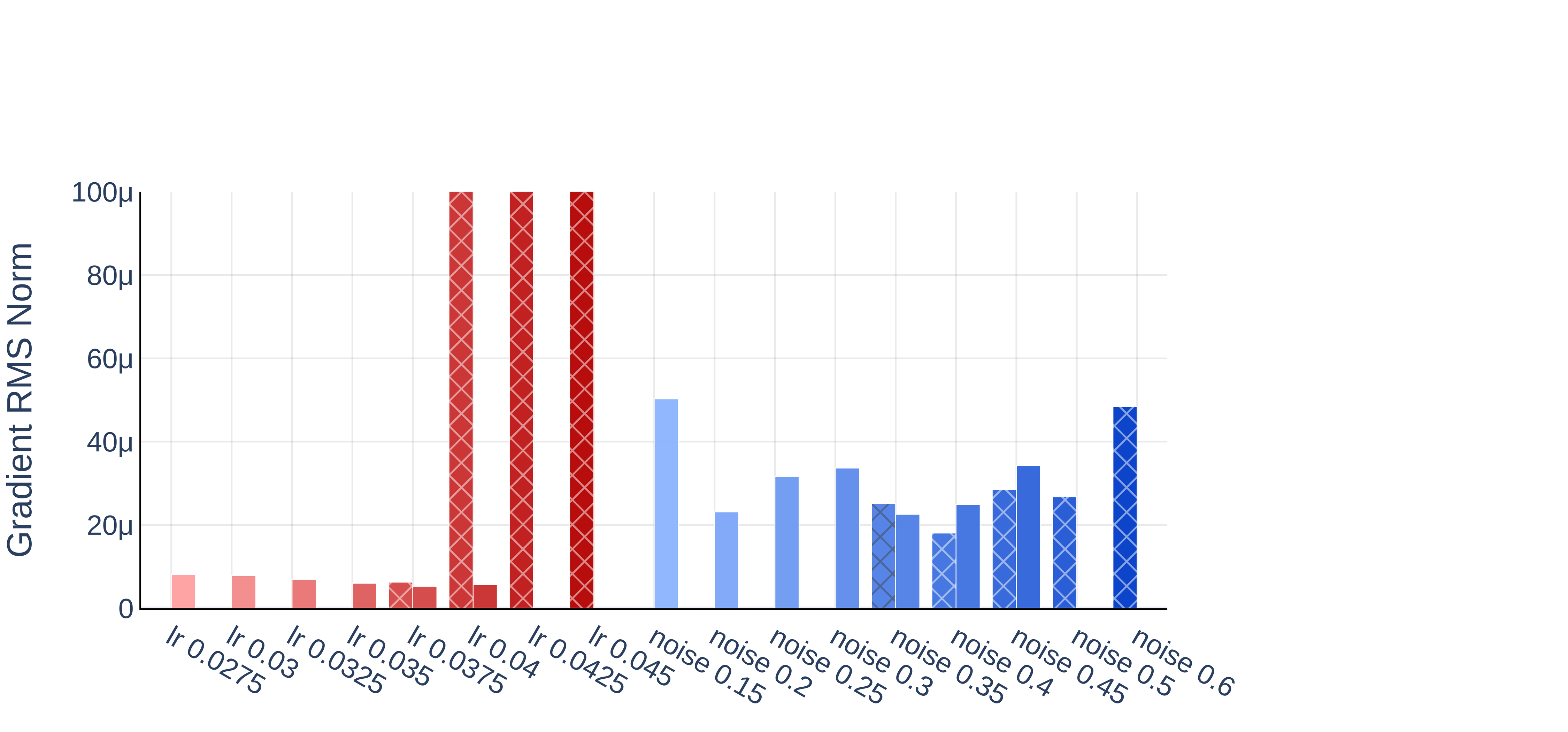}
  \caption{1.3B MoE model run statistics. \textbf{left}: Absolute mean of pre-residual attention output. \textbf{Right}: Gradient RMS norm.} 
  \label{fig:act_analysis7}
\end{figure}

\section{QK-layernorm interventions}

\begin{figure}[H]
    \centering
    \includegraphics[trim=0 0 45 0,width=0.95\textwidth]{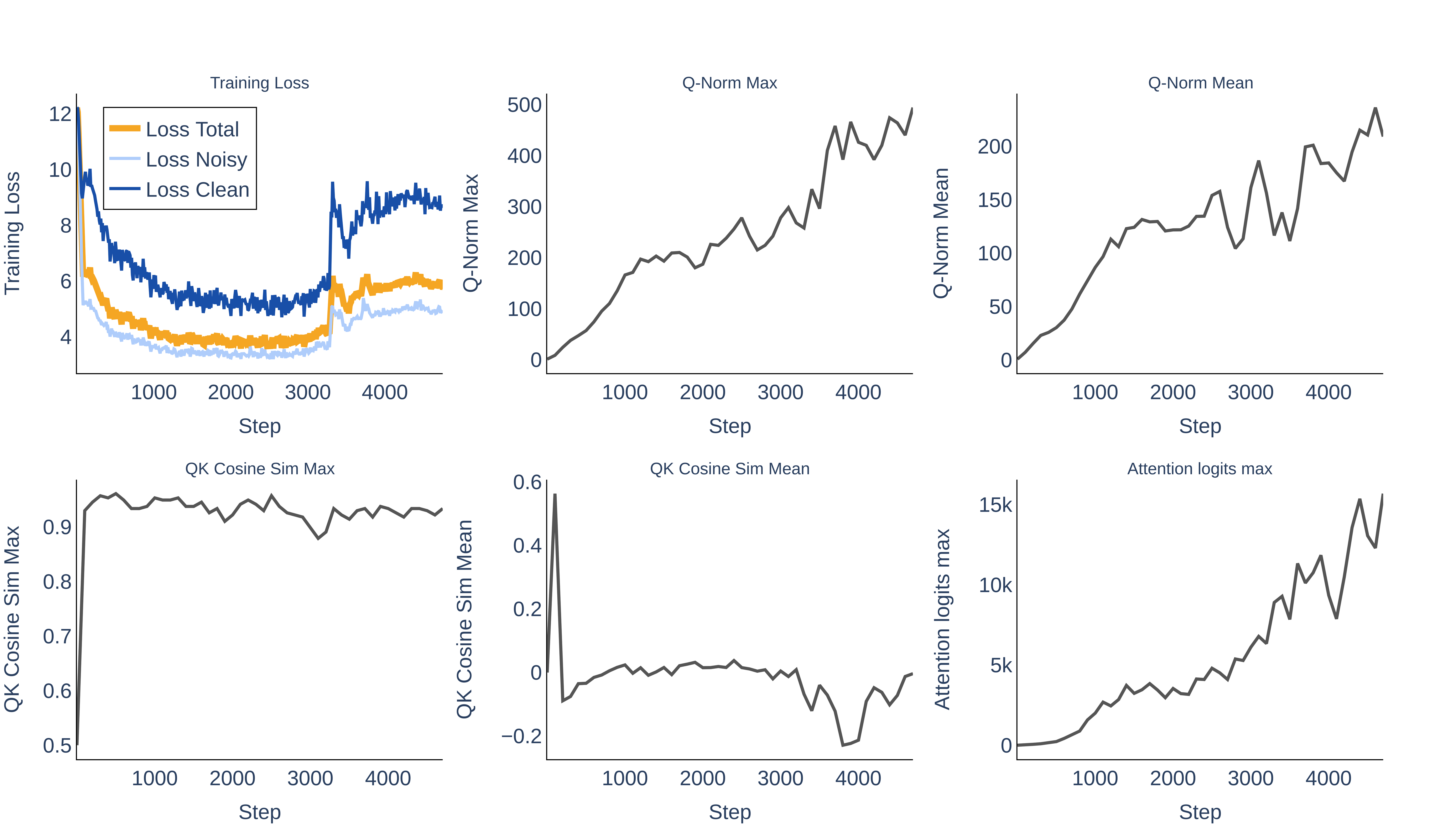}
  \caption{We show that growing Q-norms and K-norms are the cause of growing maximum attention logits, not the growing cosine similarity between the query and the key.}
  \label{fig:qnorms}
\end{figure}

\begin{figure}[H]
    \centering
    \includegraphics[trim=0 0 45 0,width=0.85\textwidth]{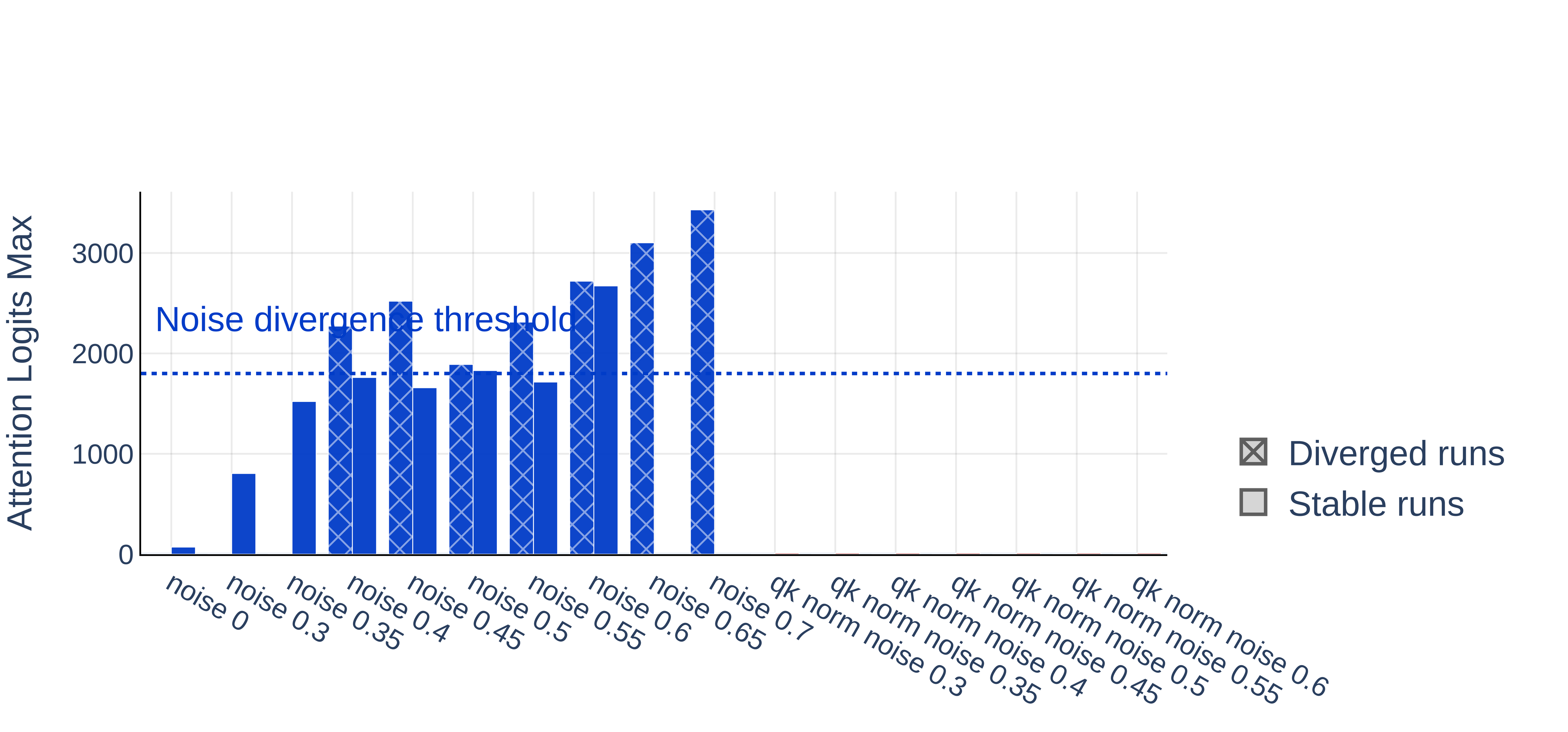}
    \includegraphics[trim=0 0 45 0,width=0.85\textwidth]{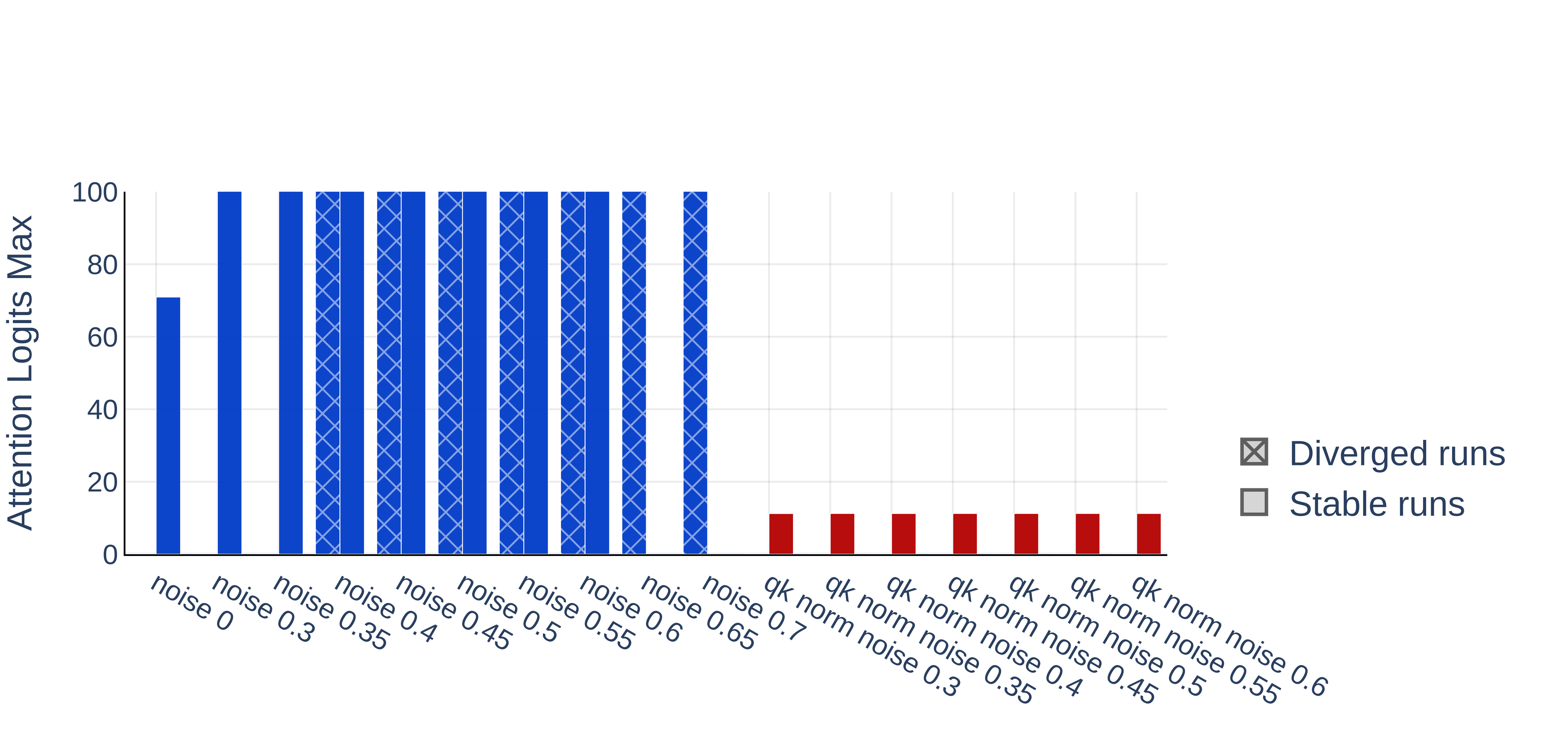}
  \caption{Maximum attention logits without (blue) versus with QK-layernorm (red). For clarity, we show a zoomed-in view of the plot on the bottom.}
  \label{fig:qnorms_zoomed}
\end{figure}

\section{MoE Router Analysis}
\label{sec:router_analysis}

\begin{figure}[H]
    \centering
    \includegraphics[trim=0 0 495 350, clip, width=0.24 \textwidth]{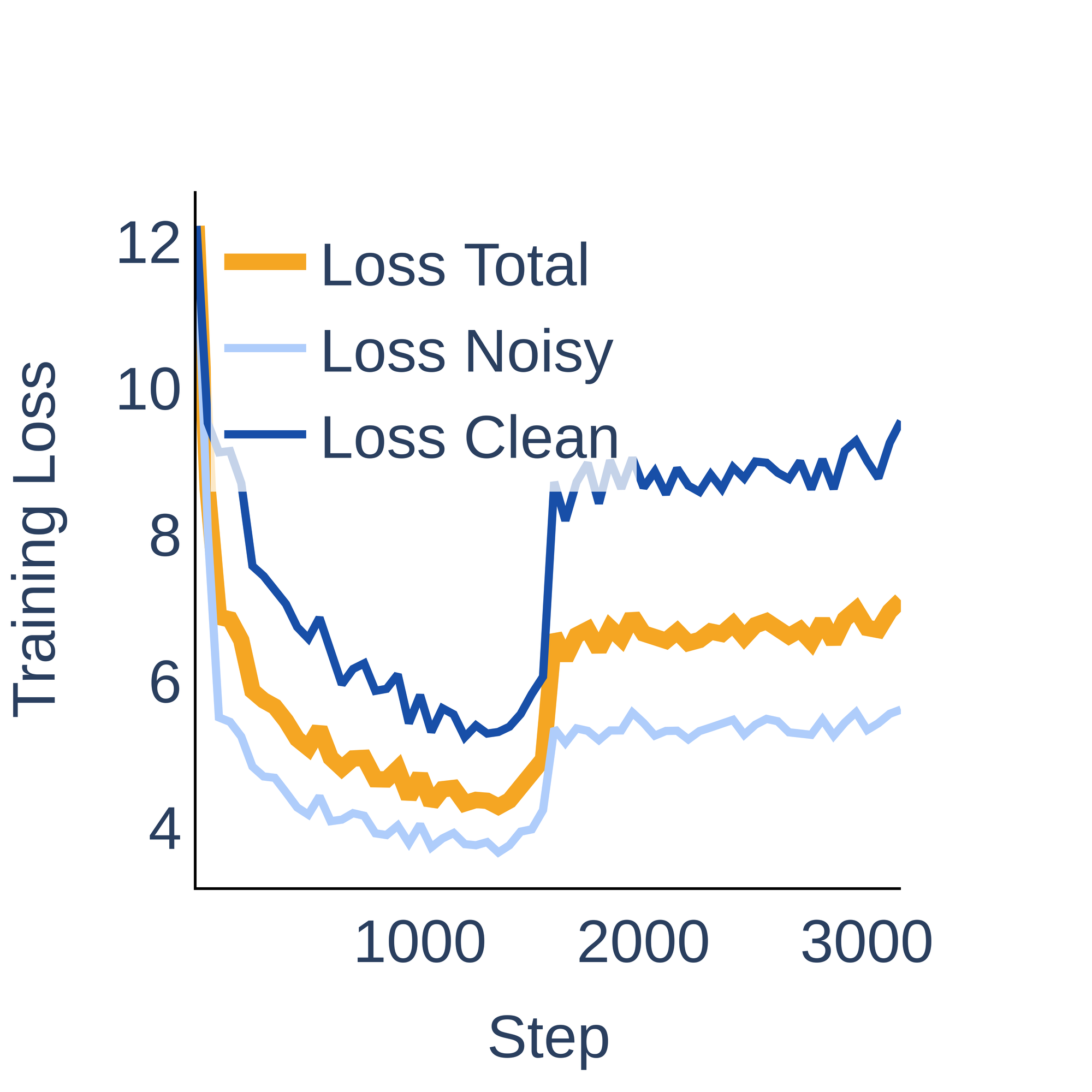}
    \includegraphics[trim=0 0 795 550, clip, width=0.24\textwidth]{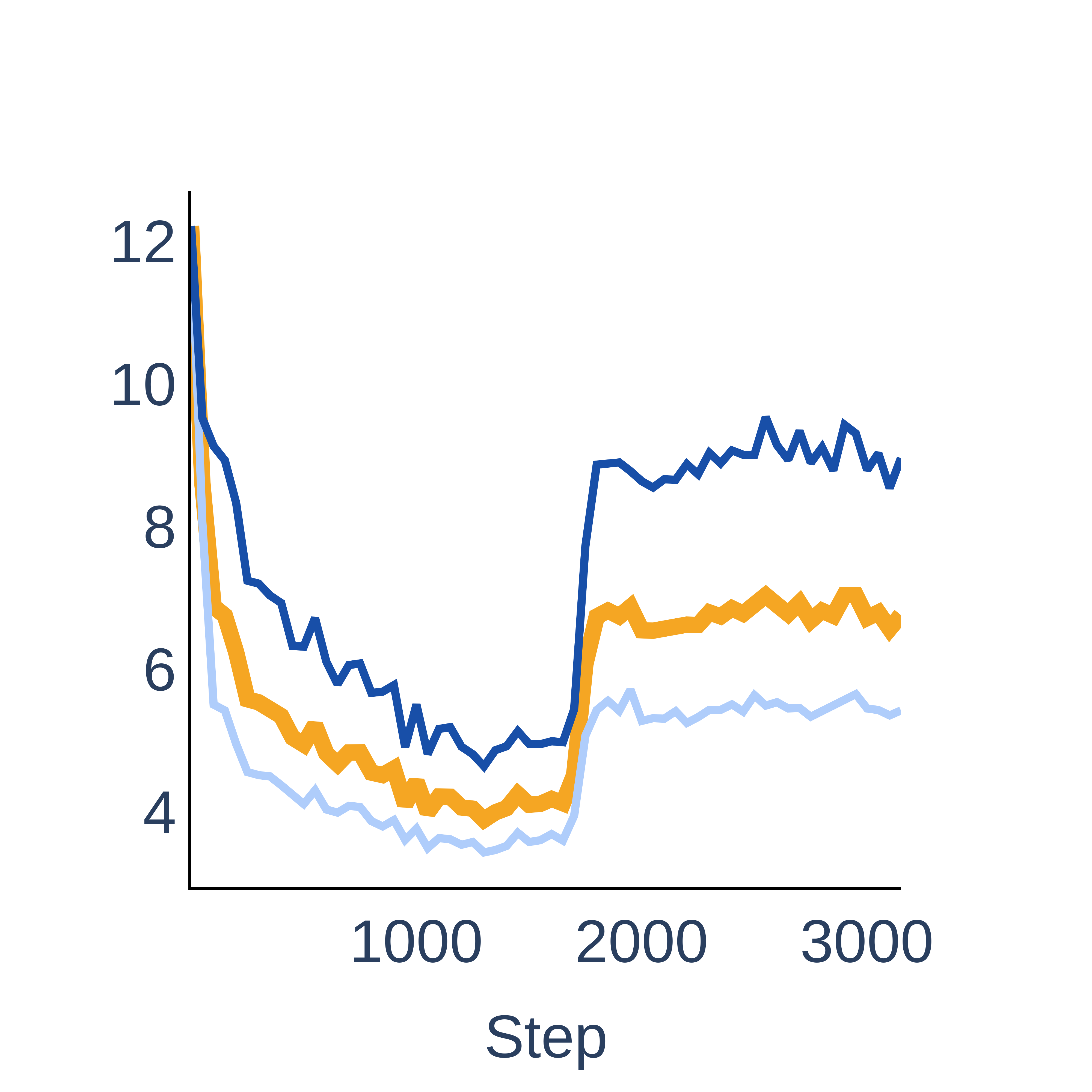}
    \includegraphics[trim=0 0 795 550, clip, width=0.24\textwidth]{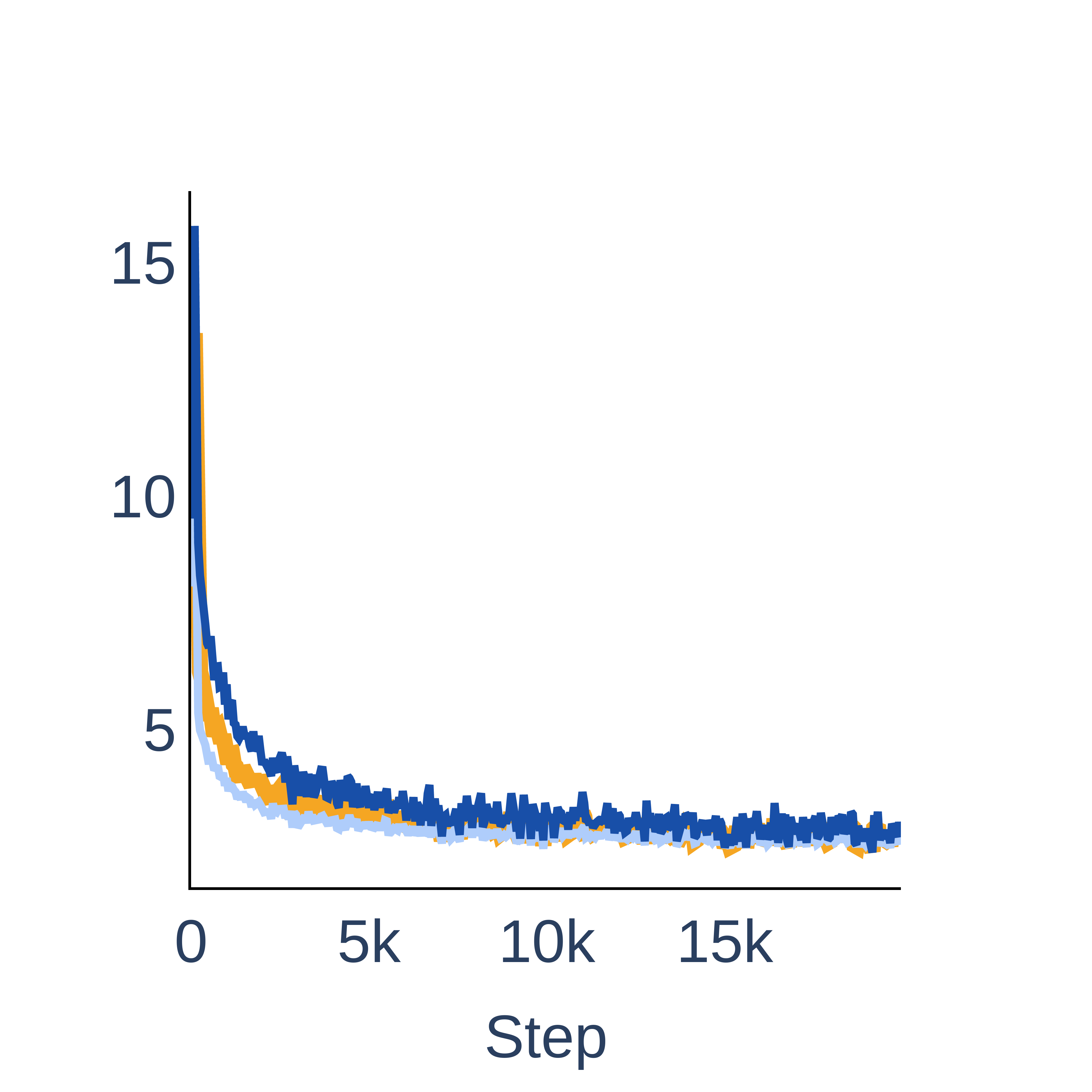}
    \includegraphics[trim=0 0 795 550, clip, width=0.24\textwidth]{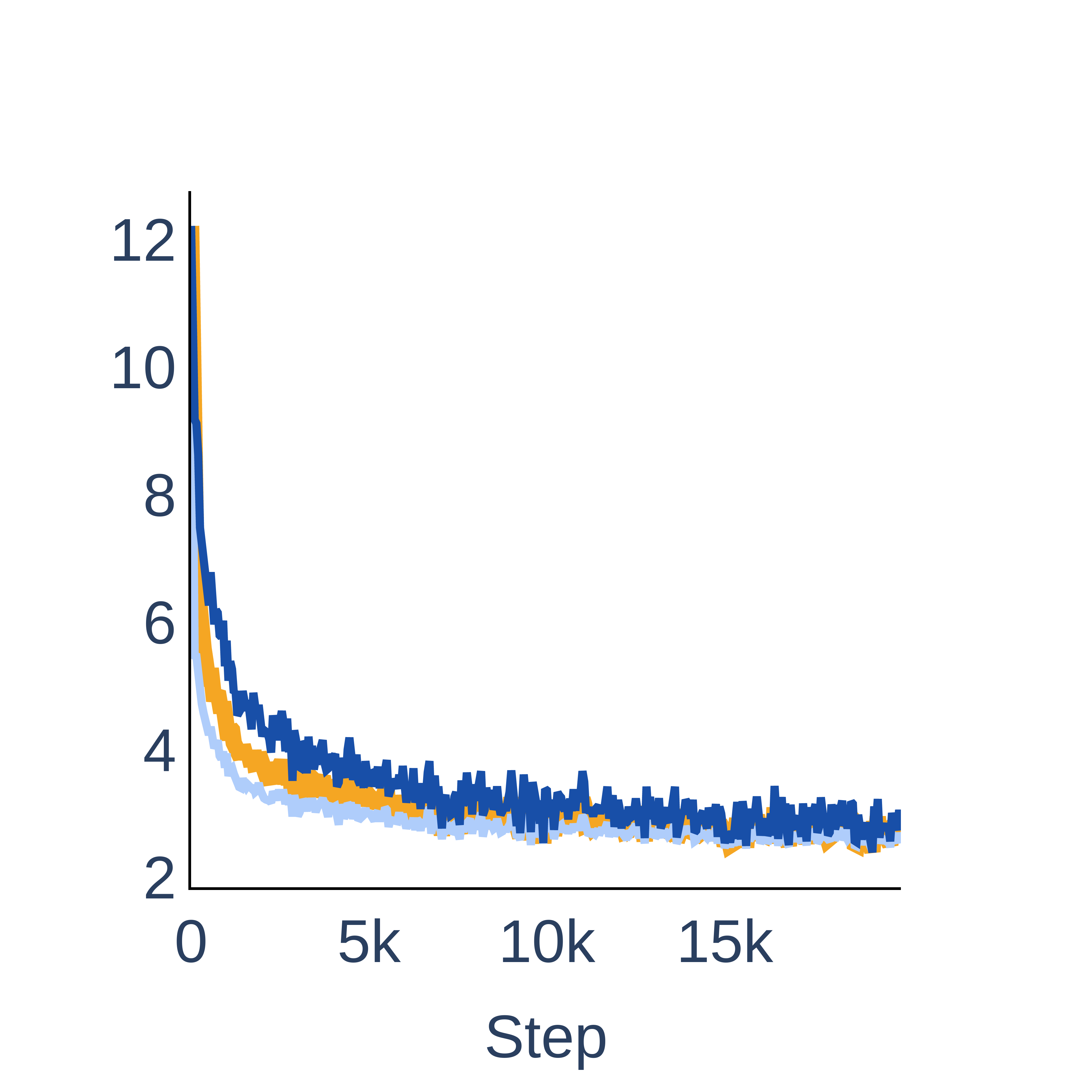}
    \caption{
    Training loss curves of 1.3B MoE runs with 35 \% noise and applying router z-loss. The only difference among the four runs is the random seed. We see that the first two runs diverge, while the latter two runs remain stable.
    }
  \label{fig:moe_divergence_exmaples}
\end{figure}

Below, we show the noisy token assignment for each of the four runs in \autoref{fig:moe_divergence_exmaples}. Although experts receive varying proportions of noisy tokens, this fraction is uncorrelated with experts' output activation magnitude, with an average Pearson correlation of near-zero across layers. This reinforces the conclusion that the MoE routing mechanism does not introduce additional sensitivity to noisy data.
\begin{figure}[H]
    \centering
    \includegraphics[trim=0 0 45 0,width=0.95\textwidth]{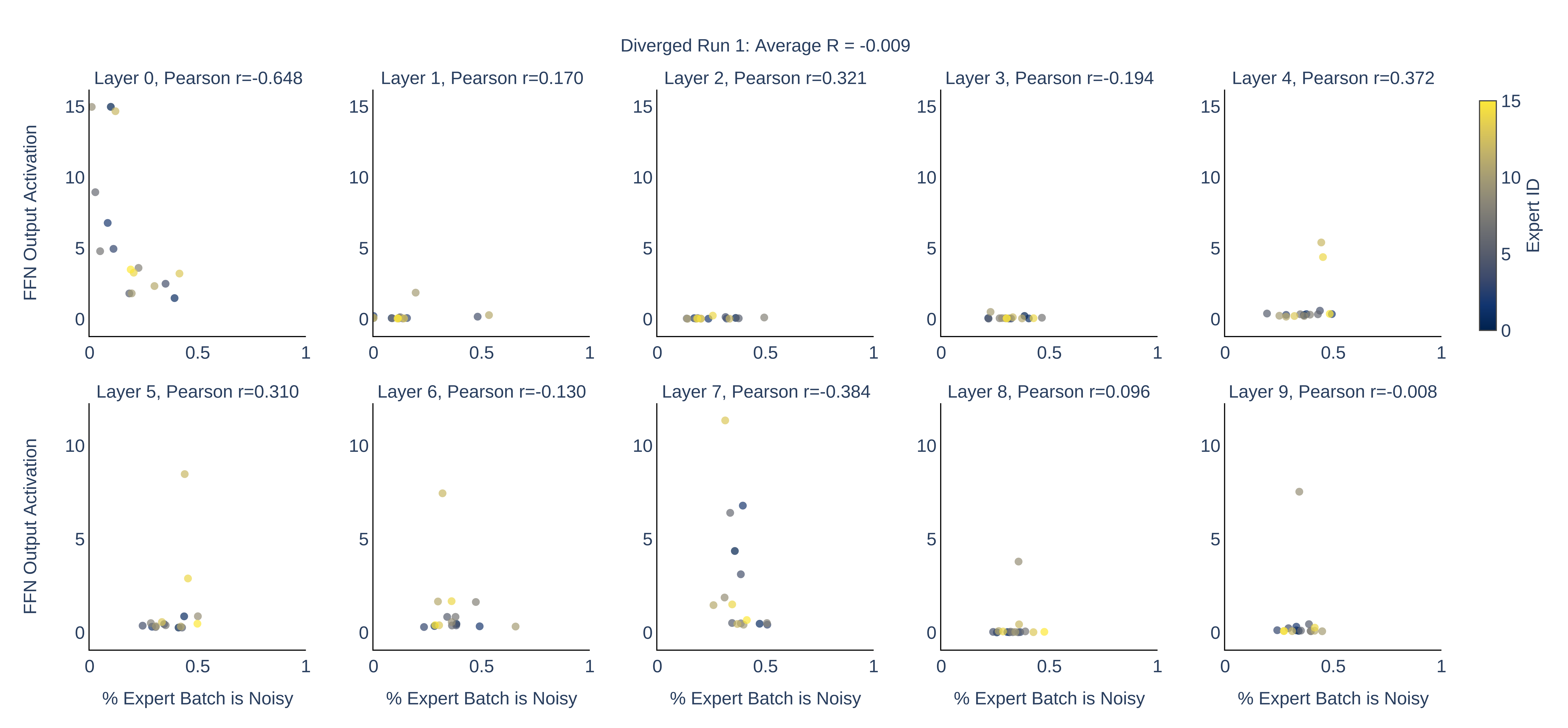}
  \caption{Noisy token assignment has very little correlation to FFN activation.}
  \label{fig:router0}
\end{figure}

\begin{figure}[H]
    \centering
    \includegraphics[trim=0 0 45 0,width=0.95\textwidth]{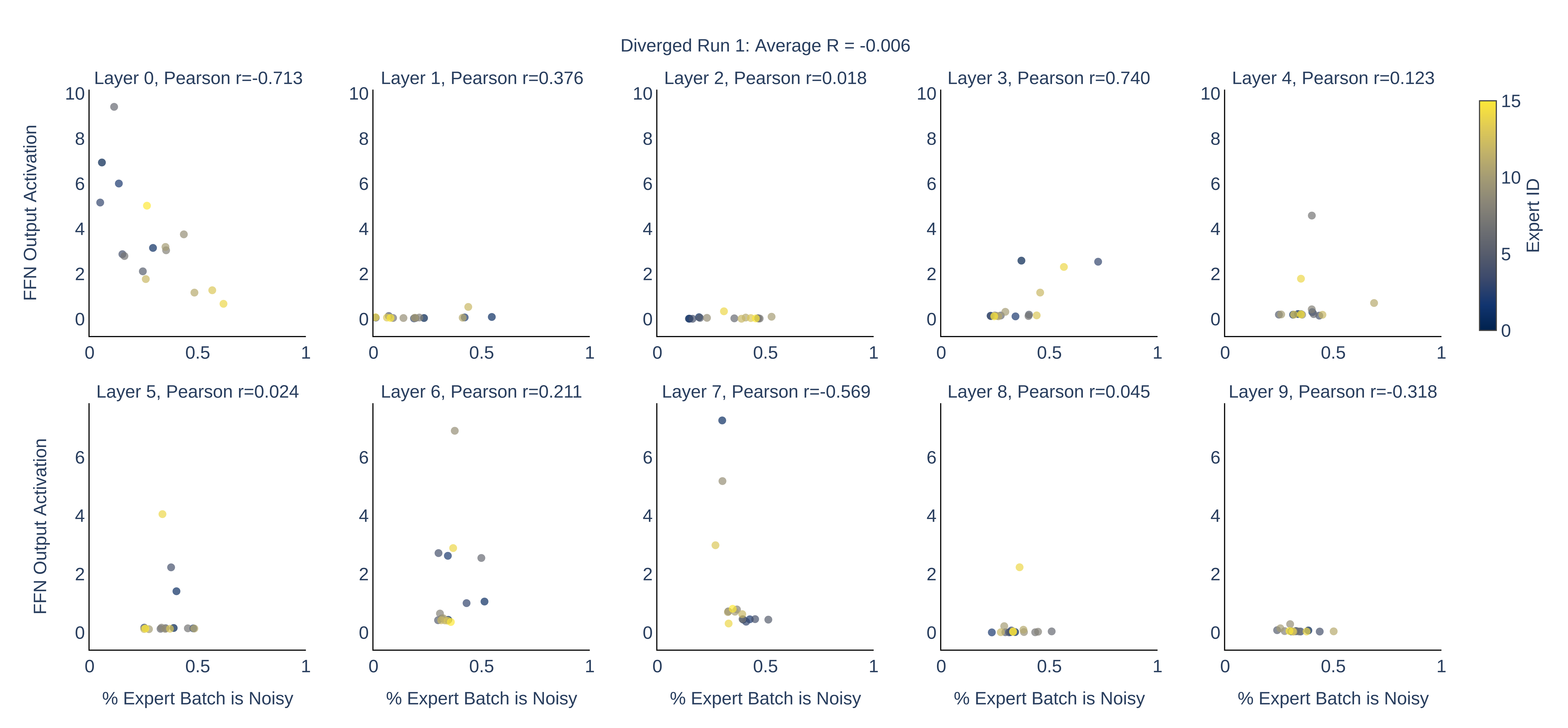}
  \caption{Noisy token assignment has very little correlation to FFN activation.}
  \label{fig:router1}
\end{figure}

\begin{figure}[H]
    \centering
    \includegraphics[trim=0 0 45 0,width=0.95\textwidth]{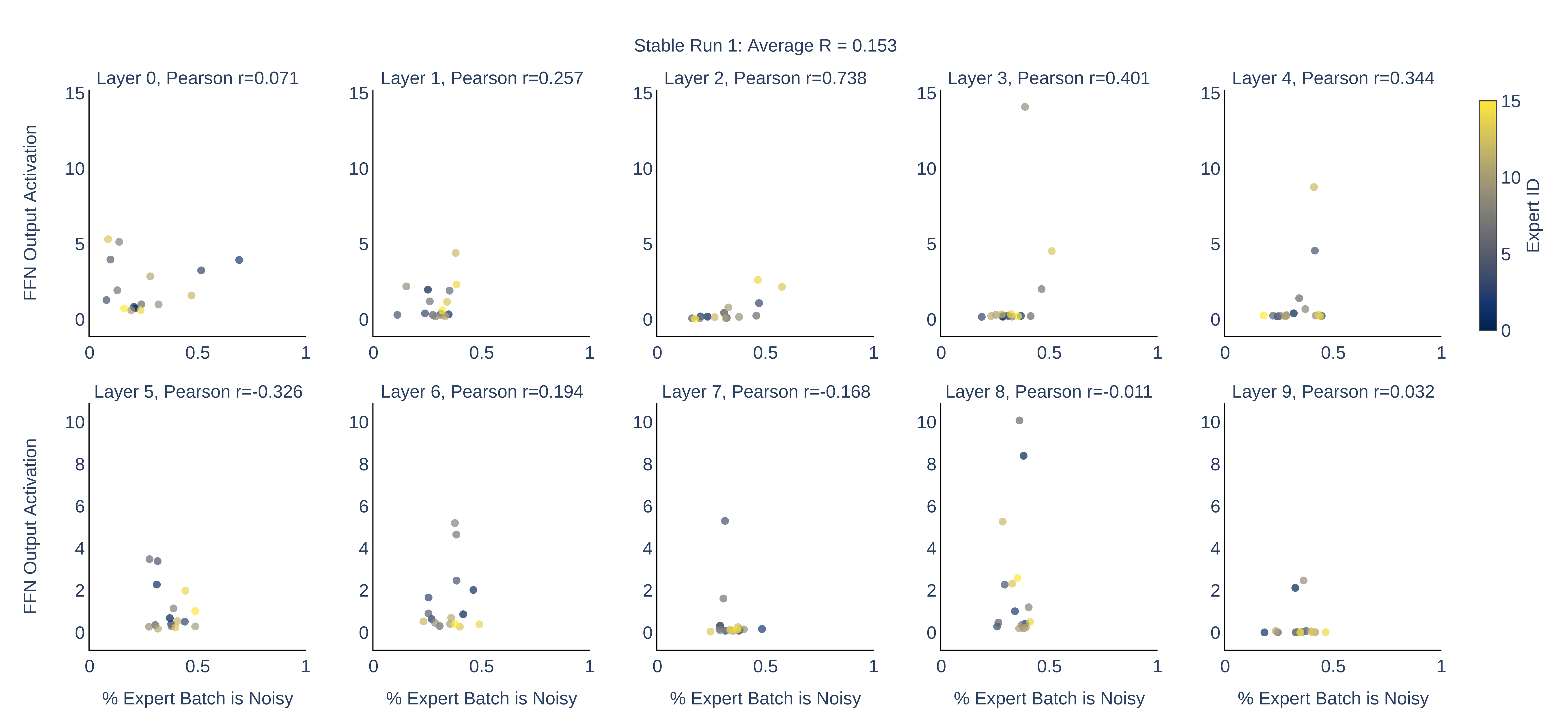}
  \caption{Noisy token assignment has very little correlation to FFN activation.}
  \label{fig:router2}
\end{figure}

\begin{figure}[H]
    \centering
    \includegraphics[trim=0 0 45 0,width=0.95\textwidth]{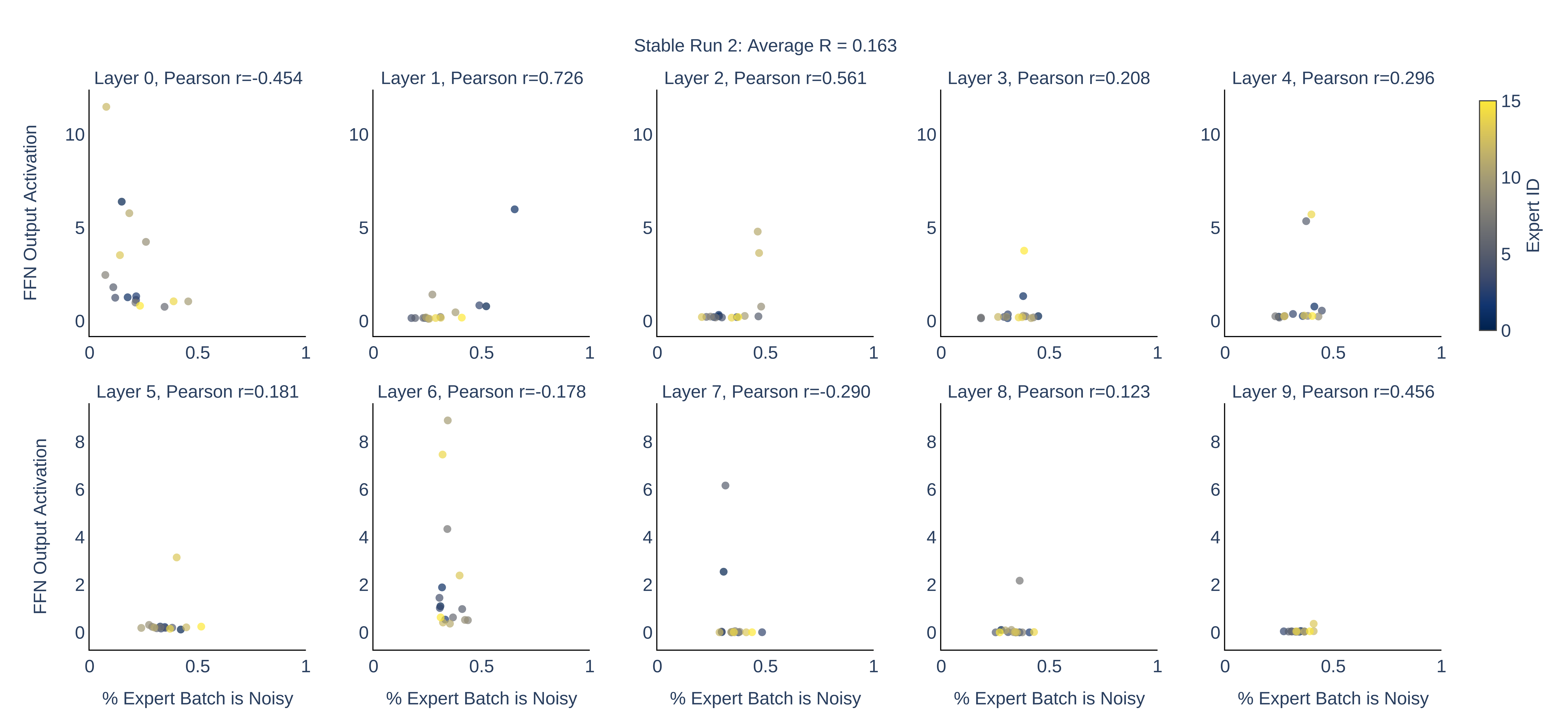}
  \caption{Noisy token assignment has very little correlation to FFN activation.}
  \label{fig:router3}
\end{figure}

\end{document}